\RequirePackage{snapshot}

\documentclass[10pt,twocolumn,letterpaper]{article}

\usepackage[pagenumbers]{cvpr} %

\usepackage{graphicx}
\usepackage{amsmath}
\usepackage{amssymb}
\usepackage{booktabs}
\usepackage[export]{adjustbox}

\usepackage{placeins}
\usepackage{graphbox}
\usepackage{breqn}
\usepackage{multirow}
 \usepackage{mathtools}

\usepackage[pagebackref,breaklinks,colorlinks]{hyperref}

\usepackage[capitalize]{cleveref}
\crefname{section}{Sec.}{Secs.}
\Crefname{section}{Section}{Sections}
\Crefname{table}{Table}{Tables}
\crefname{table}{Tab.}{Tabs.}

\def\cvprPaperID{5452} %
\def\confName{CVPR}
\def\confYear{2022}

\newcommand\blfootnote[1]{%
  \begingroup
  \renewcommand\thefootnote{}\footnote{#1}%
  \addtocounter{footnote}{-1}%
  \endgroup
}

\providecommand{\imwidth}{}

\providecommand{\impath}[1]{}
\providecommand{\impaths}[1]{}
\providecommand{\impatha}[1]{}
\providecommand{\impathb}[1]{}
\providecommand{\impathc}[1]{}
\providecommand{\impathd}[1]{}
\providecommand{\impathe}[1]{}

\newcommand{\perceptualcompression}{
\begin{figure}[!t]
	\centering
   \includegraphics[width=0.38\textwidth]{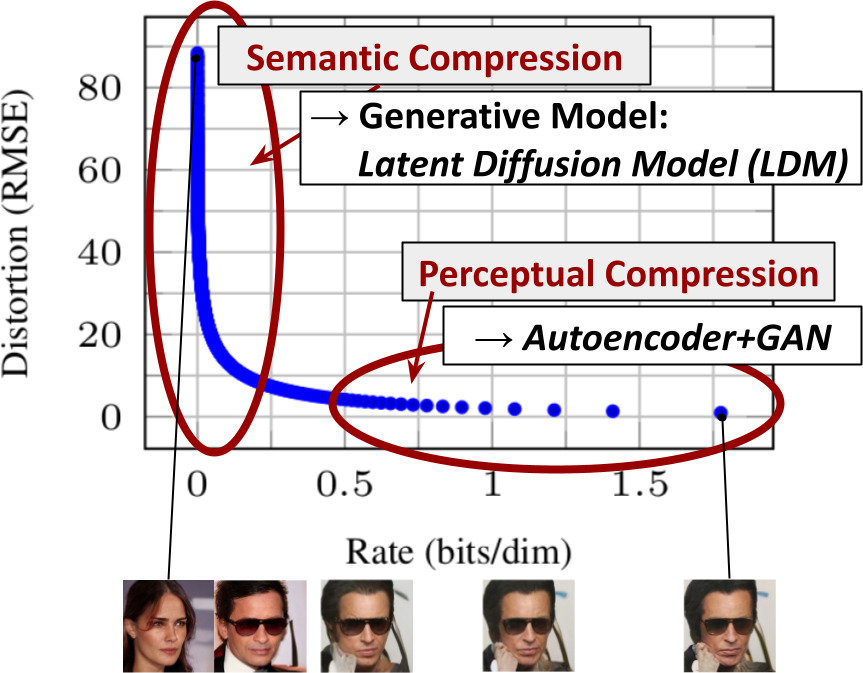}
\caption{\label{fig:perceptualcompression}
Illustrating perceptual and semantic compression: Most bits of a digital image correspond to imperceptible details. While DMs allow to suppress this semantically meaningless information by minimizing the responsible loss term, gradients (during training) and the neural network backbone (training and inference) still need to be evaluated on all pixels, leading to superfluous computations and unnecessarily expensive optimization and inference.
\\
We propose \emph{latent diffusion models (LDMs)} as an effective generative model and a separate mild compression stage that only eliminates imperceptible details.
Data and images from \cite{DBLP:conf/nips/HoJA20}.\vspace{-1.25em}
}
\end{figure}
}

\newcommand{\crossattnfig}{
\begin{figure}[tbp]
\centering
\includegraphics[trim=0 705 305 2, clip, width=0.5\textwidth]{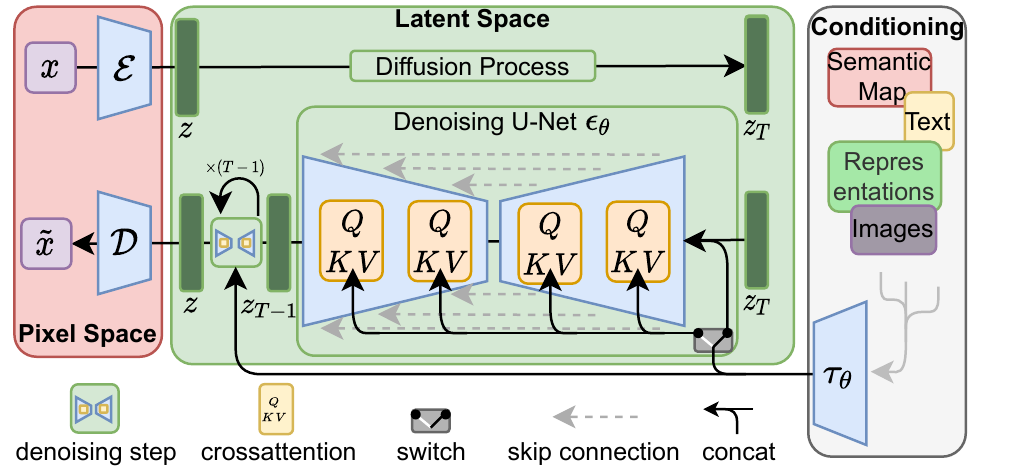}
  \caption{\label{fig:conditioning} We condition LDMs either via concatenation %
  or by a more general cross-attention mechanism. See Sec.~\ref{subsec:conditioning} \vspace{-1.5em}}
\end{figure}
}

\newcommand{\thicksample}{
\begin{figure}[htbp]
\centering
\includegraphics[width=0.5\textwidth]{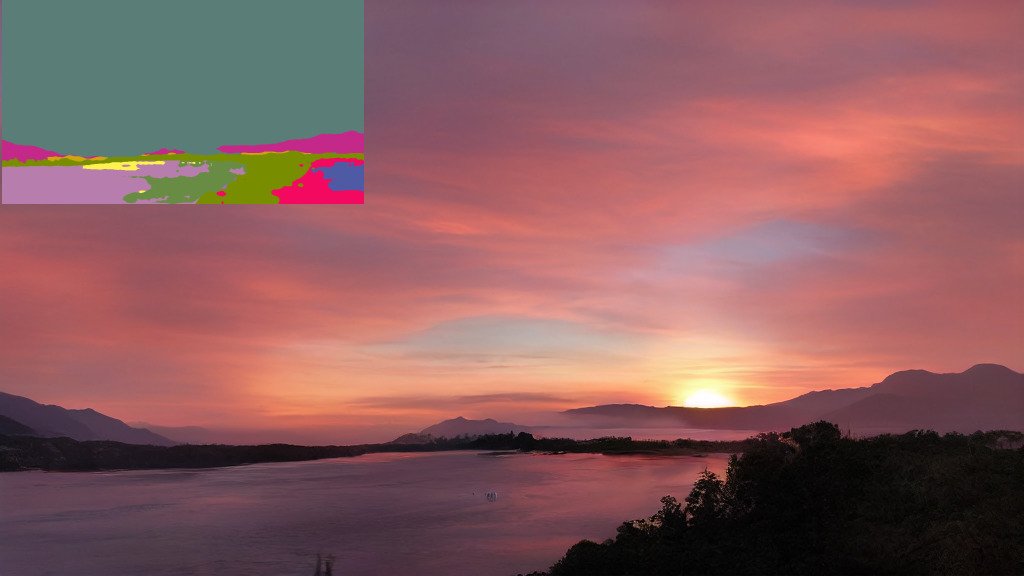}\vspace{-0.65em}
  \caption{\label{fig:thicksample} A \emph{LDM} trained on $256^2$ resolution can generalize to 
  larger resolution (here: $512\times1024$) for spatially conditioned tasks
  such as semantic synthesis of landscape images. See
  Sec.~\ref{subsubsec:beyond}.\vspace{-1em}
  }
\end{figure}
}

\newcommand{\cincompression}{
\begin{figure}[htbp]
\begin{subfigure}{.275\textwidth}
\centering
   \includegraphics[width=\linewidth]{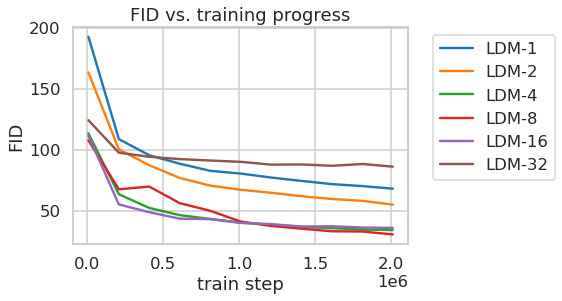}
\end{subfigure}
\hfill
\begin{subfigure}{.195\textwidth}
\centering
\includegraphics[width=\linewidth]{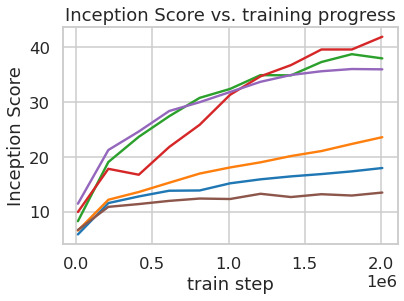}
\end{subfigure}
\caption{\label{fig:cin_traincourse}Analyzing the training of class-conditional \emph{LDMs} with different downsampling factors $f$ over 2M train steps on the ImageNet dataset. 
Pixel-based \emph{LDM-1}
requires substantially larger train times compared to models with larger downsampling factors (\emph{LDM-$\{$4-16$\}$}). 
Too much perceptual compression as in \emph{LDM-32} limits the overall sample quality.
All models are trained on a single NVIDIA A100 with the same computational budget. 
Results obtained with 100 DDIM steps~\cite{DBLP:conf/iclr/SongME21} and $\kappa = 0$.\vspace{-1em} }
\end{figure}
}

\newcommand{\firststagecomparison}{
\begin{figure}[thbp]
	\centering
    \renewcommand{\imwidth}{0.5\textwidth}
    \renewcommand{\impaths}[1]{img/firststagereconstructions/new/##1} 
    \setlength{\tabcolsep}{1pt}
    \renewcommand{\arraystretch}{1}
	\begin{adjustbox}{max width=\linewidth} %
 	\begin{tabular}{c c c c}
  	\toprule
  	\Huge \textbf{Input} & \shortstack{\Huge \textbf{ours ($f=4$)} \\ \huge
    PSNR: $27.4$ R-FID: $0.58$} &  \shortstack{\Huge \textbf{DALL-E ($f=8$)}
    \\ \huge PSNR: $22.8$ R-FID: $32.01$}  & \shortstack{\Huge \textbf{VQGAN
    ($f=16$)} \\ \huge PSNR: $19.9$ R-FID: $4.98$}\\
  	\midrule
	\includegraphics[width=\imwidth]{\impaths{up1}} &	         
		\includegraphics[width=\imwidth]{\impaths{up2}} &	         
			\includegraphics[width=\imwidth]{\impaths{up3}} &	         
				\includegraphics[width=\imwidth]{\impaths{up4}} \\

	\includegraphics[width=\imwidth]{\impaths{down1}} &	         
		\includegraphics[width=\imwidth]{\impaths{down2}} &	         
			\includegraphics[width=\imwidth]{\impaths{down3}} &	         
				\includegraphics[width=\imwidth]{\impaths{down4}} \\	         
  	\bottomrule
	\end{tabular}
	\end{adjustbox}
	\caption{\label{fig:firststagecomparison} Boosting the upper bound on achievable quality with less agressive downsampling.
	Since diffusion models offer excellent inductive biases for spatial data, we do not need the heavy spatial downsampling of related generative models in latent space, but can still greatly reduce the dimensionality of the data via suitable autoencoding models, see Sec.~\ref{sec:method}.
	Images are from the DIV2K \cite{DBLP:conf/cvpr/AgustssonT17} validation set, evaluated at $512^2$ px. We denote the spatial
	downsampling factor by $f$. Reconstruction FIDs %
	\cite{FID} and PSNR are calculated on ImageNet-val. \cite{DBLP:conf/cvpr/DengDSLL009}; 
	see also Tab.~\ref{tab:firststagetablecomplete}.
	} \vspace{-1.25em}
\end{figure}
}

\newcommand{\speeeeed}{
\begin{figure}[thbp]
\begin{subfigure}{.275\textwidth}
\centering
   \includegraphics[width=\linewidth]{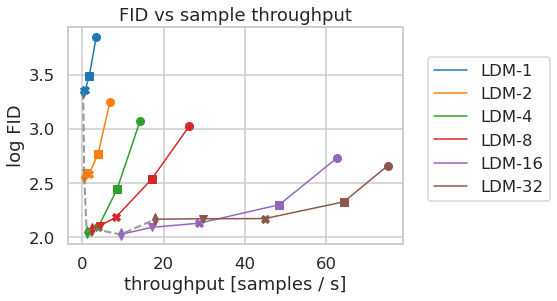}
\end{subfigure}
\hfill
\begin{subfigure}{.195\textwidth}
\centering
\includegraphics[width=\linewidth]{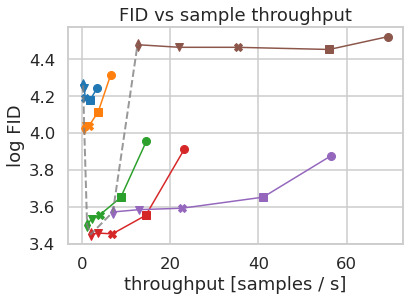}
\end{subfigure}
\caption{\label{fig:speedplot} Comparing
  \emph{LDMs} with varying compression on the CelebA-HQ (left) and
  ImageNet (right) datasets. Different markers indicate $\{10, 20, 50, 100,
  200\}$ sampling steps using DDIM, from right to left along
  each line. The dashed line shows the FID scores for 200 steps, indicating the
  strong performance of \emph{LDM-$\{$4-8$\}$}.
  FID scores assessed on 5000 samples. All models
  were trained for 500k (CelebA) / 2M (ImageNet) steps on an A100.\vspace{-2em}
}
\end{figure}
}

\newcommand{\bboxandtexttoimgsamples}{
\begin{figure}[tbp]
\centering
\parbox[b]{.5\textwidth}{
\small
	\renewcommand{\imwidth}{0.09\textwidth}
    \renewcommand{\impath}[1]{img/supplement/layouts/##1}%
    \renewcommand{\impathd}[1]{img/cr/text2img/##1}%
    \renewcommand{\impatha}[1]{img/goodsamples/bbox2img/street/##1} 
	\renewcommand{\impathb}[1]{img/goodsamples/bbox2img/table/##1} 
    \renewcommand{\impathc}[1]{img/goodsamples/bbox2img/surfer/##1}
    \renewcommand{\impathe}[1]{img/goodsamples/text2img/sunrise/##1} 
	\setlength{\tabcolsep}{0pt}
	\begin{tabular}{c@{\hskip 2pt}cccc}
	\toprule
	\includegraphics[align=c,width=\imwidth]{\impatha{cond-2_}} &	         
	\includegraphics[align=c,width=\imwidth]{\impatha{sample-216}} &	
	\includegraphics[align=c,width=\imwidth]{\impatha{sample-251}} &	        
	\includegraphics[align=c,width=\imwidth]{\impatha{sample-439}} &  
    \includegraphics[align=c,width=\imwidth]{\impatha{sample-266}} \\
    \rule{0pt}{5ex}%

	\includegraphics[align=c,width=\imwidth]{\impath{train/cond-283}} &
	\includegraphics[align=c,width=\imwidth]{\impath{train/sample-270}} &
	\includegraphics[align=c,width=\imwidth]{\impath{train/sample-330}} &
	\includegraphics[align=c,width=\imwidth]{\impath{train/sample-390}} &
	\includegraphics[align=c,width=\imwidth]{\impath{train/sample-282}} \\
    \rule{0pt}{5ex}%
	\includegraphics[align=c,width=\imwidth]{\impath{flower/cond-99}} &
	\includegraphics[align=c,width=\imwidth]{\impath{flower/sample-98}} &
	\includegraphics[align=c,width=\imwidth]{\impath{flower/sample-134}} &
	\includegraphics[align=c,width=\imwidth]{\impath{flower/sample-158}} &
	\includegraphics[align=c,width=\imwidth]{\impath{flower/sample-206}}\\
	\midrule
	\bottomrule
  \end{tabular}\vspace{-0.5em}
}
\caption{\label{fig:bboxandtxt2img} Layout-to-image synthesis with an \emph{LDM} on COCO~\cite{DBLP:conf/cvpr/CaesarUF18}, see Sec.~\ref{subsubsec:crossattn2img}. Quantitative evaluation in the supplement \ref{suppsec:bboxtoimage}. 
 \vspace{-1em}}
\end{figure}
}

\newcommand{\smallsamples}{
\begin{figure*}[htbp]
\hfill%
  \parbox[b]{\textwidth}{
\centering
\small
    \renewcommand{\imwidth}{0.065\textwidth}
    \renewcommand{\impatha}[1]{img/goodsamples/celeba/##1} 
	\renewcommand{\impathb}[1]{img/goodsamples/ffhq/##1} 
    \renewcommand{\impathc}[1]{img/goodsamples/churches/##1} 
    \renewcommand{\impathd}[1]{img/goodsamples/beds/##1} 
    \renewcommand{\impathe}[1]{img/goodsamples/cin_cherries/##1} 
	\setlength{\tabcolsep}{0pt}		
 \begin{tabular}{ccc@{\hskip 2pt}ccc@{\hskip 2pt}ccc@{\hskip 2pt}ccc@{\hskip 2pt}ccc}
    \toprule
 	 \multicolumn{3}{c}{CelebAHQ} & \multicolumn{3}{c}{FFHQ} & \multicolumn{3}{c}{LSUN-Churches} & \multicolumn{3}{c}{LSUN-Beds}& \multicolumn{3}{c}{ImageNet} \\ 
  \toprule
	\includegraphics[align=c,width=\imwidth]{\impatha{sample-7}} &	         
	\includegraphics[align=c,width=\imwidth]{\impatha{sample-10}} &	        
	\includegraphics[align=c,width=\imwidth]{\impatha{sample-11}} &	         
	
	\includegraphics[align=c,width=\imwidth]{\impathb{sample-4}} &	         
	\includegraphics[align=c,width=\imwidth]{\impathb{sample-5}} &	        
	\includegraphics[align=c,width=\imwidth]{\impathb{sample-6}} &
	
	\includegraphics[align=c,width=\imwidth]{\impathc{sample-48}} &	         
	\includegraphics[align=c,width=\imwidth]{\impathc{sample-60}} &	        
	\includegraphics[align=c,width=\imwidth]{\impathc{sample-71}} &
	
	\includegraphics[align=c,width=\imwidth]{\impathd{sample-1}} &	         
	\includegraphics[align=c,width=\imwidth]{\impathd{sample-29}} &	        
	\includegraphics[align=c,width=\imwidth]{\impathd{sample-32}} &
	
	\includegraphics[align=c,width=\imwidth]{\impathe{sample-928}} &	         
	\includegraphics[align=c,width=\imwidth]{\impathe{sample-1}} &	        
	\includegraphics[align=c,width=\imwidth]{\impathe{sample-171}} 	
	\\
	\includegraphics[align=c,width=\imwidth]{\impatha{sample-18}} &	         
	\includegraphics[align=c,width=\imwidth]{\impatha{sample-23}} &	        
	\includegraphics[align=c,width=\imwidth]{\impatha{sample-32}} &	         
	
	\includegraphics[align=c,width=\imwidth]{\impathb{sample-26}} &	         
	\includegraphics[align=c,width=\imwidth]{\impathb{sample-29}} &	        
	\includegraphics[align=c,width=\imwidth]{\impathb{sample-32}} &
	
	\includegraphics[align=c,width=\imwidth]{\impathc{sample-82}} &	         
	\includegraphics[align=c,width=\imwidth]{\impathc{sample-97}} &	        
	\includegraphics[align=c,width=\imwidth]{\impathc{sample-98}} &
	
	\includegraphics[align=c,width=\imwidth]{\impathd{sample-35}} &	         
	\includegraphics[align=c,width=\imwidth]{\impathd{sample-42}} &	        
	\includegraphics[align=c,width=\imwidth]{\impathd{sample-44}} &
	
	\includegraphics[align=c,width=\imwidth]{\impathe{sample-1042}} &	         
	\includegraphics[align=c,width=\imwidth]{\impathe{sample-48}} &	        
	\includegraphics[align=c,width=\imwidth]{\impathe{sample-116}} 	
	\\	
	\includegraphics[align=c,width=\imwidth]{\impatha{sample-34}} &	         
	\includegraphics[align=c,width=\imwidth]{\impatha{sample-42}} &	        
	\includegraphics[align=c,width=\imwidth]{\impatha{sample-45}} &	         
	
	\includegraphics[align=c,width=\imwidth]{\impathb{sample-45}} &	         
	\includegraphics[align=c,width=\imwidth]{\impathb{sample-49}} &	        
	\includegraphics[align=c,width=\imwidth]{\impathb{sample-51}} &
	
	\includegraphics[align=c,width=\imwidth]{\impathc{sample-103}} &	         
	\includegraphics[align=c,width=\imwidth]{\impathc{sample-104}} &	        
	\includegraphics[align=c,width=\imwidth]{\impathc{sample-116}} &
	
	\includegraphics[align=c,width=\imwidth]{\impathd{sample-47}} &	         
	\includegraphics[align=c,width=\imwidth]{\impathd{sample-49}} &	        
	\includegraphics[align=c,width=\imwidth]{\impathd{sample-53}} &
	
	\includegraphics[align=c,width=\imwidth]{\impathe{sample-287}} &	         
	\includegraphics[align=c,width=\imwidth]{\impathe{sample-71}} &	        
	\includegraphics[align=c,width=\imwidth]{\impathe{sample-128}} 		\\
\bottomrule
\end{tabular}
}
\caption{\label{fig:samples_mix} Samples from \emph{LDMs} trained on CelebAHQ~\cite{DBLP:journals/corr/abs-1710-10196}, FFHQ~\cite{stylegan}, LSUN-Churches~\cite{DBLP:journals/corr/YuZSSX15}, LSUN-Bedrooms~\cite{DBLP:journals/corr/YuZSSX15} and class-conditional ImageNet~\cite{DBLP:conf/cvpr/DengDSLL009}, each with a resolution of $256 \times 256$. Best viewed when zoomed in. For more samples \cf the supplement. \vspace{-1.5em}}
\end{figure*}

}

\newcommand{\srimagenet}{
\begin{figure}[tbp]
	\setlength{\tabcolsep}{1.25pt}		
\begin{small}
\begin{tabular}{ccc}
bicubic & \emph{LDM}-SR & SR3 \\
\includegraphics[width = 1.04in]{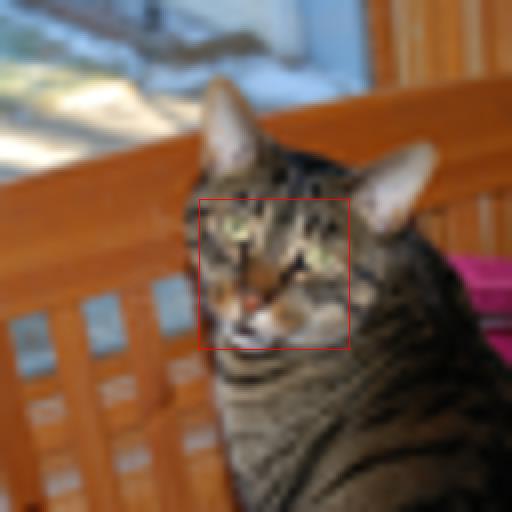}&
\includegraphics[width = 1.04in]{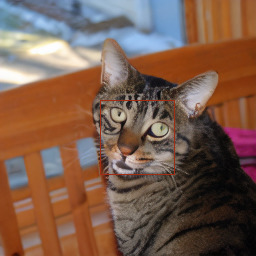}&
\includegraphics[width = 1.04in]{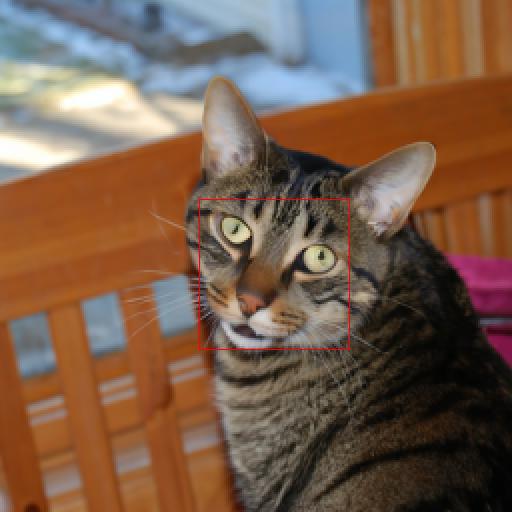}\\
\includegraphics[width = 1.04in]{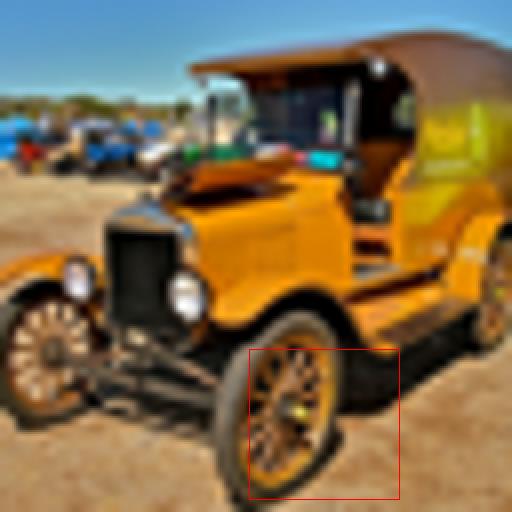}&
\includegraphics[width = 1.04in]{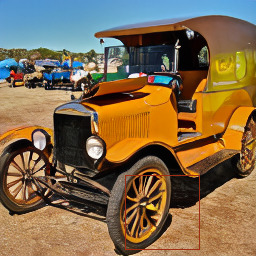}& \includegraphics[width = 1.04in]{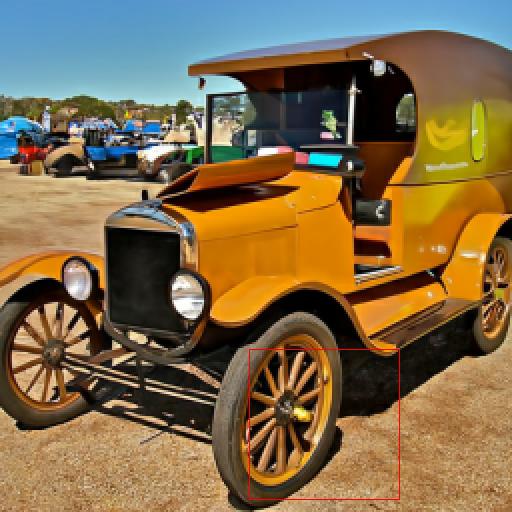}\\
\end{tabular}
\end{small}
\vspace*{-1.0em}
\caption{\label{fig:srimagenet}ImageNet 64$\rightarrow$256 super-resolution on ImageNet-Val. 
\emph{LDM-SR} has advantages at rendering realistic textures but 
SR3 can synthesize more coherent fine structures.
See appendix for additional samples and cropouts. SR3 results from \cite{DBLP:journals/corr/abs-2104-07636}.
\vspace{-1em}}
\end{figure}
}

\newcommand{\srgeneralization}{
\begin{figure*}[htbp]
\addtolength{\tabcolsep}{-4.4pt}    
\begin{tabular}{ccc}
bicubic & \emph{LDM-SR} & \emph{LDM-BSR} \\
\includegraphics[trim=32 128 16 32, clip, width = 2.25in]{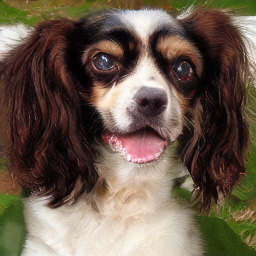}&
\includegraphics[trim=128 512 64 128, clip, width = 2.25in]{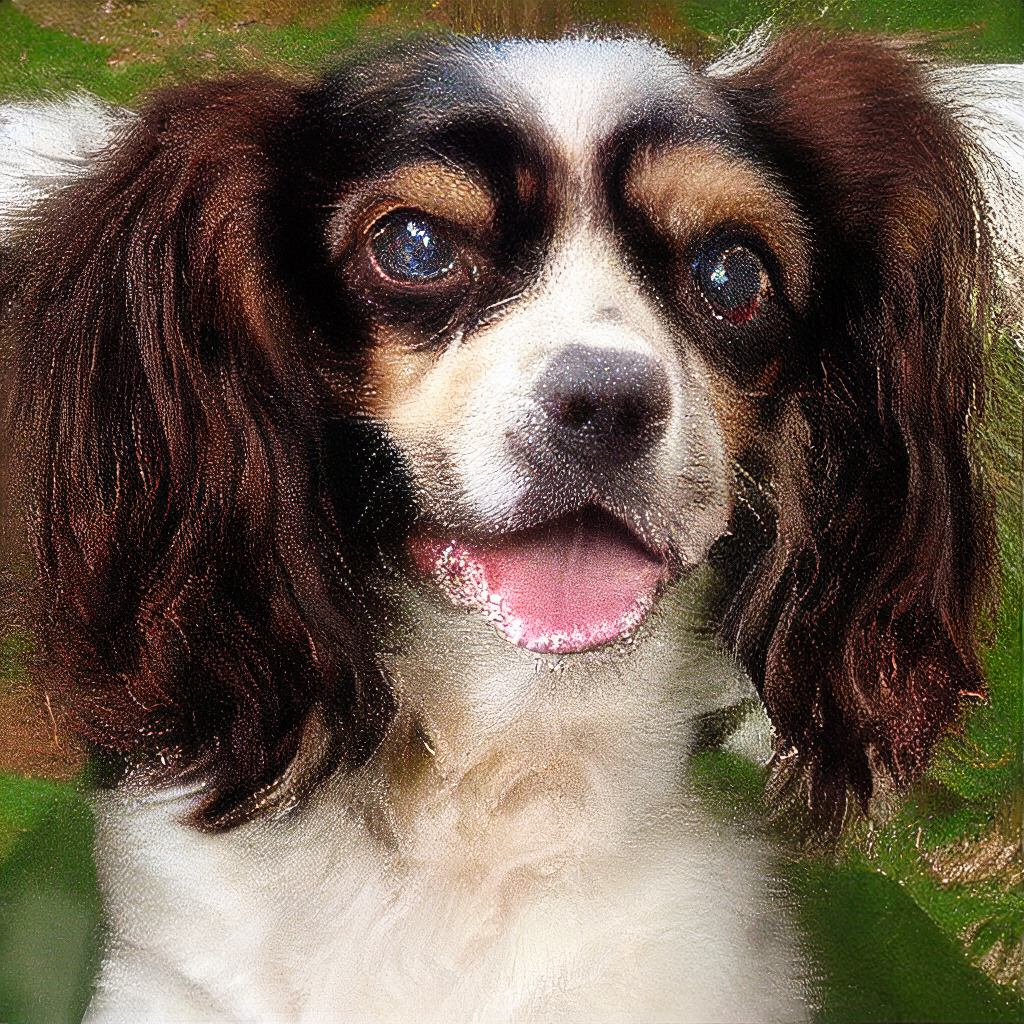}&
\includegraphics[trim=128 512 64 128, clip, width = 2.25in]{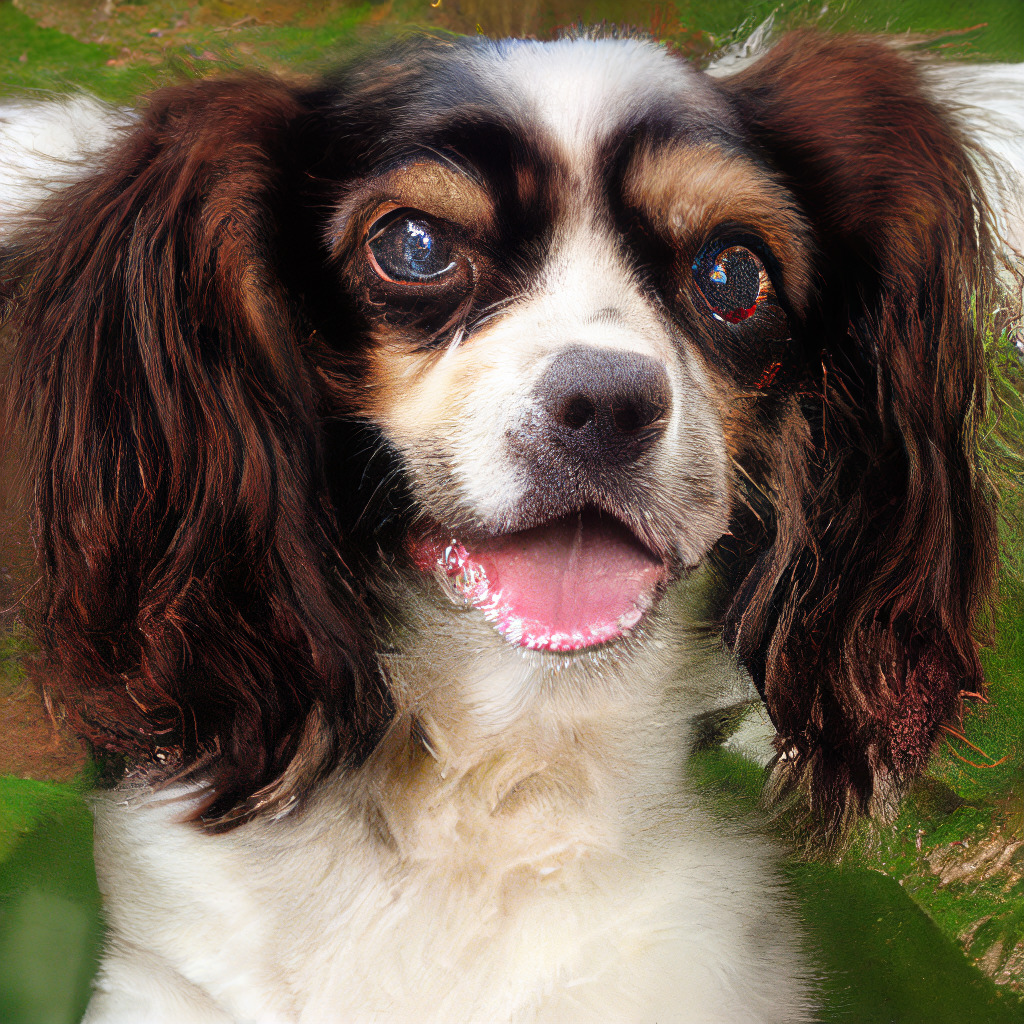}\\
\includegraphics[trim=32 128 16 32, clip, width = 2.25in]{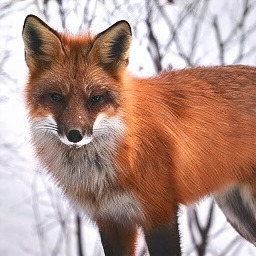}&
\includegraphics[trim=128 512 64 128, clip, width = 2.25in]{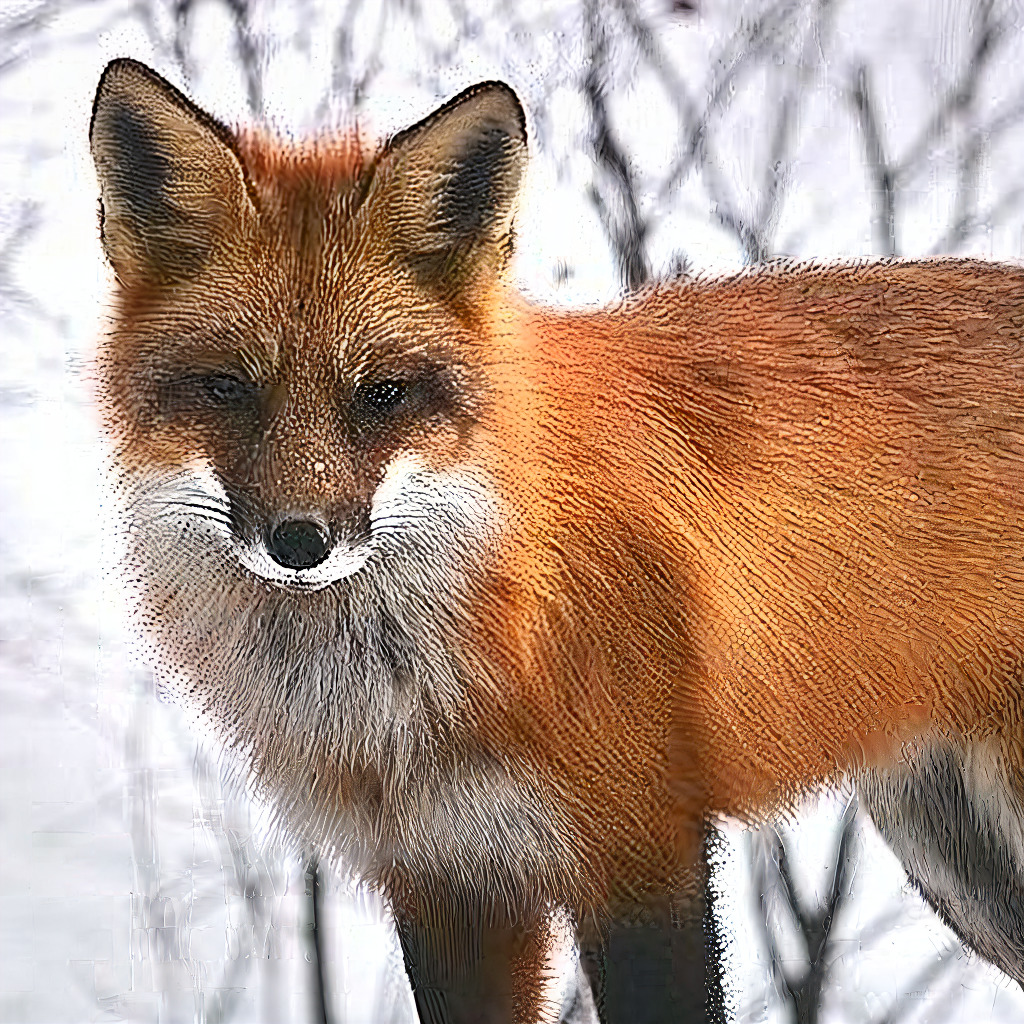}&
\includegraphics[trim=128 512 64 128, clip, width = 2.25in]{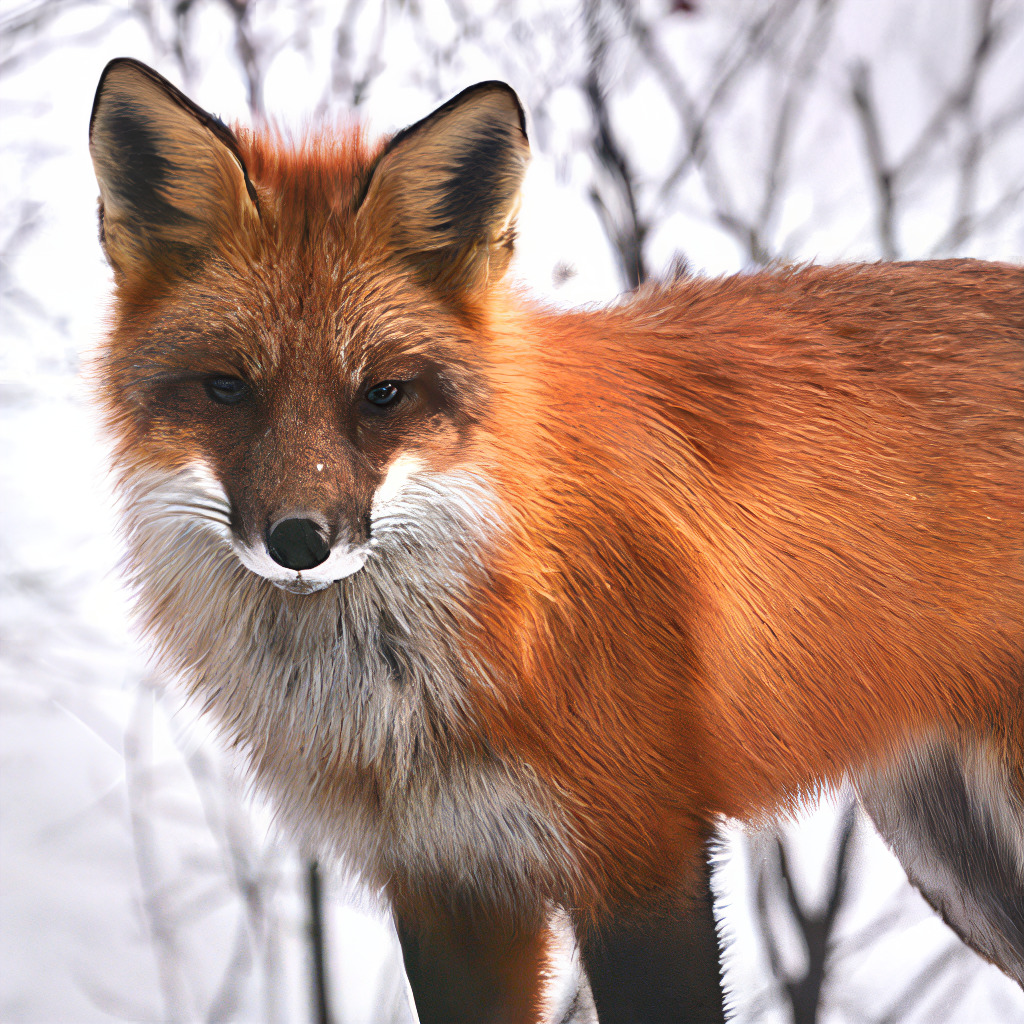}\\
\end{tabular}
\addtolength{\tabcolsep}{4.4pt}
\vspace*{-1em}
\caption{\label{fig:srgeneralization} 
\emph{LDM-BSR} generalizes to arbitrary inputs and can be used as a general-purpose upsampler,
upscaling samples from a class-conditional \emph{LDM} (image \cf Fig.~\ref{fig:samples_mix}) to $1024^2$ resolution. In contrast, 
using a fixed degradation process (see Sec.~\ref{subsec:superres}) hinders generalization.\vspace{-1em}}
\end{figure*}
}

\newcommand{\inpaintingremoval}{
\begin{figure}[!bhtp]
\centering
\scriptsize
	\renewcommand{\imwidth}{0.23\textwidth}
    \renewcommand{\impatha}[1]{img/object_removal/##1} 
	\setlength{\tabcolsep}{0pt}
	\begin{tabular}{c@{\hskip 2pt}c}
	\toprule
    input & result \\
	\midrule
    \includegraphics[align=c,width=\imwidth]{\impatha{input/000007.jpg}} &
   \includegraphics[align=c,width=\imwidth]{\impatha{our_01/000007.jpg}} \\
  \rule{0pt}{6ex}%

    \includegraphics[align=c,width=\imwidth]{\impatha{input/000012.jpg}} &
   \includegraphics[align=c,width=\imwidth]{\impatha{our_00/000012.jpg}} \\
  \rule{0pt}{6ex}%

    \includegraphics[align=c,width=\imwidth]{\impatha{input/000013.jpg}} &
   \includegraphics[align=c,width=\imwidth]{\impatha{our_00/000013.jpg}} \\

	\bottomrule
  \end{tabular}\vspace{-1em}
\caption{\label{inpaintingremoval} Qualitative results on object removal with
  our \emph{big, w/ ft} inpainting model. For more results, see
  Fig.~\ref{suppinpaintingremoval}. \vspace{-2.0em}}
\end{figure}
}

\newcommand{\texttoimgquantcocotwo}{
  \begin{center}
\begin{table}[thbp]
\begin{footnotesize}
\begin{adjustbox}{max width=\linewidth}
  \begin{tabular}[t]{lcccc}
    \toprule
    \multicolumn{5}{c}{\textbf{Text-Conditional Image Synthesis}} \\
    \midrule
    \textbf{Method} & FID $\downarrow$ & IS$\uparrow$ & $N_{\text{params}}$ & \\
    \midrule
	CogView$^{\dagger}$~\cite{DBLP:journals/corr/abs-2105-13290} & 27.10 & 18.20 & 4B & self-ranking, rejection rate 0.017\\
	LAFITE$^{\dagger}$~\cite{DBLP:journals/corr/abs-2111-13792} & 26.94 & \underline{26.02} & 75M & \\    
    GLIDE$^*$~\cite{DBLP:journals/corr/abs-2112-10741} & \underline{12.24} & - & 6B & 277 DDIM steps, c.f.g.~\cite{ho2021classifier} $s=3$ \\
	Make-A-Scene$^*$~\cite{DBLP:journals/corr/abs-2203-13131} & \textbf{11.84} & - & 4B & c.f.g for AR models~\cite{ar_cfg} $s=5$\\    
	\midrule    
    \emph{LDM-KL-8}& 23.31 & 20.03\tiny$\pm\text{0.33}$ & 1.45B & 250 DDIM steps\\
    \emph{LDM-KL-8-G}$^*$& 12.63 & \textbf{30.29\tiny$\pm\text{0.42}$}	 & 1.45B & 250 DDIM steps, c.f.g.~\cite{ho2021classifier} $s=1.5$\\    
  \bottomrule
\end{tabular}
\end{adjustbox}\vspace{-0.65em}
\end{footnotesize}
 \caption{\label{tab:txt2img} Evaluation of text-conditional image synthesis on the
  $256 \times 256$-sized MS-COCO~\cite{DBLP:journals/corr/LinMBHPRDZ14} dataset: with 250 DDIM~\cite{DBLP:conf/iclr/SongME21} steps our model 
  is on par with the most recent diffusion~\cite{DBLP:journals/corr/abs-2112-10741} and autoregressive~\cite{DBLP:journals/corr/abs-2203-13131} methods despite using significantly less parameters. $^\dagger$/$^*$:Numbers from~\cite{DBLP:journals/corr/abs-2111-13792}/~\cite{DBLP:journals/corr/abs-2203-13131} \vspace{-1em}}
\end{table}
\end{center}
}

\newcommand{\fidsnew}{
\begin{table}[t]
\begin{footnotesize}
  \begin{adjustbox}{max width=\linewidth}
  \begin{tabular}{ccccccccc}
    \toprule
    \multicolumn{4}{c}{CelebA-HQ $256\times 256$} & & \multicolumn{4}{c}{FFHQ $256\times 256$} \\
    \cmidrule{1-4} \cmidrule{6-9}
    \textbf{Method} & FID $\downarrow$ & Prec. $\uparrow$ & Recall $\uparrow$ & & \textbf{Method} & FID $\downarrow$ & Prec. $\uparrow$ & Recall $\uparrow$  \\
    \cmidrule{1-4} \cmidrule{6-9}
     DC-VAE \cite{DBLP:conf/cvpr/ParmarLLT21} & 15.8 & -&- & & ImageBART \cite{DBLP:journals/corr/abs-2108-08827}  & 9.57 & - & - \\
    VQGAN+T. \cite{DBLP:journals/corr/abs-2012-09841} (k=400) & 10.2 & - & -& & U-Net GAN (+aug) \cite{DBLP:conf/cvpr/SchonfeldSK20} & 10.9 (7.6) & - & - \\
    PGGAN \cite{DBLP:journals/corr/abs-1710-10196} & 8.0 &- &- & & UDM \cite{DBLP:journals/corr/abs-2106-05527}  & 5.54 & - & - \\
    LSGM \cite{DBLP:journals/corr/abs-2106-05931} & 7.22 & -&- & & StyleGAN \cite{stylegan} & \underline{4.16} & \underline{0.71} & \underline{0.46} \\
     UDM \cite{DBLP:journals/corr/abs-2106-05527} & \underline{7.16} & -&- & & ProjectedGAN\cite{DBLP:journals/corr/abs-2111-01007} & \textbf{3.08} & 0.65 & \underline{0.46} \\
    \midrule
    \emph{LDM-4} (ours, 500-s$^\dagger$) & \textbf{5.11} & 0.72 & 0.49 & &  \emph{LDM-4} (ours, 200-s) & 4.98 & \textbf{0.73} & \textbf{0.50} \\
\bottomrule
    \end{tabular}
    \end{adjustbox}
    \vspace{2em}
  \begin{adjustbox}{max width=\linewidth}
  \begin{tabular}{c c c c c c c c c}
    \toprule
    \multicolumn{4}{c}{LSUN-Churches $256\times 256$} & & \multicolumn{4}{c}{LSUN-Bedrooms $256\times 256$} \\
    \cmidrule{1-4} \cmidrule{6-9}
    \textbf{Method} & FID $\downarrow$& Prec. $\uparrow$ & Recall $\uparrow$  & & \textbf{Method} & FID $\downarrow$& Prec. $\uparrow$ & Recall $\uparrow$  \\
    \cmidrule{1-4} \cmidrule{6-9}
	DDPM \cite{DBLP:conf/nips/HoJA20} & 7.89 & - & - & &ImageBART \cite{DBLP:journals/corr/abs-2108-08827} & 5.51 & -&- \\
    ImageBART\cite{DBLP:journals/corr/abs-2108-08827} & 7.32 & - & - & &  DDPM \cite{DBLP:conf/nips/HoJA20} &4.9 &- &- \\
    PGGAN \cite{DBLP:journals/corr/abs-1710-10196} & 6.42 & - & - & & UDM \cite{DBLP:journals/corr/abs-2106-05527} & 4.57 & -&-\\
    StyleGAN\cite{stylegan} & 4.21 & - & - & &  StyleGAN\cite{stylegan} & 2.35 & 0.59 &  \underline{0.48} \\
    StyleGAN2\cite{DBLP:journals/corr/abs-1912-04958} & \underline{3.86} & - & - &  & ADM \cite{DBLP:journals/corr/abs-2105-05233} & \underline{1.90} & \textbf{0.66} & \textbf{0.51} \\
    ProjectedGAN\cite{DBLP:journals/corr/abs-2111-01007} & \textbf{1.59} & \underline{0.61} & \underline{0.44} & & ProjectedGAN\cite{DBLP:journals/corr/abs-2111-01007} & \textbf{1.52} & \underline{0.61} & 0.34\\
    \midrule
    \emph{LDM-8}$^*$ (ours, 200-s) & 4.02 & \textbf{0.64} & \textbf{0.52} & & \emph{LDM-4} (ours, 200-s) & 2.95 & \textbf{0.66} & \underline{0.48} \\
    \bottomrule
  \end{tabular}
  \end{adjustbox}\vspace{-2.6em}
\end{footnotesize}
  \caption{\label{tab:fids} Evaluation metrics for unconditional image
  synthesis. CelebA-HQ results reproduced from
  \cite{DBLP:conf/cvpr/ParmarLLT21,DBLP:conf/iclr/XiaoKKV21,DBLP:journals/corr/abs-2106-05527},
  FFHQ from
  \cite{DBLP:journals/corr/abs-1912-04958,DBLP:journals/corr/abs-2106-05527}. $^\dagger$: $N$-s refers to $N$ sampling steps with the DDIM~\cite{DBLP:conf/iclr/SongME21} sampler. $^*$: trained in \emph{KL}-regularized latent space.
  Additional results can be found in the supplementary.\vspace{-1.6em}}
\end{table}
}

\newcommand{\srtable}{
  \begin{table}[!bpth]\vspace{-1em}
\begin{footnotesize}
  \begin{adjustbox}{max width=\linewidth}
  \begin{tabular}{l c c c c c c}
  	\toprule
  	\textbf{Method} & FID $\downarrow$ & IS $\uparrow$ & PSNR  $\uparrow$  & SSIM  $\uparrow$  & $N_{\text{params}}$& \tiny$[\frac{\text{samples}}{s}] (^*)$ \\
  	\midrule
  	Image Regression \cite{DBLP:journals/corr/abs-2104-07636} & 15.2 & 121.1 & \textbf{27.9} & \textbf{0.801} & 625M & N/A \\
  	SR3 \cite{DBLP:journals/corr/abs-2104-07636} & 5.2 & \textbf{180.1} & \underline{26.4} & \underline{0.762} & 625M & N/A \\
  	\midrule
  	\emph{LDM-4} (ours, 100 steps) & \underline{2.8}$^\dagger$/\underline{4.8}$^\ddagger$ & 166.3 & 24.4\tiny{$\pm $3.8} & 0.69\tiny{$\pm $0.14} & \textbf{169M} & 4.62 \\
  	emph{LDM-4} (ours, big, 100 steps) & \textbf{2.4}$^\dagger$/\textbf{4.3}$^\ddagger$ & \underline{174.9} & 24.7\tiny{$\pm $4.1} & 0.71\tiny{$\pm $0.15} & 552M & 4.5 \\
  	\emph{LDM-4} (ours, 50 steps, guiding) & 4.4$^\dagger$/6.4$^\ddagger$ & 153.7 & 25.8\tiny{$\pm $3.7} & 0.74\tiny{$\pm $0.12} & 	\underline{184M}  & 0.38 \\
  	\bottomrule

  \end{tabular}
  \end{adjustbox}
\end{footnotesize}\vspace{-0.85em}
  \caption{\label{tab:srtable}  $\times 4$ upscaling results on ImageNet-Val. ($256^2$); $^\dagger$: FID features computed on validation split, $^\ddagger$: FID features computed on train split; $^*$: Assessed on a NVIDIA A100\vspace{-1em}}
\end{table}
}

\newcommand{\inpaintingtable}{
\begin{table}\vspace{-1.5em}
\begin{center}
\begin{footnotesize}
 \begin{adjustbox}{max width=\linewidth}
  \begin{tabular}[t]{lrrrr}
   \toprule
    & \multicolumn{2}{c}{\textbf{40-50\% masked}}& \multicolumn{2}{c}{\textbf{All samples}} \\
    \cmidrule(lr){2-3} \cmidrule(lr){4-5}
    \textbf{Method} & FID $\downarrow$ & LPIPS $\downarrow$ & FID $\downarrow$ & LPIPS $\downarrow$\\
    \midrule
    \emph{LDM-4} (ours, big, w/ ft)&\textbf{9.39}&\underline{0.246}\tiny{$\pm$ 0.042}&\textbf{1.50}&\underline{0.137}\tiny{$\pm$ 0.080}\\
    \emph{LDM-4} (ours, big, w/o ft)&12.89&0.257\tiny{$\pm$ 0.047}&2.40&\underline{0.142}\tiny{$\pm$ 0.085}\\
    \emph{LDM-4} (ours, w/ attn)&11.87&0.257\tiny{$\pm$ 0.042}&2.15&\underline{0.144}\tiny{$\pm$ 0.084}\\
    \emph{LDM-4} (ours, w/o attn)&12.60&0.259\tiny{$\pm$ 0.041}&2.37&\underline{0.145}\tiny{$\pm$ 0.084}\\
    \midrule
    LaMa\cite{lama}$^\dagger$&12.31&\textbf{0.243}\tiny{$\pm$ 0.038}&2.23&\textbf{0.134}\tiny{$\pm$ 0.080}\\
    LaMa\cite{lama}&12.0\phantom{0}&\textbf{0.24}\phantom{\tiny{$\pm$ 0.000}}&2.21&\underline{0.14}\phantom{\tiny{$\pm$ 0.000}}\\
    CoModGAN\cite{comodgan}&\underline{10.4}\phantom{0}&0.26\phantom{\tiny{$\pm$ 0.000}}&\underline{1.82}&0.15\phantom{\tiny{$\pm$ 0.000}}\\
    RegionWise\cite{regionwise}&21.3\phantom{0}&0.27\phantom{\tiny{$\pm$ 0.000}}&4.75&0.15\phantom{\tiny{$\pm$ 0.000}}\\
    DeepFill v2\cite{deepfillv2}&22.1\phantom{0}&0.28\phantom{\tiny{$\pm$ 0.000}}&5.20&0.16\phantom{\tiny{$\pm$ 0.000}}\\
    EdgeConnect\cite{edgeconnect}&30.5\phantom{0}&0.28\phantom{\tiny{$\pm$ 0.000}}&8.37&0.16\phantom{\tiny{$\pm$ 0.000}}\\
    \bottomrule

\end{tabular}
 \end{adjustbox}\vspace{-0.85em}
\end{footnotesize}
 \caption{\label{inpaintingtable}Comparison of inpainting performance on 30k
  crops of size $512\times 512$ from test images of Places\cite{places}. The column
  \emph{40-50\%} reports  metrics computed over hard examples where 40-50\% of
  the image region have to be inpainted. $^\dagger$recomputed on our test set,
  since the original test set used in \cite{lama} was not available.}
\end{center}\vspace{-2em}
\end{table}
}

\newcommand{\inpaintingefficiency}{%
  \begin{table}\vspace{-0.5em}
\begin{center}
 \begin{adjustbox}{max width=\linewidth}
  \begin{tabular}[t]{lcrrcc}
    \toprule
     & train throughput & \multicolumn{2}{c}{sampling throughput$^{\dagger}$} & train+val & FID@2k \\
    \textbf{Model} (\emph{reg.}-type) & samples/sec.& @256 & @512 & hours/epoch & epoch 6 \\
    \midrule
	\emph{LDM-1} (no first stage) &0.11&0.26 &0.07 & 20.66&24.74\\
    \emph{LDM-4} (\emph{KL}, w/ attn) &0.32&0.97 &0.34 & 7.66&15.21\\
    \emph{LDM-4} (\emph{VQ}, w/ attn) &0.33& 0.97& 0.34& 7.04&14.99 \\
    \emph{LDM-4} (\emph{VQ}, w/o attn) &0.35& 0.99& 0.36& 6.66&15.95\\
    \bottomrule
\end{tabular}
 \end{adjustbox}\vspace{-0.85em}
 \caption{\label{inpaintingefficiency}
  Assessing inpainting efficiency. $^{\dagger}$: Deviations from Fig.~\ref{fig:speedplot} due to varying GPU settings/batch sizes \cf the supplement.}
\end{center}\vspace{-2.5em}
\end{table}
}

\newcommand{\cinmainmetrics}{
\begin{table}[htbp]
\centering
\begin{footnotesize}
\begin{adjustbox}{max width=\linewidth}
\footnotesize
\begin{tabular}{lcccccc}
\toprule
\textbf{Method} & FID$\downarrow$ & IS$\uparrow$ & Precision$\uparrow$ & Recall$\uparrow$ & $N_{\text{params}}$ & \\
\midrule
BigGan-deep~\cite{bigganbrock}& 6.95 & \underline{203.6\tiny$\pm\text{2.6}$} & \textbf{0.87} & 0.28  & 340M & - \\
ADM~\cite{DBLP:journals/corr/abs-2105-05233} & 10.94 & 100.98& 0.69 & \textbf{0.63}  & 554M & 250 DDIM steps\\
ADM-G~\cite{DBLP:journals/corr/abs-2105-05233} & \underline{4.59}  & 186.7& \underline{0.82} & 0.52 & 608M & 250 DDIM steps \\
\midrule
\emph{LDM-4} (ours) & 10.56 & 103.49\tiny$\pm\text{1.24}$  & 0.71  & \underline{0.62} & 400M & 250 DDIM steps\\
\emph{LDM-4}-G (ours) & \textbf{3.60} & \textbf{247.67\tiny$\pm\text{5.59}$}  & \textbf{0.87}  & 0.48 & 400M & 250 steps, c.f.g~\cite{ho2021classifier}, $s=1.5$\\
\bottomrule
\end{tabular}
\end{adjustbox}
\end{footnotesize}
\caption{\label{tab:imagenet_main_numbers} Comparison of a class-conditional ImageNet \emph{LDM} with recent state-of-the-art methods for class-conditional image generation on ImageNet~\cite{DBLP:conf/cvpr/DengDSLL009}. A more detailed comparison with additional baselines can be found in \ref{suppsec:cin}, Tab.~\ref{tab:imagenet_numbers} and \ref{suppsec:compute2}. \emph{c.f.g.} denotes classifier-free guidance with a scale $s$ as proposed in ~\cite{ho2021classifier}.}
\end{table}
}

\newcommand{\userstudy}{
\begin{table}[htbp]
\begin{footnotesize}
\begin{adjustbox}{max width=\linewidth}
  \begin{tabular}{lccccc}
    \toprule
    & \multicolumn{2}{c}{SR on ImageNet}& & \multicolumn{2}{c}{Inpainting on Places} \\
    \cmidrule{2-3} \cmidrule{5-6}
    \textbf{User Study} & Pixel-DM ($f1$) & \emph{LDM-4} & & LAMA~\cite{lama} & \emph{LDM-4} \\
    \cmidrule{2-3} \cmidrule{5-6}
   \textbf{Task 1:} Preference vs GT $\uparrow$ & 16.0\% & \textbf{30.4\%} & & 13.6\% & \textbf{21.0\%} \\ 
   \textbf{Task 2:} Preference Score $\uparrow$ & 29.4\% & \textbf{70.6\%} & & 31.9\% & \textbf{68.1\%} \\ 
\bottomrule
    \end{tabular}
    \end{adjustbox}
    \vspace{-1em}
\end{footnotesize}
\caption{\label{tab:user_study}Task 1: Subjects were shown ground truth and generated image and asked for preference. %
Task 2: Subjects had to decide between two generated images. More details in \ref{suppsubsubsec:user_study}     \vspace{-1em}
}
\end{table}
}

\newcommand{\projectpath}{https://github.com/CompVis/latent-diffusion}

\newcommand{\expec}{\mathbb{E}}

\newcommand{\hpixel}{H}
\newcommand{\wpixel}{W}
\newcommand{\cpixel}{3}
\newcommand{\xrec}{\tilde{x}}

\newcommand{\hlatent}{h}
\newcommand{\wlatent}{w}
\newcommand{\clatent}{c}

\newcommand{\zt}[1]{z_{#1}}

\newcommand{\R}{\mathbb{R}}

\newcommand{\KL}{\mathbb{KL}}
\newcommand{\expect}{\mathbb{E}}

\newcommand{\disc}{D_{\psi}}

\newcommand{\lrec}{L_{rec}}
\newcommand{\ladv}{L_{adv}}
\newcommand{\lreg}{L_{reg}}

\newcommand{\lsimple}{L_{DM}}
\newcommand{\lsimpleldm}{L_{LDM}}
\newcommand{\lsimplelcm}{L_{LDM}}

\newcommand{\model}{\epsilon_\theta}
\newcommand{\conditioner}{\tau_\theta}

\newcommand{\encoder}{\mathcal{E}}
\newcommand{\decoder}{\mathcal{D}}

\newcommand{\cond}{y}

\providecommand{\imwidth}{}

\providecommand{\impath}[1]{}
\providecommand{\impaths}[1]{}
\providecommand{\impatha}[1]{}
\providecommand{\impathb}[1]{}
\providecommand{\impathc}[1]{}
\providecommand{\impathd}[1]{}
\providecommand{\impathe}[1]{}

\newcommand{\goodlandscapes}{

\begin{figure}[htbp]
	\centering
    \renewcommand{\imwidth}{.8\textwidth}
	\renewcommand{\impath}[1]{img/supplement/landscapes/##1}
	\begin{tabular}{c}
  	\toprule
 Semantic Synthesis on Flickr-Landscapes~\cite{DBLP:journals/corr/abs-2012-09841} \\
  	\toprule
	\includegraphics[width=\imwidth]{\impath{cond-48}} \\
	\includegraphics[width=\imwidth]{\impath{sample-51}} \\
	\includegraphics[width=\imwidth]{\impath{sample-45}} \\
	\includegraphics[width=\imwidth]{\impath{sample-47}} \\
	\end{tabular}
	
\caption{\label{fig:landscapes1} When provided a semantic map as conditioning, our \emph{LDMs} generalize to substantially larger resolutions than those seen during training. Although this model was trained on inputs of size $256^{2}$ it can be used to create high-resolution samples as the ones shown here, which are of resolution $1024 \times 384$.}	
\end{figure}

}

\newcommand{\texttoimgconvsamples}{
\begin{figure}[htbp]
	\resizebox{.75\totalheight}{!}{
	\centering
	\renewcommand{\impath}[1]{img/cr/text2img/convolutional/##1}
	\setlength{\tabcolsep}{2pt} 
		\begin{tabular}{cc}
		\toprule
		\multicolumn{2}{c}{\footnotesize{\emph{'A painting of the last supper by Picasso.'}}}\\
		\toprule
		\multicolumn{2}{c}{\includegraphics[width=\textwidth]{\impath{a_painting_of_the_last_supper_by_picasso}}}\\
		\midrule
		\footnotesize{\emph{'An oil painting of a latent space.'}}& \shortstack{\footnotesize{\emph{'An epic painting of Gandalf the Black}} \\ \footnotesize{\emph{summoning thunder and lightning in the mountains.'}}} \\
		\cmidrule{1-1} \cmidrule{2-2}
		\includegraphics[height=.285\textheight]{\impath{anoilpaintingoflatentspace}} & 
		\includegraphics[height=.285\textheight]{\impath{an_epic_painting_of_gandalf_the_black_summoning_thunder_and_lightning_in_the_mountains}} \\
		\midrule
		\multicolumn{2}{c}{\footnotesize{\emph{'A sunset over a mountain range, vector image.'}}}\\
		\toprule
		\multicolumn{2}{c}{\includegraphics[width=\textwidth]{\impath{sunsetmountainvector}}}\\
		\bottomrule
		\end{tabular}
		}
		\caption{\label{fig:text2img_conv} Combining classifier free diffusion guidance with the convolutional sampling strategy from Sec.~\ref{subsubsec:beyond}, our 1.45B parameter text-to-image model can be used for rendering images larger than the native $256^2$ resolution the model was trained on.}
\end{figure}
}

\newcommand{\landscapestune}{
\begin{figure}[!h]
\centering
    \renewcommand{\imwidth}{.835\textwidth}
	\renewcommand{\impath}[1]{img/highlandscapes/512-ft/##1}

\includegraphics[width=\imwidth]{\impath{10095_s1}}
\includegraphics[width=\imwidth]{\impath{1107_s1_cropped}}
\includegraphics[width=\imwidth]{\impath{5262_s2}}
\caption{\label{fig:landscapestune} Convolutional samples from the semantic landscapes model as in Sec.~\ref{subsubsec:beyond}, finetuned on $512^2$ images.}
\end{figure}
}

\newcommand{\landscapestunewithmap}{
\begin{figure}[htbp]
	\centering
    \renewcommand{\imwidth}{.725\textwidth}
	\renewcommand{\impath}[1]{img/highlandscapes/512-ft/##1}
	\begin{tabular}{c}
  	\toprule
	 Semantic Synthesis on Flickr-Landscapes~\cite{DBLP:journals/corr/abs-2012-09841} ($512^2$ finetuning) \\
  	\toprule
	\includegraphics[trim=0 100 0 0, clip, width=\imwidth]{\impath{5240_cond}} \\
	\includegraphics[trim=0 100 0 0, clip, width=\imwidth]{\impath{5240_s1}} \\
	\includegraphics[trim=0 100 0 0, clip, width=\imwidth]{\impath{5240_s4}} \\
	\end{tabular}
	
\caption{\label{fig:landscapestunewithmap} Convolutional samples from the semantic landscapes model as in Sec.~\ref{subsubsec:beyond}, finetuned on $512^2$ images.}	
\end{figure}

}

\newcommand{\thickersample}{
\begin{figure*}[htbp]
\centering
\includegraphics[width=\linewidth]{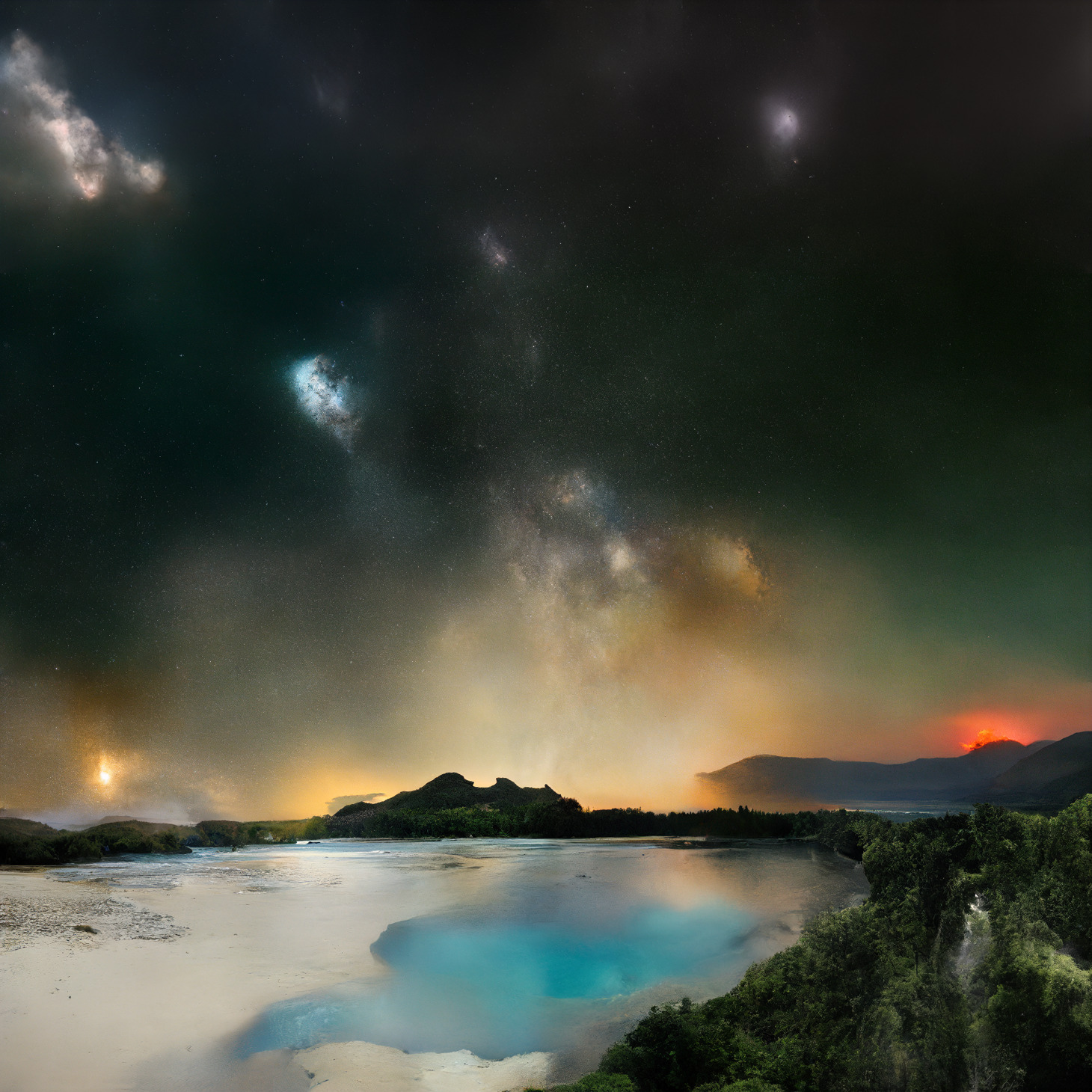}\vspace{-0.65em}  %
  \caption{\label{fig:thickersample} A \emph{LDM} trained on $256^2$ resolution can generalize to 
  larger resolution for spatially conditioned tasks
  such as semantic synthesis of landscape images. See
  Sec.~\ref{subsubsec:beyond}.\vspace{-1em}
  }
\end{figure*}
}

\newcommand{\cinprogressvdays}{
\begin{figure}[htbp]
\begin{subfigure}{.59\textwidth}
\centering
   \includegraphics[width=\linewidth]{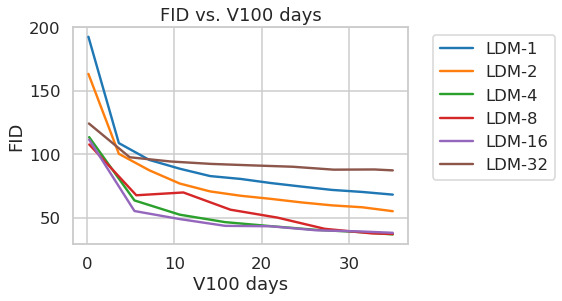}
\end{subfigure}
\hfill
\begin{subfigure}{.39\textwidth}
\centering
\includegraphics[width=\linewidth]{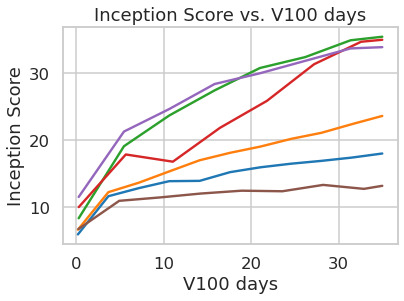}
\end{subfigure}
\caption{\label{fig:cin_traincourse_v100} For completeness we also report the training progress of class-conditional \emph{LDMs} on the ImageNet dataset for a fixed number of 35 V100 days. 
Results obtained with 100 DDIM steps~\cite{DBLP:conf/iclr/SongME21} and $\kappa = 0$. FIDs computed on 5000 samples for efficiency reasons.}
\end{figure}
}

\newcommand{\supercows}{
\begin{figure*}[htbp]
\addtolength{\tabcolsep}{-4.4pt}    
\renewcommand{\impath}[1]{img/supplement/cows/##1}
\centering
\begin{tabular}{cc}
bicubic & \emph{LDM-BSR} \\
\includegraphics[trim=0 0 0 0, clip, width = 0.4\textwidth, height=0.3\textheight]{\impath{cow_540_bicubic}}&
\includegraphics[trim=0 0 0 0, clip, width = 0.4\textwidth, height=0.3\textheight]{\impath{cow_540_hq_1}}\\
\includegraphics[trim=0 0 0 0, clip, width = 0.4\textwidth, height=0.3\textheight]{\impath{cow_999_bicubic}}&
\includegraphics[trim=0 0 0 0, clip, width = 0.4\textwidth, height=0.3\textheight]{\impath{cow_hq_999_1}}\\
\includegraphics[trim=0 0 0 0, clip, width = 0.4\textwidth, height=0.3\textheight]{\impath{cow_10_bicubic}}&
\includegraphics[trim=0 0 0 0, clip, width = 0.4\textwidth, height=0.3\textheight]{\impath{cow_10_hq_3}}\\
\end{tabular}
\addtolength{\tabcolsep}{4.4pt}
\vspace*{-1em}
\caption{\label{fig:supercows} 
\emph{LDM-BSR} generalizes to arbitrary inputs and can be used as a general-purpose upsampler,
upscaling samples from the LSUN-Cows dataset to $1024^2$ resolution.}
\end{figure*}
}

\newcommand{\convolutionalrescaling}{
\begin{figure*}[htbp]
\centering
\addtolength{\tabcolsep}{-4.4pt}    
\renewcommand{\impath}[1]{img/supplement/rescalingeffects/##1}
\begin{tabular}{ccc}
\toprule
KL-reg, w/o rescaling & KL-reg, w/ rescaling & VQ-reg, w/o rescaling \\
\toprule
\includegraphics[trim=0 0 0 0, clip, width = 0.34\textwidth]{\impath{7_norevae_2}}&
\includegraphics[trim=0 0 0 0, clip, width = 0.34\textwidth]{\impath{7_revae_1}}&
\includegraphics[trim=0 0 0 0, clip, width = 0.34\textwidth]{\impath{7_vq_1}}\\
\includegraphics[trim=0 0 0 0, clip, width = 0.34\textwidth]{\impath{1_nore_vae_2}}&
\includegraphics[trim=0 0 0 0, clip, width = 0.34\textwidth]{\impath{1_re_vae}}&
\includegraphics[trim=0 0 0 0, clip, width = 0.34\textwidth]{\impath{1_vq}}\\
\includegraphics[trim=0 0 0 0, clip, width = 0.34\textwidth]{\impath{22_norevae_3}}&
\includegraphics[trim=0 0 0 0, clip, width = 0.34\textwidth]{\impath{22_revae_3}}&
\includegraphics[trim=0 0 0 0, clip, width = 0.34\textwidth]{\impath{22_vq_1}}\\
\end{tabular}
\addtolength{\tabcolsep}{4.4pt}
\vspace*{-1em}
\caption{\label{fig:convolutionalrescaling} Illustrating the effect of latent space rescaling on convolutional sampling, here for semantic image synthesis on landscapes.
See Sec.~\ref{subsubsec:beyond}~and~Sec.~\ref{suppsec:rescale}.}
\end{figure*}
}

\newcommand{\convolutionalguiding}{
\begin{figure*}[!htbp]
\centering
\addtolength{\tabcolsep}{-4.4pt}    
\renewcommand{\impath}[1]{img/guided_sr/##1}
\begin{tabular}{ccc}
\toprule
Samples $256^2$ & Guided Convolutional Samples $512^2$ & Convolutional Samples $512^2$\\
\toprule
\includegraphics[trim=-80 -80 -80 -80, clip, width = 0.25\textwidth]{\impath{256/4.jpg}}&
\includegraphics[trim=0 0 0 0, clip, width = 0.25\textwidth]{\impath{guided2/1.jpg}}&
\includegraphics[trim=0 0 0 0, clip, width = 0.25\textwidth]{\impath{512/1.jpg}}\\
\includegraphics[trim=-80 -80 -80 -80, clip, width = 0.25\textwidth]{\impath{256/5.jpg}}&
\includegraphics[trim=0 0 0 0, clip, width = 0.25\textwidth]{\impath{guided/1.jpg}}&
\includegraphics[trim=0 0 0 0, clip, width = 0.25\textwidth]{\impath{512/5.jpg}}\\
\includegraphics[trim=-80 -80 -80 -80, clip, width = 0.25\textwidth]{\impath{256/2.jpg}}&
\includegraphics[trim=0 0 0 0, clip, width = 0.25\textwidth]{\impath{guided2/3.jpg}}&
\includegraphics[trim=0 0 0 0, clip, width = 0.25\textwidth]{\impath{512/7.jpg}}\\
\end{tabular}
\addtolength{\tabcolsep}{4.4pt}
\vspace*{-1em}
\caption{\label{fig:convolutionalguiding} On landscapes, convolutional sampling with unconditional models can lead to homogeneous and incoherent global structures (see column 2). $L_2$-guiding with a low resolution image can help to reestablish coherent global structures.}
\end{figure*}
}

\newcommand{\suppsrimagenetbaseline}{
\begin{figure}
\centering
\scriptsize
	\renewcommand{\imwidth}{0.1525\textwidth}
    \renewcommand{\impatha}[1]{img/sr_ho_comparison/Ho_imagenet_64x_256x/##1}
    \renewcommand{\impathb}[1]{img/sr_ho_comparison/sr_imnet_1152256/##1}
        \renewcommand{\impathc}[1]{img/sr_ho_comparison/baseline_1152200/##1}
	\setlength{\tabcolsep}{0pt}
	\begin{tabular}{c@{\hskip 2pt}c@{\hskip 2pt}cc@{\hskip 2pt}cc}
	\toprule
    input & GT & \textbf{\emph{Pixel Baseline} \#1} & \textbf{\emph{Pixel Baseline} \#2} & \textbf{\emph{LDM} \#1} & \textbf{\emph{LDM} \#2} \\
	\midrule
   \includegraphics[align=c,width=\imwidth]{\impatha{bicubic_interp/bullet_train1_bb.jpeg}} &
   \includegraphics[align=c,width=\imwidth]{\impatha{reference/bullet_train1_bb.jpeg}} &
   \includegraphics[align=c,width=\imwidth]{\impathc{22480/1.jpg}} &
   \includegraphics[align=c,width=\imwidth]{\impathc{22480/2.jpg}} &
   \includegraphics[align=c,width=\imwidth]{\impathb{22480/1.jpg}} &
   \includegraphics[align=c,width=\imwidth]{\impathb{22480/2.jpg}} \\
   \rule{0pt}{6ex}%

   \includegraphics[align=c,width=\imwidth]{\impatha{bicubic_interp/candy1_bb.jpeg}} &
   \includegraphics[align=c,width=\imwidth]{\impatha{reference/candy1_bb.jpeg}} &
   \includegraphics[align=c,width=\imwidth]{\impathc{24583/1.jpg}} &
   \includegraphics[align=c,width=\imwidth]{\impathc{24583/2.jpg}} &
   \includegraphics[align=c,width=\imwidth]{\impathb{24583/1.jpg}} &
   \includegraphics[align=c,width=\imwidth]{\impathb{24583/2.jpg}} \\
   \rule{0pt}{6ex}%

   \includegraphics[align=c,width=\imwidth]{\impatha{bicubic_interp/cat1_bb.jpeg}} &
   \includegraphics[align=c,width=\imwidth]{\impatha{reference/cat1_bb.jpeg}} &
   \includegraphics[align=c,width=\imwidth]{\impathc{13923/1.jpg}} &
   \includegraphics[align=c,width=\imwidth]{\impathc{13923/2.jpg}} &
   \includegraphics[align=c,width=\imwidth]{\impathb{13923/1.jpg}} &
   \includegraphics[align=c,width=\imwidth]{\impathb{13923/2.jpg}} \\
   \rule{0pt}{6ex}%

   \includegraphics[align=c,width=\imwidth]{\impatha{bicubic_interp/dock1_bb.jpeg}} &
   \includegraphics[align=c,width=\imwidth]{\impatha{reference/dock1_bb.jpeg}} &
   \includegraphics[align=c,width=\imwidth]{\impathc{25922/1.jpg}} &
   \includegraphics[align=c,width=\imwidth]{\impathc{25922/2.jpg}} &
   \includegraphics[align=c,width=\imwidth]{\impathb{25922/1.jpg}} &
   \includegraphics[align=c,width=\imwidth]{\impathb{25922/2.jpg}} \\
   \rule{0pt}{6ex}%
   
   \includegraphics[align=c,width=\imwidth]{\impatha{bicubic_interp/flowerpot1_bb.jpeg}} &
   \includegraphics[align=c,width=\imwidth]{\impatha{reference/flowerpot1_bb.jpeg}} &
   \includegraphics[align=c,width=\imwidth]{\impathc{35693/1.jpg}} &
   \includegraphics[align=c,width=\imwidth]{\impathc{35693/2.jpg}} &
   \includegraphics[align=c,width=\imwidth]{\impathb{35693/1.jpg}} &
   \includegraphics[align=c,width=\imwidth]{\impathb{35693/2.jpg}} \\
   \rule{0pt}{6ex}%

   \includegraphics[align=c,width=\imwidth]{\impatha{bicubic_interp/greenhouse1_bb.jpeg}} &
   \includegraphics[align=c,width=\imwidth]{\impatha{reference/greenhouse1_bb.jpeg}} &
   \includegraphics[align=c,width=\imwidth]{\impathc{27998/1.jpg}} &
   \includegraphics[align=c,width=\imwidth]{\impathb{27998/2.jpg}} &
   \includegraphics[align=c,width=\imwidth]{\impathb{27998/1.jpg}} &
   \includegraphics[align=c,width=\imwidth]{\impathb{27998/2.jpg}} \\
   \rule{0pt}{6ex}%
   
   \includegraphics[align=c,width=\imwidth]{\impatha{bicubic_interp/jaguar_bb.jpeg}} &
   \includegraphics[align=c,width=\imwidth]{\impatha{reference/jaguar_bb.jpeg}} &
   \includegraphics[align=c,width=\imwidth]{\impathc{14234/1.jpg}} &
   \includegraphics[align=c,width=\imwidth]{\impathc{14234/2.jpg}} &
   \includegraphics[align=c,width=\imwidth]{\impathb{14234/1.jpg}} &
   \includegraphics[align=c,width=\imwidth]{\impathb{14234/2.jpg}} \\
   \rule{0pt}{6ex}%

   \includegraphics[align=c,width=\imwidth]{\impatha{bicubic_interp/model_t_bb.jpeg}} &
   \includegraphics[align=c,width=\imwidth]{\impatha{reference/model_t_bb.jpeg}} &
   \includegraphics[align=c,width=\imwidth]{\impathc{31979/1.jpg}} &
   \includegraphics[align=c,width=\imwidth]{\impathc{31979/2.jpg}} &
   \includegraphics[align=c,width=\imwidth]{\impathb{31979/1.jpg}} &
   \includegraphics[align=c,width=\imwidth]{\impathb{31979/2.jpg}} \\	
	\bottomrule
  \end{tabular}
\caption{\label{suppsrimagenet} Qualitative superresolution comparison of two random samples between LDM-SR and baseline-diffusionmodel in Pixelspace. Evaluated on imagenet validation-set after same amount of training steps.}
\end{figure}
}

\newcommand{\suppinpaintingsamples}{
\begin{figure}
\centering
\scriptsize
	\renewcommand{\imwidth}{0.1525\textwidth}
    \renewcommand{\impatha}[1]{img/inpainting_results/##1} 
	\setlength{\tabcolsep}{0pt}
	\begin{tabular}{c@{\hskip 2pt}c@{\hskip 2pt}c@{\hskip 2pt}ccc}
	\toprule
    input & GT & LaMa\cite{lama} & \textbf{\emph{LDM} \#1} & \textbf{\emph{LDM} \#2} & \textbf{\emph{LDM} \#3} \\
	\midrule
   \includegraphics[align=c,width=\imwidth]{\impatha{input/Places365_val_00000919_crop000_mask000.jpg}} &
	    \includegraphics[align=c,width=\imwidth]{\impatha{gt/Places365_val_00000919_crop000_mask000.jpg}} &
	  \includegraphics[align=c,width=\imwidth]{\impatha{lama/Places365_val_00000919_crop000_mask000.jpg}} &
	\includegraphics[align=c,width=\imwidth]{\impatha{our_00/Places365_val_00000919_crop000_mask000.jpg}} &
	\includegraphics[align=c,width=\imwidth]{\impatha{our_09/Places365_val_00000919_crop000_mask000.jpg}} &
	\includegraphics[align=c,width=\imwidth]{\impatha{our_02/Places365_val_00000919_crop000_mask000.jpg}} \\
  \rule{0pt}{6ex}%

   \includegraphics[align=c,width=\imwidth]{\impatha{input/Places365_val_00000532_crop000_mask000.jpg}} &
	    \includegraphics[align=c,width=\imwidth]{\impatha{gt/Places365_val_00000532_crop000_mask000.jpg}} &
	  \includegraphics[align=c,width=\imwidth]{\impatha{lama/Places365_val_00000532_crop000_mask000.jpg}} &
	\includegraphics[align=c,width=\imwidth]{\impatha{our_00/Places365_val_00000532_crop000_mask000.jpg}} &
	\includegraphics[align=c,width=\imwidth]{\impatha{our_01/Places365_val_00000532_crop000_mask000.jpg}} &
	\includegraphics[align=c,width=\imwidth]{\impatha{our_02/Places365_val_00000532_crop000_mask000.jpg}} \\
  \rule{0pt}{6ex}%

   \includegraphics[align=c,width=\imwidth]{\impatha{input/Places365_val_00000641_crop000_mask000.jpg}} &
	    \includegraphics[align=c,width=\imwidth]{\impatha{gt/Places365_val_00000641_crop000_mask000.jpg}} &
	  \includegraphics[align=c,width=\imwidth]{\impatha{lama/Places365_val_00000641_crop000_mask000.jpg}} &
	\includegraphics[align=c,width=\imwidth]{\impatha{our_12/Places365_val_00000641_crop000_mask000.jpg}} &
	\includegraphics[align=c,width=\imwidth]{\impatha{our_05/Places365_val_00000641_crop000_mask000.jpg}} &
	\includegraphics[align=c,width=\imwidth]{\impatha{our_04/Places365_val_00000641_crop000_mask000.jpg}} \\
  \rule{0pt}{6ex}%

   \includegraphics[align=c,width=\imwidth]{\impatha{input/Places365_val_00000642_crop000_mask000.jpg}} &
	    \includegraphics[align=c,width=\imwidth]{\impatha{gt/Places365_val_00000642_crop000_mask000.jpg}} &
	  \includegraphics[align=c,width=\imwidth]{\impatha{lama/Places365_val_00000642_crop000_mask000.jpg}} &
	\includegraphics[align=c,width=\imwidth]{\impatha{our_06/Places365_val_00000642_crop000_mask000.jpg}} &
	\includegraphics[align=c,width=\imwidth]{\impatha{our_00/Places365_val_00000642_crop000_mask000.jpg}} &
	\includegraphics[align=c,width=\imwidth]{\impatha{our_02/Places365_val_00000642_crop000_mask000.jpg}} \\
  \rule{0pt}{6ex}%

   \includegraphics[align=c,width=\imwidth]{\impatha{input/Places365_val_00000721_crop000_mask000.jpg}} &
	    \includegraphics[align=c,width=\imwidth]{\impatha{gt/Places365_val_00000721_crop000_mask000.jpg}} &
	  \includegraphics[align=c,width=\imwidth]{\impatha{lama/Places365_val_00000721_crop000_mask000.jpg}} &
	\includegraphics[align=c,width=\imwidth]{\impatha{our_13/Places365_val_00000721_crop000_mask000.jpg}} &
	\includegraphics[align=c,width=\imwidth]{\impatha{our_07/Places365_val_00000721_crop000_mask000.jpg}} &
	\includegraphics[align=c,width=\imwidth]{\impatha{our_10/Places365_val_00000721_crop000_mask000.jpg}} \\
  \rule{0pt}{6ex}%

   \includegraphics[align=c,width=\imwidth]{\impatha{input/Places365_val_00000729_crop000_mask000.jpg}} &
	    \includegraphics[align=c,width=\imwidth]{\impatha{gt/Places365_val_00000729_crop000_mask000.jpg}} &
	  \includegraphics[align=c,width=\imwidth]{\impatha{lama/Places365_val_00000729_crop000_mask000.jpg}} &
	\includegraphics[align=c,width=\imwidth]{\impatha{our_00/Places365_val_00000729_crop000_mask000.jpg}} &
	\includegraphics[align=c,width=\imwidth]{\impatha{our_07/Places365_val_00000729_crop000_mask000.jpg}} &
	\includegraphics[align=c,width=\imwidth]{\impatha{our_15/Places365_val_00000729_crop000_mask000.jpg}} \\
  \rule{0pt}{6ex}%

   \includegraphics[align=c,width=\imwidth]{\impatha{input/Places365_val_00000812_crop000_mask000.jpg}} &
	    \includegraphics[align=c,width=\imwidth]{\impatha{gt/Places365_val_00000812_crop000_mask000.jpg}} &
	  \includegraphics[align=c,width=\imwidth]{\impatha{lama/Places365_val_00000812_crop000_mask000.jpg}} &
	\includegraphics[align=c,width=\imwidth]{\impatha{our_08/Places365_val_00000812_crop000_mask000.jpg}} &
	\includegraphics[align=c,width=\imwidth]{\impatha{our_00/Places365_val_00000812_crop000_mask000.jpg}} &
	\includegraphics[align=c,width=\imwidth]{\impatha{our_02/Places365_val_00000812_crop000_mask000.jpg}} \\
  \rule{0pt}{6ex}%

   \includegraphics[align=c,width=\imwidth]{\impatha{input/Places365_val_00000556_crop000_mask000.jpg}} &
	    \includegraphics[align=c,width=\imwidth]{\impatha{gt/Places365_val_00000556_crop000_mask000.jpg}} &
	  \includegraphics[align=c,width=\imwidth]{\impatha{lama/Places365_val_00000556_crop000_mask000.jpg}} &
	\includegraphics[align=c,width=\imwidth]{\impatha{our_00/Places365_val_00000556_crop000_mask000.jpg}} &
	\includegraphics[align=c,width=\imwidth]{\impatha{our_04/Places365_val_00000556_crop000_mask000.jpg}} &
	\includegraphics[align=c,width=\imwidth]{\impatha{our_02/Places365_val_00000556_crop000_mask000.jpg}} \\
	
	\bottomrule
  \end{tabular}
\caption{\label{fig:suppinpaintingsamples} Qualitative results on image inpainting. In contrast to \cite{lama}, our generative approach enables generation of multiple diverse samples for a given input.}
\end{figure}
}

\newcommand{\suppinpaintingremoval}{
\begin{figure}
\centering
\scriptsize
	\renewcommand{\imwidth}{0.25\textwidth}
    \renewcommand{\impatha}[1]{img/object_removal/##1} 
	\setlength{\tabcolsep}{0pt}
  \begin{tabular}{c@{\hskip 2pt}c@{\hskip 6pt}c@{\hskip 2pt}c}
	\toprule
    input & result & input & result \\
	\midrule

    \includegraphics[align=c,width=\imwidth]{\impatha{input/000026.jpg}} &
    \includegraphics[align=c,width=\imwidth]{\impatha{our_02/000026.jpg}} &

    \includegraphics[align=c,width=\imwidth]{\impatha{input/000028.jpg}} &
   \includegraphics[align=c,width=\imwidth]{\impatha{our_02/000028.jpg}} \\
  \rule{0pt}{6ex}%

    \includegraphics[align=c,width=\imwidth]{\impatha{input/000029.jpg}} &
    \includegraphics[align=c,width=\imwidth]{\impatha{our_01/000029.jpg}} &

    \includegraphics[align=c,width=\imwidth]{\impatha{input/000031.jpg}} &
   \includegraphics[align=c,width=\imwidth]{\impatha{our_01/000031.jpg}} \\
  \rule{0pt}{6ex}%

    \includegraphics[align=c,width=\imwidth]{\impatha{input/000032.jpg}} &
    \includegraphics[align=c,width=\imwidth]{\impatha{our_02/000032.jpg}} &

    \includegraphics[align=c,width=\imwidth]{\impatha{input/000034.jpg}} &
   \includegraphics[align=c,width=\imwidth]{\impatha{our_02/000034.jpg}} \\
  \rule{0pt}{6ex}%

    \includegraphics[align=c,width=\imwidth]{\impatha{input/000038.jpg}} &
    \includegraphics[align=c,width=\imwidth]{\impatha{our_02/000038.jpg}} &

    \includegraphics[align=c,width=\imwidth]{\impatha{input/000039.jpg}} &
   \includegraphics[align=c,width=\imwidth]{\impatha{our_01/000039.jpg}} \\
  \rule{0pt}{6ex}%

    \includegraphics[align=c,width=\imwidth]{\impatha{input/000011.jpg}} &
    \includegraphics[align=c,width=\imwidth]{\impatha{our_01/000011.jpg}} &

    \includegraphics[align=c,width=\imwidth]{\impatha{input/000042.jpg}} &
   \includegraphics[align=c,width=\imwidth]{\impatha{our_01/000042.jpg}} \\
	\bottomrule
  \end{tabular}
\caption{\label{suppinpaintingremoval} More qualitative results on object
removal as in Fig.~\ref{inpaintingremoval}.}
\end{figure}
}

\newcommand{\imagenetsamplesone}{
\begin{figure}[htbp]
\centering
	\renewcommand{\imwidth}{0.9\textwidth}
	\setlength{\tabcolsep}{0pt}
	\begin{tabular}{c}
	\toprule
	Random class conditional samples on the ImageNet dataset \\
	\midrule
	\includegraphics[align=c,width=\imwidth]{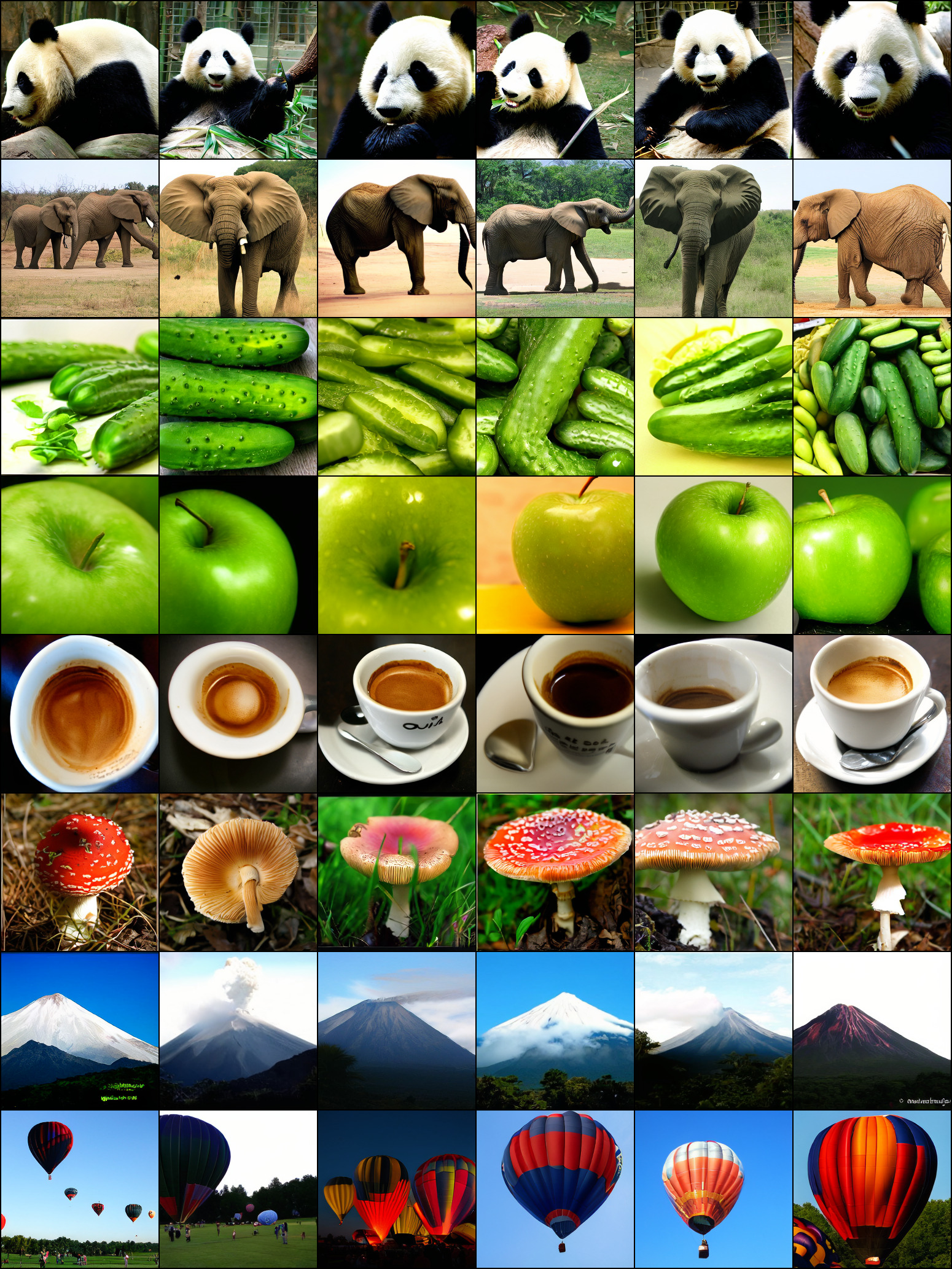}\\
	\bottomrule
	\end{tabular}

\caption{\label{fig:imagenet_samples_1} Random samples from \emph{LDM-4} trained on the ImageNet dataset. Sampled with classifier-free guidance \cite{ho2021classifier} scale $s=5.0$ and 200 DDIM steps with $\eta = 1.0$. }
\end{figure}
}

\newcommand{\imagenetsamplestwo}{
\begin{figure}[htbp]
\centering
	\renewcommand{\imwidth}{0.9\textwidth}
	\setlength{\tabcolsep}{0pt}
	\begin{tabular}{c}
	\toprule
	Random class conditional samples on the ImageNet dataset \\
	\midrule
	\includegraphics[align=c,width=\imwidth]{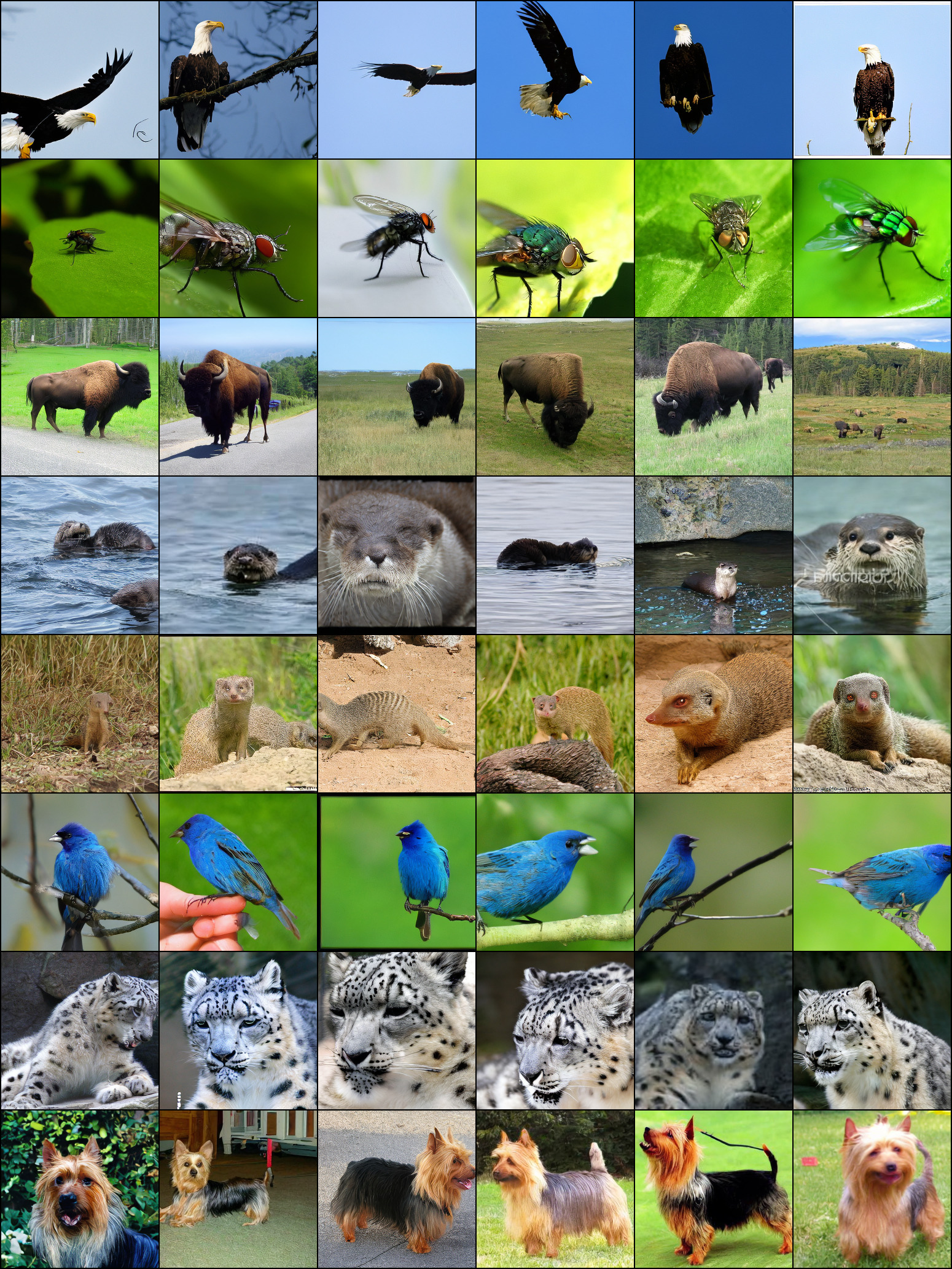}\\
	\bottomrule
	\end{tabular}

\caption{\label{fig:imagenet_samples_2} Random samples from \emph{LDM-4} trained on the ImageNet dataset. Sampled with classifier-free guidance \cite{ho2021classifier} scale $s=3.0$ and 200 DDIM steps with $\eta = 1.0$. }
\end{figure}
}

\newcommand{\randomcelebasamples}{
\begin{figure}[htbp]
\centering
	\renewcommand{\imwidth}{0.9\textwidth}
	\setlength{\tabcolsep}{0pt}
	\begin{tabular}{c}
	\toprule
	Random samples on the CelebA-HQ dataset \\
	\midrule
	\includegraphics[align=c,width=\imwidth]{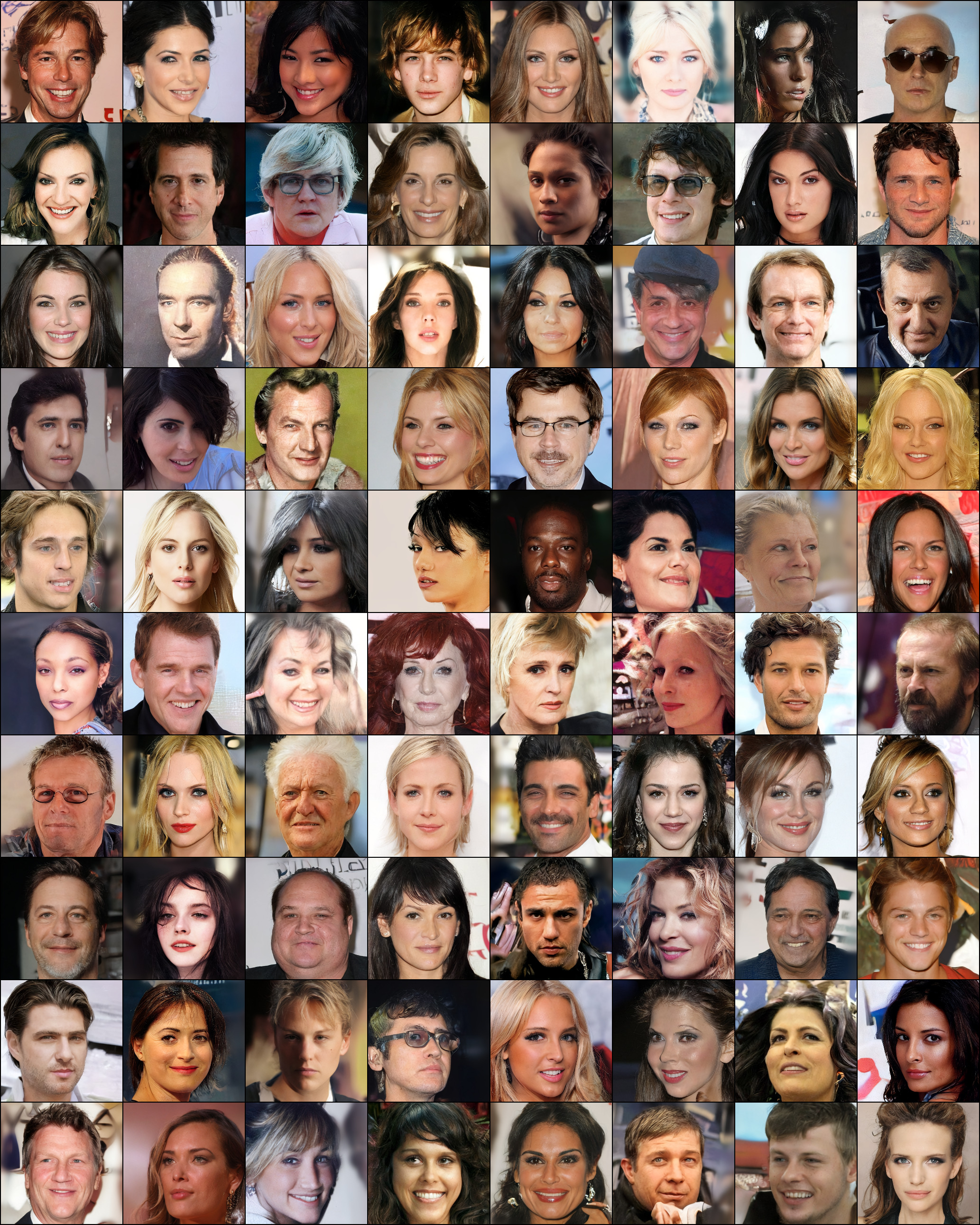}\\
	\bottomrule
	\end{tabular}
	
\caption{\label{fig:celeba_rsamples} Random samples of our best performing model \emph{LDM-4} on the CelebA-HQ dataset. Sampled with 500 DDIM steps and $\eta=0$ (FID = 5.15).}
\end{figure}
}

\newcommand{\randomffhqsamples}{
\begin{figure}[htbp]
\centering
	\renewcommand{\imwidth}{0.9\textwidth}
	\setlength{\tabcolsep}{0pt}
	\begin{tabular}{c}
	\toprule
	Random samples on the FFHQ dataset \\
	\midrule
	\includegraphics[align=c,width=\imwidth]{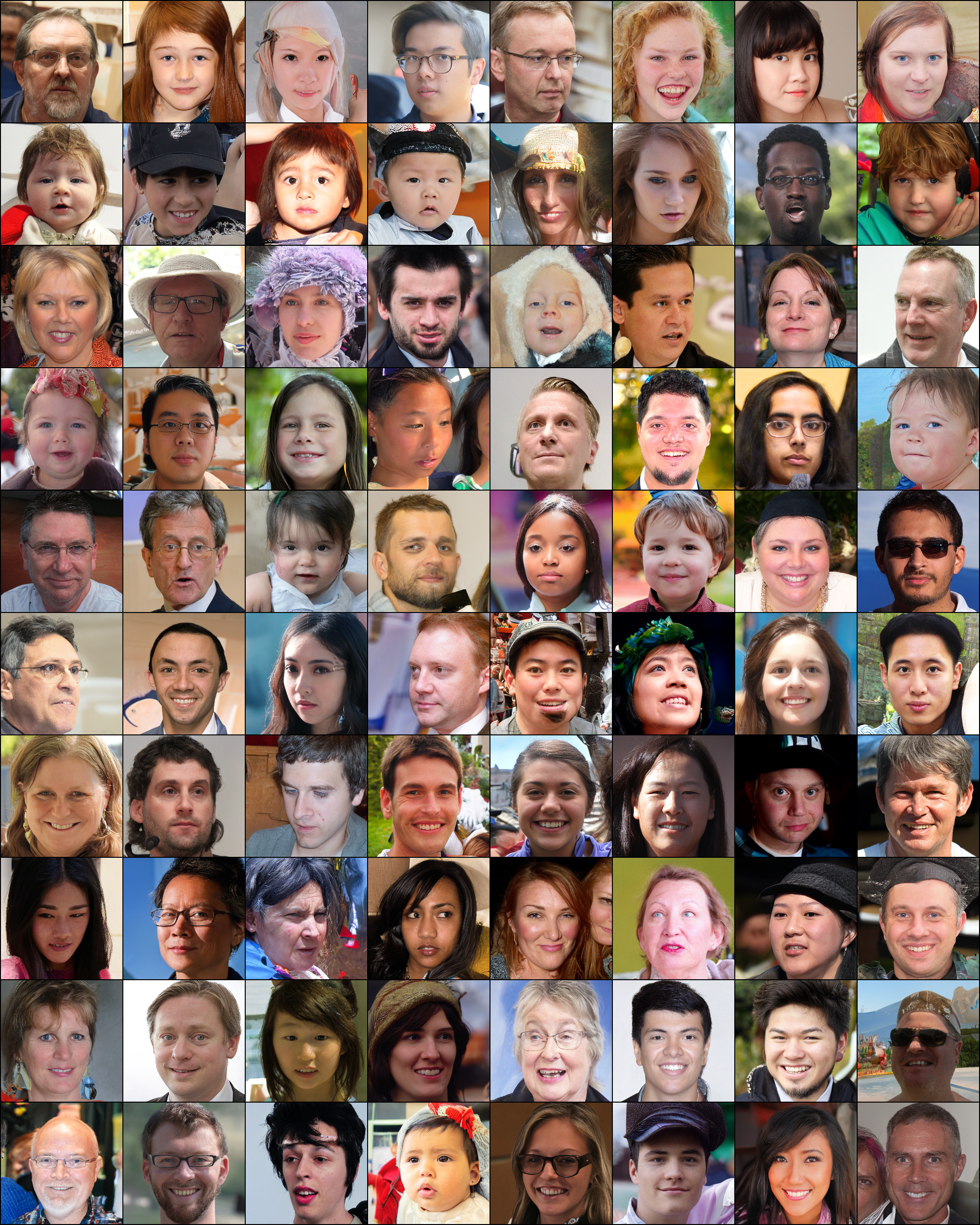}\\
	\bottomrule
	\end{tabular}
	
\caption{\label{fig:ffhq_rsamples} Random samples of our best performing model \emph{LDM-4} on the FFHQ dataset. Sampled with 200 DDIM steps and $\eta=1$ (FID = 4.98).}
\end{figure}
}

\newcommand{\randomchurchsamples}{
\begin{figure}[htbp]
\centering
	\renewcommand{\imwidth}{0.9\textwidth}
	\setlength{\tabcolsep}{0pt}
	\begin{tabular}{c}
	\toprule
	Random samples on the LSUN-Churches dataset \\
	\midrule
	\includegraphics[align=c,width=\imwidth]{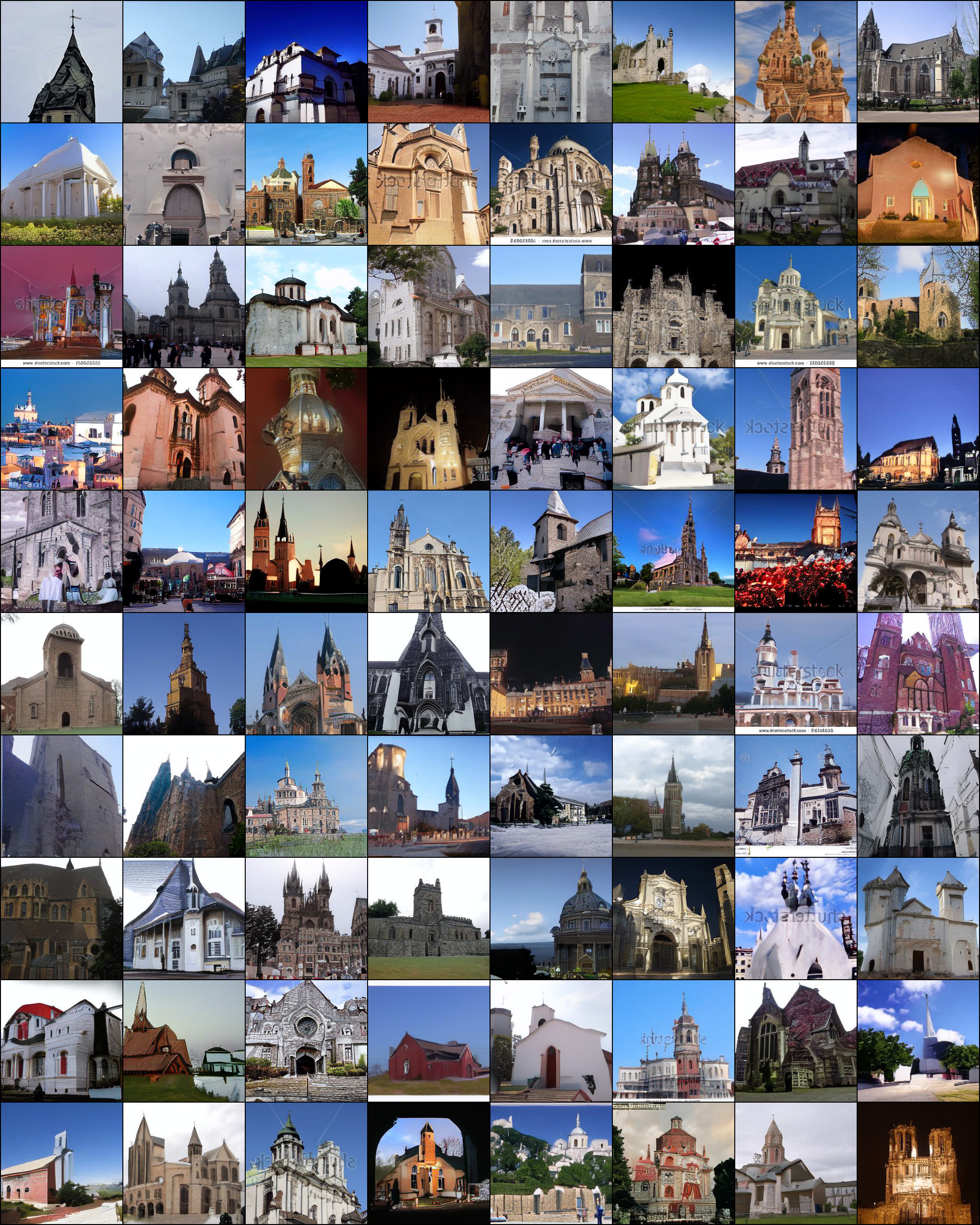}\\
	\bottomrule
	\end{tabular}
	
\caption{\label{fig:church_rsamples} Random samples of our best performing model \emph{LDM-8} on the LSUN-Churches dataset. Sampled with 200 DDIM steps and $\eta=0$ (FID = 4.48).}
\end{figure}
}

\newcommand{\randombedssamples}{
\begin{figure}[htbp]
\centering
	\renewcommand{\imwidth}{0.9\textwidth}
	\setlength{\tabcolsep}{0pt}
	\begin{tabular}{c}
	\toprule
	Random samples on the LSUN-Bedrooms dataset \\
	\midrule
	\includegraphics[align=c,width=\imwidth]{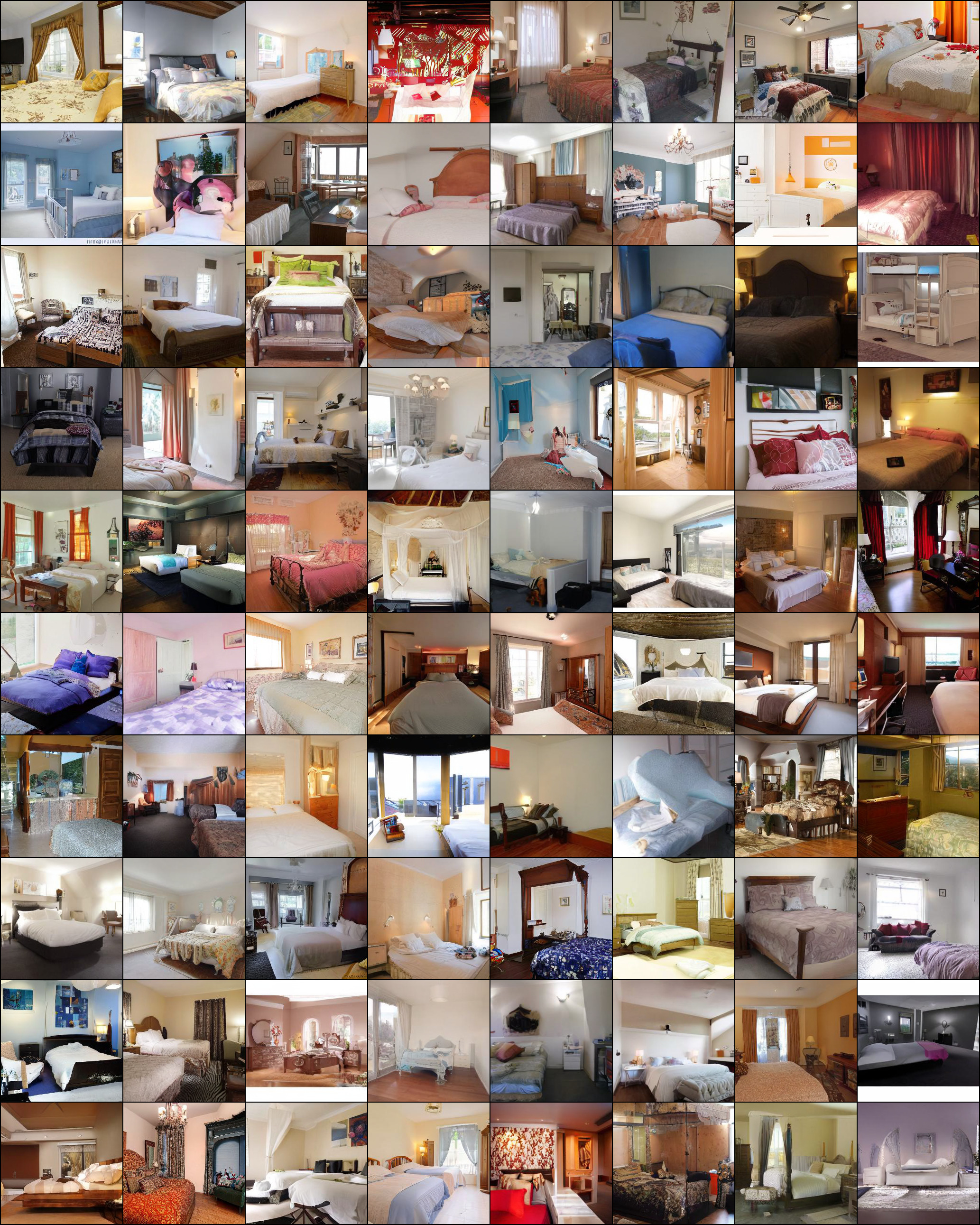}\\
	\bottomrule
	\end{tabular}
	
\caption{\label{fig:beds_rsamples} Random samples of our best performing model \emph{LDM-4} on the LSUN-Bedrooms dataset. Sampled with 200 DDIM steps and $\eta=1$ (FID = 2.95).}
\end{figure}

}

\newcommand{\nnsceleba}{
\begin{figure}[htbp]
\centering
	\renewcommand{\imwidth}{0.9\textwidth}
	\setlength{\tabcolsep}{0pt}
	\begin{tabular}{c}
	\toprule
	Nearest Neighbors on the CelebA-HQ dataset \\
	\midrule
	\includegraphics[align=c,width=\imwidth]{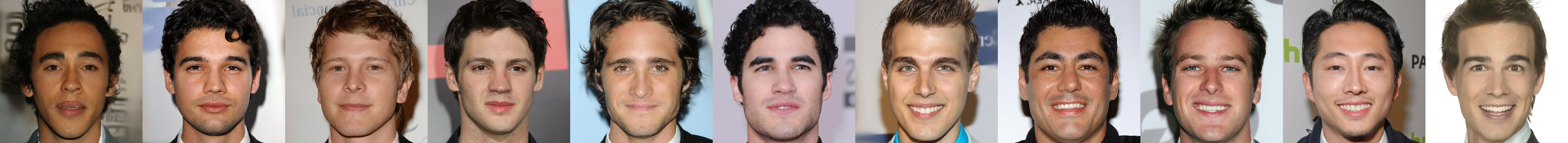}\\
	\includegraphics[align=c,width=\imwidth]{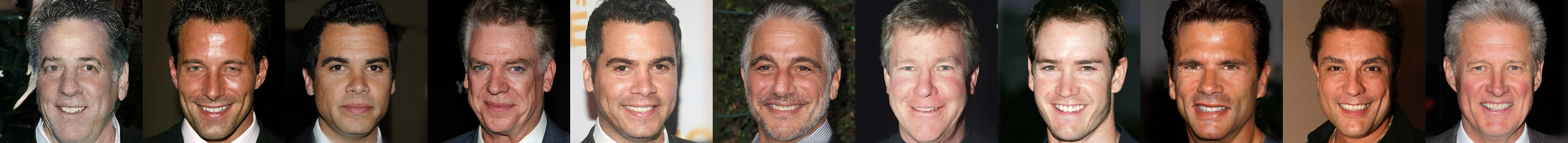}\\
	\includegraphics[align=c,width=\imwidth]{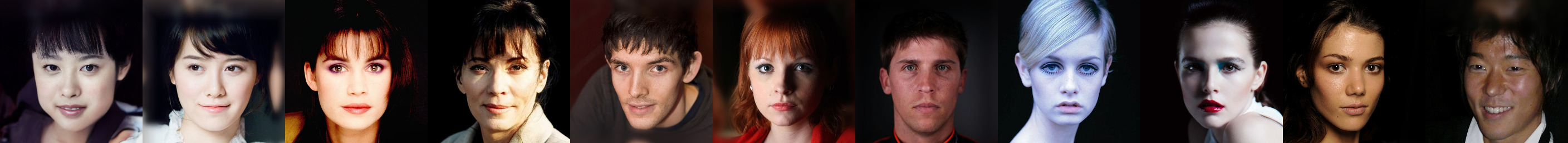}\\
	\includegraphics[align=c,width=\imwidth]{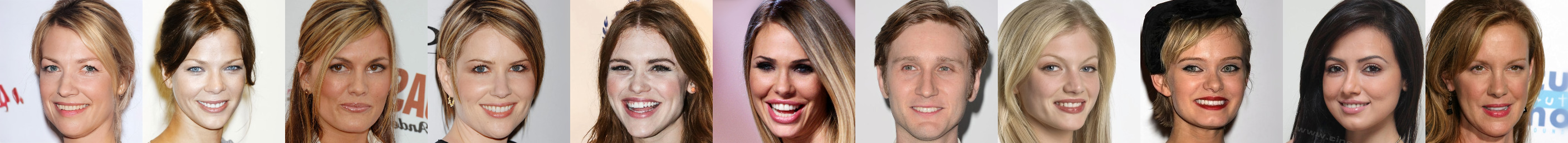}\\
	\includegraphics[align=c,width=\imwidth]{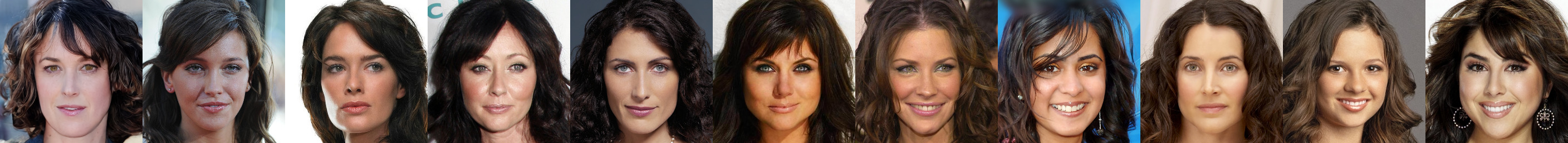}\\
	\includegraphics[align=c,width=\imwidth]{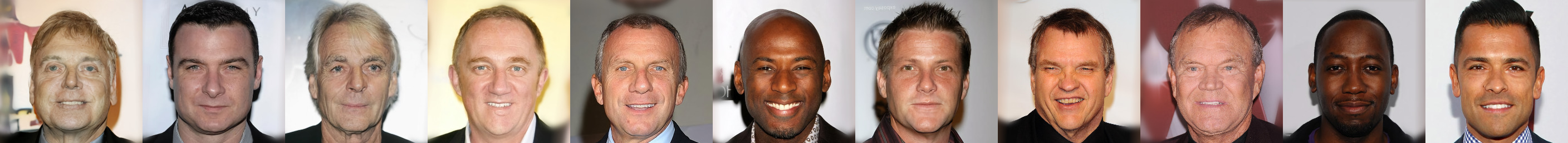}\\
	\includegraphics[align=c,width=\imwidth]{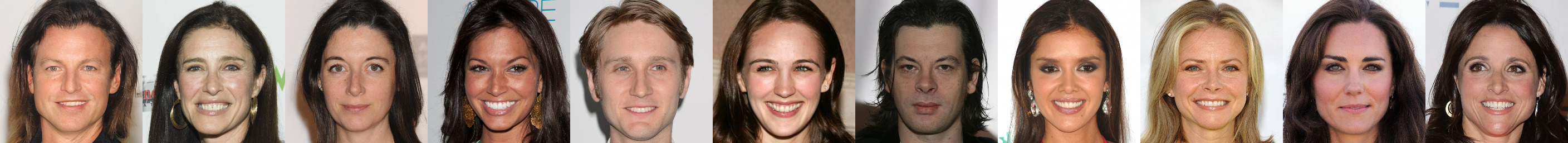}\\
	\includegraphics[align=c,width=\imwidth]{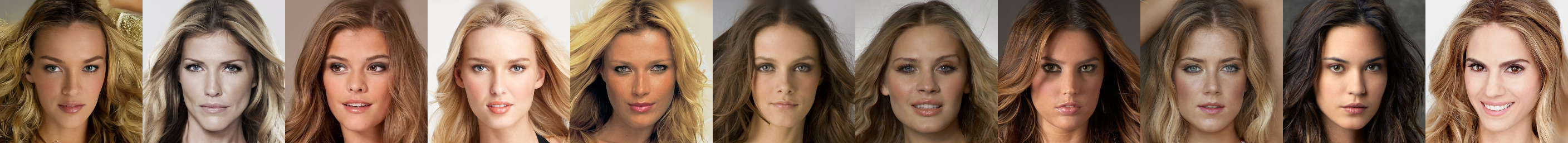}\\
	\includegraphics[align=c,width=\imwidth]{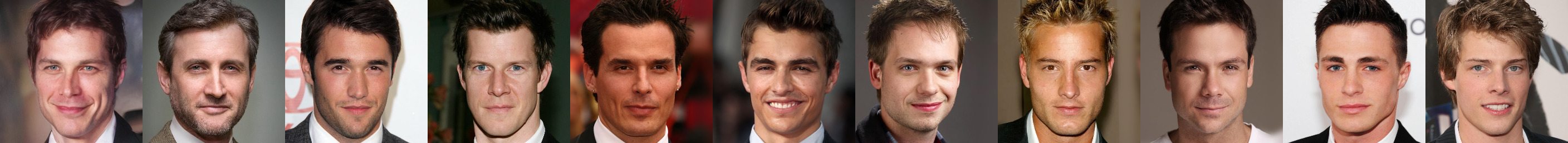}\\
	\includegraphics[align=c,width=\imwidth]{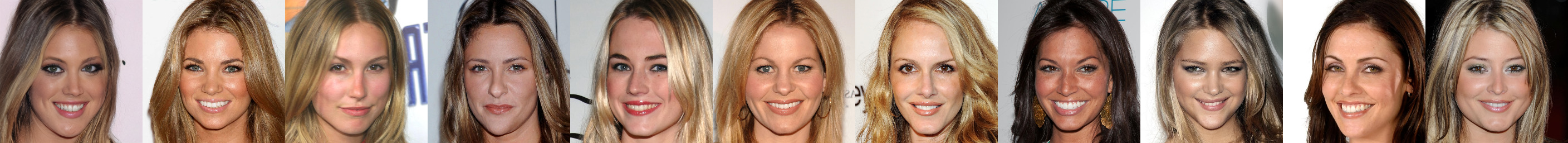}\\
	\bottomrule
	\end{tabular}
	
\caption{\label{fig:celeba_nns} Nearest neighbors of our best CelebA-HQ model, computed in the feature space of a VGG-16~\cite{simonyan2015VGG}. The leftmost sample is from our model. The remaining samples in each row are its 10 nearest neighbors.}
\end{figure}

}

\newcommand{\nnsffhq}{
\begin{figure}[htbp]
\centering
	\renewcommand{\imwidth}{0.9\textwidth}
    \renewcommand{\impath}[1]{img/supplement/nns/ffhq/##1}%
	\setlength{\tabcolsep}{0pt}
	\begin{tabular}{c}
	\toprule
	Nearest Neighbors on the FFHQ dataset \\
	\midrule
	\includegraphics[align=c,width=\imwidth]{\impath{example_0-10nns}}\\
	\includegraphics[align=c,width=\imwidth]{\impath{example_4-10nns}}\\
	\includegraphics[align=c,width=\imwidth]{\impath{example_5-10nns}}\\
	\includegraphics[align=c,width=\imwidth]{\impath{example_6-10nns}}\\
	\includegraphics[align=c,width=\imwidth]{\impath{example_8-10nns}}\\
	\includegraphics[align=c,width=\imwidth]{\impath{example_12-10nns}}\\
	\includegraphics[align=c,width=\imwidth]{\impath{example_13-10nns}}\\
	\includegraphics[align=c,width=\imwidth]{\impath{example_14-10nns}}\\
	\includegraphics[align=c,width=\imwidth]{\impath{example_17-10nns}}\\
	\includegraphics[align=c,width=\imwidth]{\impath{example_18-10nns}}\\
	\bottomrule
	\end{tabular}
	
\caption{\label{fig:ffhq_nns} Nearest neighbors of our best FFHQ model, computed in the feature space of a VGG-16~\cite{simonyan2015VGG}. The leftmost sample is from our model. The remaining samples in each row are its 10 nearest neighbors.}
\end{figure}

}

\newcommand{\nnschurches}{
\begin{figure}[htbp]
\centering
	\renewcommand{\imwidth}{0.9\textwidth}
    \renewcommand{\impath}[1]{img/supplement/nns/churches/##1}%
	\setlength{\tabcolsep}{0pt}
	\begin{tabular}{c}
	\toprule
	Nearest Neighbors on the LSUN-Churches dataset \\
	\midrule
	\includegraphics[align=c,width=\imwidth]{\impath{example_0-10nns}}\\
	\includegraphics[align=c,width=\imwidth]{\impath{example_4-10nns}}\\
	\includegraphics[align=c,width=\imwidth]{\impath{example_5-10nns}}\\
	\includegraphics[align=c,width=\imwidth]{\impath{example_6-10nns}}\\
	\includegraphics[align=c,width=\imwidth]{\impath{example_8-10nns}}\\
	\includegraphics[align=c,width=\imwidth]{\impath{example_1-10nns}}\\
	\includegraphics[align=c,width=\imwidth]{\impath{example_10-10nns}}\\
	\includegraphics[align=c,width=\imwidth]{\impath{example_11-10nns}}\\
	\includegraphics[align=c,width=\imwidth]{\impath{example_19-10nns}}\\
	\includegraphics[align=c,width=\imwidth]{\impath{example_18-10nns}}\\
	\bottomrule
	\end{tabular}
	
\caption{\label{fig:churches_nns} Nearest neighbors of our best LSUN-Churches model, computed in the feature space of a VGG-16~\cite{simonyan2015VGG}. The leftmost sample is from our model. The remaining samples in each row are its 10 nearest neighbors.}
\end{figure}

}

\newcommand{\layouttoimagesamples}{
\begin{figure}[htbp]
\centering
	\renewcommand{\imwidth}{0.12\textwidth}
    \renewcommand{\impath}[1]{img/supplement/layouts/##1}%
    \renewcommand{\impatha}[1]{img/goodsamples/bbox2img/street/##1} 
    \renewcommand{\impathb}[1]{img/goodsamples/bbox2img/table/##1} 
	\setlength{\tabcolsep}{0pt}
	\begin{tabular}{c@{\hskip 2pt}c@{\hskip 2pt}c@{\hskip 2pt}c@{\hskip 2pt}c@{\hskip 2pt}c@{\hskip 2pt}c@{\hskip 2pt}c}
	\toprule
	\multicolumn{8}{c}{layout-to-image synthesis on the COCO dataset} \\
	\midrule
    \includegraphics[align=c,width=\imwidth]{\impathb{cond-185_}} &	         
    \includegraphics[align=c,width=\imwidth]{\impatha{cond-2_}} &	         
	\includegraphics[align=c,width=\imwidth]{\impath{indoor/cond-451}} &
	\includegraphics[align=c,width=\imwidth]{\impath{rider/cond-494}} &
	\includegraphics[align=c,width=\imwidth]{\impath{giraffe/cond-252}} &
	\includegraphics[align=c,width=\imwidth]{\impath{bathroom/cond-104}} &
	\includegraphics[align=c,width=\imwidth]{\impath{elefant/cond-508}} &
	\includegraphics[align=c,width=\imwidth]{\impath{person/cond-233}} \\
	
	\includegraphics[align=c,width=\imwidth]{\impathb{sample-219}} &	        
	\includegraphics[align=c,width=\imwidth]{\impatha{sample-216}} &	
	\includegraphics[align=c,width=\imwidth]{\impath{indoor/sample-450}} &
	\includegraphics[align=c,width=\imwidth]{\impath{rider/sample-445}} &
	\includegraphics[align=c,width=\imwidth]{\impath{giraffe/sample-239}} &
	\includegraphics[align=c,width=\imwidth]{\impath{bathroom/sample-91}} &
	\includegraphics[align=c,width=\imwidth]{\impath{elefant/sample-507}} &
	\includegraphics[align=c,width=\imwidth]{\impath{person/sample-400}} \\
	
	\includegraphics[align=c,width=\imwidth]{\impathb{sample-319}} &	
	\includegraphics[align=c,width=\imwidth]{\impatha{sample-251}} &	        
	\includegraphics[align=c,width=\imwidth]{\impath{indoor/sample-462}} &
	\includegraphics[align=c,width=\imwidth]{\impath{rider/sample-493}} &
	\includegraphics[align=c,width=\imwidth]{\impath{giraffe/sample-251}} &
	\includegraphics[align=c,width=\imwidth]{\impath{bathroom/sample-127}} &
	\includegraphics[align=c,width=\imwidth]{\impath{elefant/sample-579}} &
	\includegraphics[align=c,width=\imwidth]{\impath{person/sample-292}} \\
	
	\includegraphics[align=c,width=\imwidth]{\impathb{sample-249}} &	        
	\includegraphics[align=c,width=\imwidth]{\impatha{sample-439}} &  
	\includegraphics[align=c,width=\imwidth]{\impath{indoor/sample-474}} &
	\includegraphics[align=c,width=\imwidth]{\impath{rider/sample-541}} &
	\includegraphics[align=c,width=\imwidth]{\impath{giraffe/sample-275}} &
	\includegraphics[align=c,width=\imwidth]{\impath{bathroom/sample-175}} &
	\includegraphics[align=c,width=\imwidth]{\impath{elefant/sample-555}} &
	\includegraphics[align=c,width=\imwidth]{\impath{person/sample-232}} \\
	
	\includegraphics[align=c,width=\imwidth]{\impathb{sample-259}} &
    \includegraphics[align=c,width=\imwidth]{\impatha{sample-266}} &
	\includegraphics[align=c,width=\imwidth]{\impath{indoor/sample-486}} &
	\includegraphics[align=c,width=\imwidth]{\impath{rider/sample-505}} &
	\includegraphics[align=c,width=\imwidth]{\impath{giraffe/sample-311}} &
	\includegraphics[align=c,width=\imwidth]{\impath{bathroom/sample-187}} &
	\includegraphics[align=c,width=\imwidth]{\impath{elefant/sample-567}} &
	\includegraphics[align=c,width=\imwidth]{\impath{person/sample-340}} \\
	
	\bottomrule
	\end{tabular}
	
\caption{\label{fig:lay2img_samples} More samples from our best model for layout-to-image synthesis, \emph{LDM-4}, which was trained on the OpenImages dataset and finetuned on the COCO dataset. Samples generated with 100 DDIM steps and $\eta = 0$. Layouts are from the COCO validation set.}
\end{figure}
}

\newcommand{\newtexttoimagesamples}{
\begin{figure*}[htbp]
\centering
	\renewcommand{\imwidth}{0.14\textwidth}
    \renewcommand{\impath}[1]{img/cr/text2img/##1}%
	\setlength{\tabcolsep}{1pt}
	\begin{tabular}{ccccccc}
	\toprule
	\multicolumn{7}{c}{\small{Text-to-Image Synthesis on LAION. 1.45B Model.}} \\
	\midrule

    \shortstack{\tiny{\emph{'A street sign that reads}} \\ \tiny{\emph{``Latent Diffusion'' '}}} & \shortstack{\tiny{\emph{'A zombie in the}} \\ \tiny{\emph{style of Picasso'}}} & \shortstack{\tiny{\emph{'An image of an animal}} \\ \tiny{\emph{half mouse half octopus'}}} & \shortstack{\tiny{\emph{'An illustration of a slightly}} \\ \tiny{\emph{conscious neural network'}}} & \shortstack{\tiny{\emph{'A painting of a}} \\ \tiny{\emph{squirrel eating a burger'}}}  & \shortstack{\tiny{\emph{'A watercolor painting of a}} \\ \tiny{\emph{chair that looks like an octopus'}}} & \shortstack{\tiny{\emph{'A shirt with the inscription:}}\\ \tiny{\emph{``I love generative models!'' '}}} \\

	\midrule
	\includegraphics[align=c,width=\imwidth]{\impath{sign/sample-43}} &
	\includegraphics[align=c,width=\imwidth]{\impath{zombiepicasso/sample-983}} &
	\includegraphics[align=c,width=\imwidth]{\impath{halfmousehalfoct/sample-850}} &
	\includegraphics[align=c,width=\imwidth]{\impath{network/sample-401}} &
	\includegraphics[align=c,width=\imwidth]{\impath{squirrelburger/sample-80}} &
	\includegraphics[align=c,width=\imwidth]{\impath{octochair/sample-366}} &
	\includegraphics[align=c,width=\imwidth]{\impath{generativeshirt/sample-968}} \\
	
	\includegraphics[align=c,width=\imwidth]{\impath{sign/sample-79}} &
	\includegraphics[align=c,width=\imwidth]{\impath{zombiepicasso/sample-991}} &
	\includegraphics[align=c,width=\imwidth]{\impath{halfmousehalfoct/sample-868}} &
	\includegraphics[align=c,width=\imwidth]{\impath{network/sample-409}} &
	\includegraphics[align=c,width=\imwidth]{\impath{squirrelburger/sample-59}} &
	\includegraphics[align=c,width=\imwidth]{\impath{octochair/sample-383}} &
	\includegraphics[align=c,width=\imwidth]{\impath{generativeshirt/sample-975}} \\

	\bottomrule
	\end{tabular}
\caption{\label{fig:text2img_samples} Samples for user-defined text prompts from our model for text-to-image synthesis, \emph{LDM-8 (KL)}, which was trained on the LAION~\cite{schuhmann2021laion400m} database. Samples generated with 200 DDIM steps and $\eta = 1.0$. We use unconditional guidance \cite{ho2021classifier} with $s=10.0$.}
\end{figure*}
}

\newcommand{\firststagetablecomplete}{
\begin{table}[thbp]
    \begin{center}
	\begin{footnotesize}
  \begin{adjustbox}{max width=0.9\textwidth}
      \begin{tabular}{c c c c c c c c}
        \toprule
$f$ & $\vert \mathcal{Z} \vert $ & $\clatent$ & \textbf{R-FID} $\downarrow$ & \textbf{R-IS} $\uparrow$ & \textbf{PSNR} $\uparrow$ & \textbf{PSIM} $\downarrow$ & \textbf{SSIM} $\uparrow$\\
        \midrule 
16 \emph{VQGAN}~\cite{DBLP:journals/corr/abs-2012-09841} & 16384 & 256 & 4.98 & -- & 19.9 \tiny{$\pm3.4$} & 1.83 \tiny{$\pm0.42$} & 0.51 \tiny{$\pm0.18$}\\
16 \emph{VQGAN}~\cite{DBLP:journals/corr/abs-2012-09841} & 1024 & 256 & 7.94 &  -- & 19.4 \tiny{$\pm3.3$} & 1.98 \tiny{$\pm0.43$} & 0.50 \tiny{$\pm0.18$} \\
8 \emph{DALL-E}~\cite{DBLP:journals/corr/abs-2102-12092} & 8192 & - & 32.01 & -- & 22.8 \tiny{$\pm2.1$} & 1.95 \tiny{$\pm0.51$} & 0.73 \tiny{$\pm0.13$}\\
\midrule 
 32 & 16384 & 16 & 31.83 & 40.40 \tiny{$\pm1.07$} & 17.45 \tiny{$\pm2.90$} & 2.58 \tiny{$\pm0.48$} & 0.41 \tiny{$\pm0.18$} \\
 16 & 16384 & 8 & 5.15 & 144.55 \tiny{$\pm3.74$} & 20.83 \tiny{$\pm3.61$} & 1.73 \tiny{$\pm0.43$} & 0.54 \tiny{$\pm0.18$} \\
 8 & 16384 & 4 & 1.14 & 201.92 \tiny{$\pm3.97$} & 23.07 \tiny{$\pm3.99$} & 1.17 \tiny{$\pm0.36$} & 0.65 \tiny{$\pm0.16$} \\
 8 & 256 & 4 & 1.49 & 194.20 \tiny{$\pm3.87$} & 22.35 \tiny{$\pm3.81$} & 1.26 \tiny{$\pm0.37$} & 0.62 \tiny{$\pm0.16$} \\
 4 & 8192 & 3 & 0.58 & 224.78 \tiny{$\pm5.35$} & 27.43 \tiny{$\pm4.26$} & 0.53 \tiny{$\pm0.21$} & 0.82 \tiny{$\pm0.10$} \\

 4$^\dagger$ & 8192 & 3 & 1.06 & 221.94 \tiny{$\pm4.58$} & 25.21 \tiny{$\pm4.17$} & 0.72 \tiny{$\pm0.26$} & 0.76 \tiny{$\pm0.12$} \\
 4 & 256 & 3 & 0.47 & 223.81 \tiny{$\pm4.58$} & 26.43 \tiny{$\pm4.22$} & 0.62 \tiny{$\pm0.24$} & 0.80 \tiny{$\pm0.11$} \\
2 & 2048 & 2 & 0.16 & 232.75 \tiny{$\pm5.09$} & 30.85 \tiny{$\pm4.12$} & 0.27 \tiny{$\pm0.12$} & 0.91 \tiny{$\pm0.05$} \\
2 & 64 & 2 & 0.40 & 226.62 \tiny{$\pm4.83$} & 29.13 \tiny{$\pm3.46$} & 0.38 \tiny{$\pm0.13$} & 0.90 \tiny{$\pm0.05$} \\
\midrule
32 & KL & 64 & 2.04 & 189.53 \tiny{$\pm3.68$} & 22.27 \tiny{$\pm3.93$} & 1.41 \tiny{$\pm0.40$} & 0.61 \tiny{$\pm0.17$} \\
32 & KL & 16 & 7.3 & 132.75 \tiny{$\pm2.71$} & 20.38 \tiny{$\pm3.56$} & 1.88 \tiny{$\pm0.45$} & 0.53 \tiny{$\pm0.18$} \\
16 & KL & 16 & 0.87 & 210.31 \tiny{$\pm3.97$} & 24.08 \tiny{$\pm4.22$} & 1.07 \tiny{$\pm0.36$} & 0.68 \tiny{$\pm0.15$} \\
16 & KL & 8 & 2.63 & 178.68 \tiny{$\pm4.08$} & 21.94 \tiny{$\pm3.92$} & 1.49 \tiny{$\pm0.42$} & 0.59 \tiny{$\pm0.17$} \\
8  & KL & 4 & 0.90 & 209.90 \tiny{$\pm4.92$} & 24.19 \tiny{$\pm4.19$} & 1.02 \tiny{$\pm0.35$} & 0.69 \tiny{$\pm0.15$} \\
4  & KL & 3 & 0.27 & 227.57 \tiny{$\pm4.89$} & 27.53 \tiny{$\pm4.54$} & 0.55 \tiny{$\pm0.24$} & 0.82 \tiny{$\pm0.11$} \\
2  & KL & 2 & 0.086 & 232.66 \tiny{$\pm5.16$} & 32.47 \tiny{$\pm4.19$} & 0.20 \tiny{$\pm0.09$} & 0.93 \tiny{$\pm0.04$} \\
        \bottomrule 
      \end{tabular}  
      \end{adjustbox}
            \caption{\label{tab:firststagetablecomplete} Complete autoencoder zoo trained on OpenImages, evaluated on ImageNet-Val. $\dagger$ denotes an attention-free autoencoder. \vspace{-1em}}
      	\end{footnotesize}
    \end{center}
  \end{table}
}

\newcommand{\srsuppptable}{
  \begin{table*}[pbth]
  \centering
\begin{footnotesize}
  \begin{adjustbox}{max width=\linewidth}
  \begin{tabular}{l c c c c}
  	\toprule
  	\textbf{Method} & FID $\downarrow$ & IS $\uparrow$ & PSNR  $\uparrow$  & SSIM  $\uparrow$  \\
  	\midrule
  	Image Regression \cite{DBLP:journals/corr/abs-2104-07636} & 15.2 & 121.1 & \textbf{27.9} & \textbf{0.801} \\
  	SR3 \cite{DBLP:journals/corr/abs-2104-07636} & 5.2 & \textbf{180.1} & 26.4 & 0.762 \\
  	\midrule
  	\emph{LDM-4} (ours, 100 steps) & \textbf{2.8}$^\dagger$/\textbf{4.8}$^\ddagger$ & 166.3 & 24.4\tiny{$\pm $3.8} & 0.69\tiny{$\pm $0.14} \\
  	\emph{LDM-4} (ours, 50 steps, guiding) & 4.4$^\dagger$/6.4$^\ddagger$ & 153.7 & 25.8\tiny{$\pm $3.7} & 0.74\tiny{$\pm $0.12} \\
  	\emph{LDM-4} (ours, 100 steps, guiding) & 4.4$^\dagger$/6.4$^\ddagger$ & 154.1 & 25.7\tiny{$\pm $3.7} & 0.73\tiny{$\pm $0.12} \\
  	\midrule
  	\emph{LDM-4} (ours, 100 steps, +15 ep.) & \textbf{2.6}$^\dagger$ / \textbf{4.6}$^\ddagger$ & 169.76\tiny{$\pm $5.03} & 24.4\tiny{$\pm $3.8} & 0.69\tiny{$\pm $0.14} \\
  	Pixel-DM (100 steps, +15 ep.) & 5.1$^\dagger$ / 7.1$^\ddagger$ & 163.06\tiny{$\pm $4.67} & 24.1\tiny{$\pm $3.3} & 0.59\tiny{$\pm $0.12}  \\
  	\bottomrule

  \end{tabular}
  \end{adjustbox}
\end{footnotesize}\vspace{-0.85em}
  \caption{\label{tab:srsupptable}  $\times 4$ upscaling results on ImageNet-Val. ($256^2$); $^\dagger$: FID features computed on validation split, $^\ddagger$: FID features computed on train split. We also include a pixel-space baseline that receives the same amount of compute as \emph{LDM-4}. The last two rows received 15 epochs of additional training compared to the former results. }
\end{table*}
}

\newcommand{\computevsfid}{
\begin{table}[thbp]
\begin{center}
\begin{adjustbox}{max width=.95\textwidth}

\begin{tabular}{lccccccccc}
\toprule
\textbf{Method} & Generator  & Classifier & Overall & Inference & $N_{\text{params}}$ & FID$\downarrow$  & IS$\uparrow$ & Precision$\uparrow$ & Recall$\uparrow$\\
&Compute&Compute&Compute& Throughput$^*$ &&&&& \\
\toprule
&&&&&&&&&\\
\textbf{LSUN Churches $256^{2}$} & & & &&& &&&\\
\midrule
StyleGAN2~\cite{DBLP:journals/corr/abs-1912-04958}$^\dagger$& 64 & - & 64 & - & 59M  & 3.86& - & - & - \\
\emph{LDM-8} (ours, 100 steps, 410K)& 18 & - & 18 & 6.80  & 256M & 4.02& - & 0.64 & 0.52\\
&&&&&&&&&\\
\textbf{LSUN Bedrooms $256^{2}$} &  &  &&& & &&&\\
\midrule
ADM~\cite{DBLP:journals/corr/abs-2105-05233}$^\dagger$ (1000 steps)& 232 & - & 232 & 0.03 & 552M & 1.9 & - & 0.66 & 0.51 \\
\emph{LDM-4} (ours, 200 steps, 1.9M)& 60 & - & 55 & 1.07 & 274M & 2.95& - & 0.66 & 0.48 \\
&&&&&&&&&\\
\textbf{CelebA-HQ $256^{2}$} &&&&&& &&&\\
\midrule
\emph{LDM-4} (ours, 500 steps, 410K)& 14.4 & - & 14.4 & 0.43 & 274M & 5.11& - & 0.72 & 0.49 \\
&&&&&&&&&\\
\textbf{FFHQ $256^{2}$} &&&&&& &&&\\
\midrule

StyleGAN2~\cite{DBLP:journals/corr/abs-1912-04958}& 32.13$^\ddagger$ & - & 32.13$^\dagger$  & - &  59M & 3.8 & - & - & - \\
\emph{LDM-4} (ours, 200 steps, 635K)& 26 & - & 26 & 1.07 & 274M & 4.98& - & 0.73 & 0.50 \\
&&&&&&&&&\\
\textbf{ImageNet $256^{2}$} &&&&&&&&& \\
\midrule
VQGAN-f-4 (ours, first stage)& 29 & - & 29 & - & 55M & 0.58$^{\dagger\dagger}$& - & - & - \\
VQGAN-f-8 (ours, first stage) & 66 & - & 66 & - & 68M & 1.14$^{\dagger\dagger}$& - & - & - \\
\midrule
BigGAN-deep~\cite{bigganbrock}$^\dagger$& 128-256 & & 128-256 & - & 340M & 6.95 & 203.6\tiny$\pm\text{2.6}$ & 0.87 & 0.28\\
ADM~\cite{DBLP:journals/corr/abs-2105-05233} (250 steps) $^\dagger$& 916 & - & 916 & 0.12 & 554M & 10.94& 100.98 & 0.69 & 0.63 \\
ADM-G~\cite{DBLP:journals/corr/abs-2105-05233} (25 steps) $^\dagger$& 916 & 46 & 962 & 0.7 & 608M & 5.58 & - & 0.81  & 0.49 \\
ADM-G~\cite{DBLP:journals/corr/abs-2105-05233} (250 steps)$^\dagger$& 916 & 46 & 962 & 0.07 & 608M & 4.59 & 186.7 & 0.82 & 0.52 \\
ADM-G,ADM-U~\cite{DBLP:journals/corr/abs-2105-05233} (250 steps)$^\dagger$& 329 & 30 & 349 & n/a & n/a & 3.85  & 221.72 & 0.84 & 0.53 \\
\emph{LDM-8-G} (ours, 100, 2.9M)& 79 & 12 & 91 & 1.93 & 506M & 8.11 & 190.4\tiny$\pm\text{2.6}$  & 0.83 & 0.36 \\
\emph{LDM-8} (ours, 200 ddim steps 2.9M, batch size 64)& 79 & - & 79 & 1.9 & 395M & 17.41 & 72.92 & 0.65 & 0.62 \\
\emph{LDM-4} (ours, 250 ddim steps 178K, batch size 1200)& 271 & - & 271 & 0.7 & 400M & 10.56 & 103.49\tiny$\pm\text{1.24}$  & 0.71  & 0.62 \\
\emph{LDM-4-G} (ours, 250 ddim steps 178K, batch size 1200, classifier-free guidance~\cite{ho2021classifier} scale 1.25)& 271 & - & 271 & 0.4 & 400M & 3.95 & 178.22\tiny$\pm\text{2.43}$  & 0.81  & 0.55 \\
\emph{LDM-4-G} (ours, 250 ddim steps 178K, batch size 1200, classifier-free guidance~\cite{ho2021classifier} scale 1.5)& 271 & - & 271 & 0.4 & 400M & 3.60 & 247.67\tiny$\pm\text{5.59}$  & 0.87  & 0.48 \\

\bottomrule
\end{tabular}

\end{adjustbox}
\end{center}
\caption{\label{tab:compute_vs_fid} Comparing compute requirements during training and inference throughput with state-of-the-art generative models. Compute during training in V100-days, numbers of competing methods taken from~\cite{DBLP:journals/corr/abs-2105-05233} unless stated differently;$^*$: Throughput measured in samples/sec on a single NVIDIA A100;$^\dagger$: Numbers taken from~\cite{DBLP:journals/corr/abs-2105-05233} ;$^\ddagger$: Assumed to be trained on 25M train examples; $^{\dagger\dagger}$: R-FID vs. ImageNet validation set}
\end{table}

}

\newcommand{\cinhyperparams}{
\begin{table}[thbp]
\begin{center}
\begin{adjustbox}{max width=.8\textwidth}

\begin{tabular}{lcccccc}
\toprule
& \emph{LDM-1}& \emph{LDM-2} & \emph{LDM-4} & \emph{LDM-8}& \emph{LDM-16} & \emph{LDM-32} \\
\midrule
$z$-shape & $256 \times 256 \times 3$ &$128  \times 128 \times 2$ &$64 \times 64 \times 3$ & $32 \times 32 \times 4$&$16 \times 16 \times 8$ & $88 \times 8 \times 32$\\
$\vert \mathcal{Z} \vert$ & - & 2048 & 8192 & 16384 & 16384 & 16384 \\
Diffusion steps &1000 & 1000 &1000&1000&1000&1000 \\
Noise Schedule & linear&linear&linear&linear&linear&linear \\
Model Size &396M&391M&391M&395M&395M&395M \\
Channels & 192 & 192 & 192 & 256 & 256 & 256\\
Depth &2& 2&2&2&2& 2\\
Channel Multiplier & 1,1,2,2,4,4 & 1,2,2,4,4 & 1,2,3,5 & 1,2,4 & 1,2,4 & 1,2,4 \\
Number of Heads & 1 & 1 & 1 & 1 & 1 & 1 \\
Batch Size &7& 9 & 40 & 64 & 112 & 112 \\
Iterations &2M& 2M & 2M & 2M & 2M & 2M\\
Learning Rate& $\text{4.9e-5}$ & $\text{6.3e-5}$ & $\text{8e-5}$ & $\text{6.4e-5}$ & $\text{4.5e-5}$ & $\text{4.5e-5}$ \\
\midrule
Conditioning & CA & CA &CA &CA&CA& CA\\
CA-resolutions& 32, 16, 8 & 32, 16, 8 & 32, 16, 8 & 32, 16, 8 & 16, 8, 4 & 8, 4, 2\\
Embedding Dimension &  512&512&512&512&512&512 \\
Transformers Depth & 1 & 1&1&1&1&1 \\
\bottomrule
\end{tabular}

\end{adjustbox}
\end{center}
\caption{\label{tab:cin_hyperparams} Hyperparameters for the conditional \emph{LDMs} trained on the ImageNet dataset for the analysis in Sec.~\ref{subsec:reduced_compute}. All models trained on a single NVIDIA A100.}
\end{table}
}

\newcommand{\transformertable}{
\begin{table}[thbp]
\begin{center}
\begin{footnotesize}
\begin{adjustbox}{max width=.45\textwidth}
\begin{tabular}{l c}
\toprule
\textbf{input} & $\R^{h \times w \times c}$ \\
\midrule
LayerNorm  & $\R^{h \times w \times c}$ \\
Conv1x1  & $\R^{h \times w \times d\cdot n_h}$ \\
Reshape & $ \R^{h\cdot w \times d\cdot n_h}$ \\
\multirow{3}{*}{ $\times T \begin{cases*} \text{SelfAttention} \\ \text{MLP} \\ \text{CrossAttention} \end{cases*}$} & $\R^{h\cdot w \times d \cdot n_h}$ \\
 & $\R^{h \cdot w \times d\cdot n_h}$ \\
 & $\R^{h \cdot w \times d\cdot n_h}$ \vspace{0.5em} \\
Reshape & $\R^{h \times w \times d\cdot n_h}$ \\
Conv1x1  & $\R^{h \times w \times c}$ \\
\bottomrule
\end{tabular}
\end{adjustbox}
\end{footnotesize}
\end{center}
\caption{\label{tab:transformertable} Architecture of a transformer block as described in Sec.~\ref{suppsubsubsec:transformer}, 
replacing the self-attention layer of the standard ``ablated UNet'' architecture~\cite{DBLP:journals/corr/abs-2105-05233}. 
Here, $n_h$ denotes the number of attention heads
and $d$ the dimensionality per head.}
\end{table}
}

\newcommand{\transformerhyperparams}{
\begin{table}[thbp]
\begin{center}
\begin{adjustbox}{max width=.8\textwidth}

\begin{tabular}{lcc}
\toprule
& \textbf{Text-to-Image} &  \textbf{Layout-to-Image}\\
\midrule
seq-length & 77 & 92\\
depth $N$ & 32 & 16\\
dim & 1280 & 512\\
\bottomrule
\end{tabular}
\end{adjustbox}
\end{center}
\caption{\label{tab:transformerhyperparams} Hyperparameters for the experiments with transformer encoders in Sec.~\ref{subsec:conditionallatentdiffusion}.}
\end{table}
}

\newcommand{\celebahyperparams}{
\begin{table}[thbp]
\begin{center}
\begin{adjustbox}{max width=.8\textwidth}

\begin{tabular}{lcccccc}
\toprule
& \emph{LDM-1}& \emph{LDM-2} & \emph{LDM-4} & \emph{LDM-8}& \emph{LDM-16} & \emph{LDM-32} \\
\midrule
$z$-shape & $256 \times 256 \times 3$ &$128  \times 128 \times 2$ &$64 \times 64 \times 3$ & $32 \times 32 \times 4$&$16 \times 16 \times 8$ & $88 \times 8 \times 32$\\
$\vert \mathcal{Z} \vert$ & - & 2048 & 8192 & 16384 & 16384 & 16384 \\
Diffusion steps &1000 & 1000 &1000&1000&1000&1000 \\
Noise Schedule & linear&linear&linear&linear&linear&linear \\
Model Size &270M&265M&274M&258M&260M&258M \\
Channels & 192 & 192 & 224 & 256 & 256 & 256\\
Depth &2& 2&2&2&2& 2\\
Channel Multiplier & 1,1,2,2,4,4 & 1,2,2,4,4 & 1,2,3,4 & 1,2,4 & 1,2,4 & 1,2,4 \\
Attention resolutions& 32, 16, 8 & 32, 16, 8 & 32, 16, 8 & 32, 16, 8 & 16, 8, 4 & 8, 4, 2\\
Head Channels & 32 & 32 & 32 & 32 & 32 & 32 \\
Batch Size &9& 11 & 48 & 96 & 128 & 128\\
Iterations$^*$ & 500k & 500k & 500k& 500k & 500k & 500k\\
Learning Rate& $\text{9e-5}$ & $\text{1.1e-4}$ & $\text{9.6e-5}$ & $\text{9.6e-5}$ & $\text{1.3e-4}$ & $\text{1.3e-4}$ \\
\bottomrule
\end{tabular}

\end{adjustbox}
\end{center}
\caption{\label{tab:celeba_hyperparams} Hyperparameters for the unconditional \emph{LDMs} trained on the CelebA dataset for the analysis in Fig.~\ref{fig:speedplot}. All models trained on a single NVIDIA A100. $^*$: All models are trained for 500k iterations. If converging earlier, we used the best checkpoint for assessing the provided FID scores.}
\end{table}
}

\newcommand{\uncondhyperparams}{
\begin{table}[thbp]
\begin{center}
\begin{adjustbox}{max width=.8\textwidth}

\begin{tabular}{lcccc}
\toprule
& CelebA-HQ $256 \times 256$ & FFHQ $256 \times 256$ & LSUN-Churches $256 \times 256$ & LSUN-Bedrooms $256 \times 256$  \\
\midrule
$f$ & 4 & 4 & 8 & 4 \\
$z$-shape & $64 \times 64 \times 3$ & $64 \times 64 \times 3$ &  - & $64 \times 64 \times 3$\\
$\vert \mathcal{Z} \vert$ & 8192 & 8192 & - & 8192  \\
Diffusion steps &1000 & 1000 &1000&1000 \\
Noise Schedule & linear&linear&linear&linear \\
$N_{\text{params}}$ & 274M & 274M & 294M & 274M \\
Channels & 224 & 224 & 192 & 224 \\
Depth &2& 2&2&2 \\
Channel Multiplier & 1,2,3,4 & 1,2,3,4 & 1,2,2,4,4 & 1,2,3,4 \\
Attention resolutions & 32, 16, 8 & 32, 16, 8 & 32, 16, 8, 4 & 32, 16, 8 \\
Head Channels & 32 & 32 & 24 & 32 \\
Batch Size & 48 & 42 & 96 & 48 \\
Iterations$^*$ & 410k & 635k & 500k & 1.9M \\
Learning Rate& $\text{9.6e-5}$ & $\text{8.4e-5}$ & $\text{5.e-5}$ & $\text{9.6e-5}$\\
\bottomrule
\end{tabular}

\end{adjustbox}
\end{center}
\caption{\label{tab:uncond_hyperparams} Hyperparameters for the unconditional \emph{LDMs} producing the numbers shown in Tab.~\ref{tab:fids}. All models trained on a single NVIDIA A100.}
\end{table}
}

\newcommand{\condhyperparams}{
\begin{table}[thbp]
\begin{center}
\begin{adjustbox}{max width=.95\textwidth}

\begin{tabular}{lccccccc}
\toprule
\textbf{Task}& Text-to-Image & \multicolumn{2}{c}{Layout-to-Image} & Class-Label-to-Image &  Super Resolution & Inpainting & Semantic-Map-to-Image\\
\midrule
\textbf{Dataset} & LAION & OpenImages & COCO & ImageNet & ImageNet & Places  & Landscapes\\
\midrule
$f$ & 8 & 4 & 8 & 4 & 4 & 4 & 8\\
$z$-shape & $32 \times 32 \times 4$ &$64 \times 64 \times 3$ & $32 \times 32 \times 4$ & $64 \times 64\times 3$&$64 \times 64 \times 3$ & $64 \times 64 \times 3$ & $32 \times 32 \times 4$ \\
$\vert \mathcal{Z} \vert$ & - & 8192 & 16384 & 8192 & 8192 & 8192 & 16384 \\
Diffusion steps &1000 & 1000 &1000&1000&1000&1000 &1000 \\
Noise Schedule & linear&linear&linear&linear&linear&linear&linear \\
Model Size & 1.45B & 306M & 345M &395M& 169M & 215M & 215M \\
Channels & 320 & 128 & 192 & 192 & 160 & 128 & 128 \\
Depth &2& 2&2&2&2& 2& 2\\
Channel Multiplier & 1,2,4,4 & 1,2,3,4 & 1,2,4 & 1,2,3,5 & 1,2,2,4 & 1,4,8 & 1,4,8\\
Number of Heads & 8 & 1 & 1 & 1 & 1 & 1 & 1 \\
Dropout & - & -  & 0.1  & - & - & - & - \\
Batch Size & 680 & 24 & 48 & 1200 & 64 & 128 & 48 \\
Iterations & 390K & 4.4M & 170K & 178K & 860K & 360K & 360K \\
Learning Rate& $\text{1.0e-4}$ & $\text{4.8e-5}$  & $\text{4.8e-5}$ & $\text{1.0e-4}$ & $\text{6.4e-5}$ &$\text{1.0e-6}$ &$\text{4.8e-5}$ \\
\midrule
Conditioning & CA & CA & CA & CA & concat & concat & concat\\
(C)A-resolutions& 32, 16, 8 & 32, 16, 8 & 32, 16, 8 & 32, 16, 8 & - & - & - \\
Embedding Dimension & 1280 & 512 & 512 & 512 & - & - & - \\
Transformer Depth & 1 & 3 & 2 & 1 & - & - & - \\
\bottomrule
\end{tabular}

\end{adjustbox}
\end{center}
\caption{\label{tab:cond_hyperparams} Hyperparameters for the conditional
  \emph{LDMs} from Sec.~\ref{sec:experiments}. All models trained on a single NVIDIA A100
  except for the inpainting model which was trained on eight V100.}
\end{table}
}

\newcommand{\cinmetrics}{
\begin{table}[htbp]
\centering
\begin{adjustbox}{max width=\textwidth}
\footnotesize
\begin{tabular}{lcccccc}
\toprule
\textbf{Method} & FID$\downarrow$ & IS$\uparrow$ & Precision$\uparrow$ & Recall$\uparrow$ & $N_{\text{params}}$ & \\
\midrule
SR3~\cite{DBLP:journals/corr/abs-2104-07636} & 11.30 & - & - & - & 625M & - \\
ImageBART~\cite{DBLP:journals/corr/abs-2108-08827}& 21.19 & -& - & -  & 3.5B & - \\
ImageBART~\cite{DBLP:journals/corr/abs-2108-08827}& 7.44 & - & - & - & 3.5B & 0.05 acc. rate$^*$ \\
VQGAN+T~\cite{DBLP:journals/corr/abs-2012-09841}& 17.04 & 70.6\tiny$\pm\text{1.8}$& - & -  & 1.3B & - \\
VQGAN+T~\cite{DBLP:journals/corr/abs-2012-09841}& 5.88 & \textbf{304.8}\tiny$\pm\text{3.6}$& - & -  & 1.3B & 0.05 acc. rate$^*$ \\
BigGan-deep~\cite{bigganbrock}& 6.95 & 203.6\tiny$\pm\text{2.6}$& \textbf{0.87} & 0.28  & 340M & - \\
ADM~\cite{DBLP:journals/corr/abs-2105-05233} & 10.94 & 100.98& 0.69 & \textbf{0.63}  & 554M & 250 DDIM steps\\
ADM-G~\cite{DBLP:journals/corr/abs-2105-05233} & 4.59  & 186.7& 0.82 & 0.52 & 608M & 250 DDIM steps \\
ADM-G,ADM-U~\cite{DBLP:journals/corr/abs-2105-05233} & \underline{3.85}  & 221.72 & 0.84 & 0.53 & n/a & 2 $\times$ 250 DDIM steps \\
CDM~\cite{DBLP:journals/corr/abs-2106-15282}& 4.88 & 158.71\tiny$\pm\text{2.26}$ & - & - & n/a & 2 $\times$ 100 DDIM steps \\
\midrule
\emph{LDM-8} (ours) & 17.41 & 72.92\tiny$\pm\text{2.6}$& 0.65 & \underline{0.62}  & 395M & 200 DDIM steps, 2.9M train steps, batch size 64 \\
\emph{LDM-8-G} (ours) & 8.11 & 190.43\tiny$\pm\text{2.60}$& 0.83  & 0.36 & 506M& 200 DDIM steps, classifier scale 10, 2.9M train steps, batch size 64  \\
\emph{LDM-8} (ours) & 15.51 & 79.03\tiny$\pm\text{1.03}$& 0.65 & \textbf{0.63}  & 395M & 200 DDIM steps, 4.8M train steps, batch size 64 \\
\emph{LDM-8-G} (ours) & 7.76 & 209.52\tiny$\pm\text{4.24}$& \underline{0.84} & 0.35 & 506M& 200 DDIM steps, classifier scale 10, 4.8M train steps, batch size 64 \\
\emph{LDM-4} (ours) & 10.56 & 103.49\tiny$\pm\text{1.24}$   & 0.71  & \underline{0.62} & 400M & 250 DDIM steps, 178K train steps, batch size 1200 \\
\emph{LDM-4-G} (ours) & 3.95 & 178.22\tiny$\pm\text{2.43}$  & 0.81  & 0.55 & 400M& 250 DDIM steps, unconditional guidance~\cite{ho2021classifier} scale 1.25, 178K train steps, batch size 1200 \\
\emph{LDM-4-G} (ours) & \textbf{3.60} & \underline{247.67\tiny$\pm\text{5.59}$}  & \textbf{0.87}  & 0.48 & 400M & 250 DDIM steps, unconditional guidance~\cite{ho2021classifier} scale 1.5, 178K train steps, batch size 1200 \\
\bottomrule
\end{tabular}
\end{adjustbox}

\caption{\label{tab:imagenet_numbers} Comparison of a class-conditional ImageNet \emph{LDM} with recent state-of-the-art methods for class-conditional image generation on the ImageNet~\cite{DBLP:conf/cvpr/DengDSLL009} dataset.$^*$: Classifier rejection sampling with the given rejection rate as proposed in~\cite{DBLP:conf/nips/RazaviOV19}.}

\end{table}
}

\newcommand{\layouttoimagefids}{
\begin{table}[htbp]
\centering
\begin{adjustbox}{max width=.7\textwidth}
\footnotesize
\begin{tabular}{lccc}
\toprule
  & COCO$256 \times 256$  & OpenImages $256 \times 256$ & OpenImages $512\times 512$\\
  \cmidrule{2-4}
\textbf{Method} & FID$\downarrow$ & FID$\downarrow$ & FID$\downarrow$ \\
\midrule
LostGAN-V2~\cite{DBLP:journals/corr/abs-2003-11571}& 42.55 & - & - \\
OC-GAN~\cite{DBLP:conf/aaai/SylvainZBH021}& 41.65 & - & - \\
SPADE~\cite{Park_2019_CVPR}& \underline{41.11} & -& -  \\
VQGAN+T~\cite{DBLP:journals/corr/abs-2105-06458}& 56.58 & \underline{45.33} & \underline{48.11} \\
\midrule
\emph{LDM-8} (100 steps, ours) & 42.06$^\dagger$ & -& - \\
\emph{LDM-4} (200 steps, ours) & \textbf{40.91}$^*$ & \textbf{32.02}& \textbf{35.80} \\
\bottomrule
\end{tabular}
\end{adjustbox}

\caption{\label{tab:layout2img} Quantitative comparison of our layout-to-image models on the COCO~\cite{DBLP:conf/cvpr/CaesarUF18} and OpenImages~\cite{DBLP:journals/corr/abs-1811-00982} datasets. $^\dagger$: Training from scratch on COCO; $^*$: Finetuning from OpenImages.}

\end{table}
}

\begin{document}

\title{\vspace{-2.5em} High-Resolution Image Synthesis with Latent Diffusion Models}  %

\author{%
  Robin Rombach$^1$ \thanks{The first two authors contributed equally to this
  work.} \qquad Andreas Blattmann$^1$ $^*$\qquad Dominik Lorenz$^1$ \qquad
  Patrick Esser\textsuperscript{\,\includegraphics[width=0.70em]{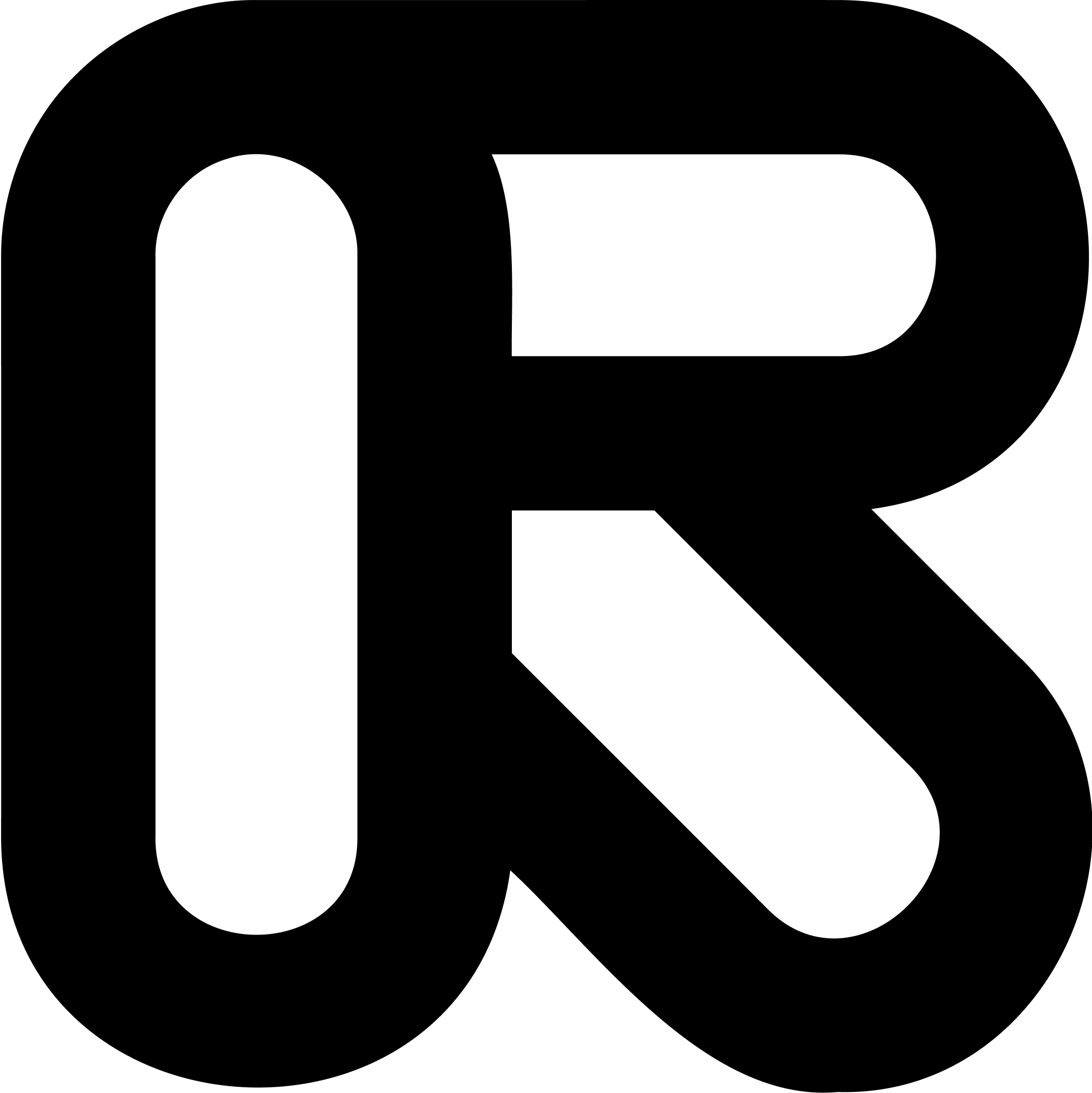}} \qquad Bj\"orn Ommer$^1$ \\
  \small{$^1$\href{https://ommer-lab.com/}{Ludwig Maximilian University of Munich \& IWR, Heidelberg
  University, Germany} \qquad \textsuperscript{\includegraphics[width=0.87em]{runway}}\href{https://runwayml.com/}{Runway ML}}\\
  \url{\projectpath} 
}
\maketitle

\begin{abstract}
By decomposing the image formation process into a sequential application of %
denoising autoencoders, diffusion models (DMs) achieve state-of-the-art synthesis results on image data and beyond.
Additionally, their formulation allows for a guiding mechanism to control the image generation process without retraining.
However, since these models typically operate directly in pixel space, optimization of powerful DMs often consumes hundreds of GPU days and inference is expensive due to 
sequential evaluations.
To enable DM training on limited computational resources while retaining their quality and flexibility, 
we apply them in the latent space of powerful pretrained autoencoders.
In contrast to previous work, training diffusion models on such a representation allows for the first time to reach a near-optimal point 
between complexity reduction and detail preservation,
greatly boosting visual fidelity.
By introducing cross-attention layers into the model architecture, 
we turn diffusion models into powerful and flexible 
generators %
for general conditioning inputs such as text or bounding boxes and high-resolution synthesis becomes possible in a convolutional manner.
Our latent diffusion models (LDMs) achieve new state-of-the-art scores for image
inpainting and class-conditional image synthesis and highly competitive performance on various tasks, including
text-to-image synthesis, unconditional image generation and super-resolution, while 
significantly reducing computational requirements compared to pixel-based DMs.

\end{abstract}

\vspace{-1.5em}
\section{Introduction}
\vspace{-0.5em}
\label{sec:intro}
\enlargethispage{\baselineskip}
\firststagecomparison
Image synthesis is one of the computer vision fields with the most spectacular recent development, but also among those with the greatest computational demands. Especially high-resolution synthesis of complex, natural scenes is presently dominated by %
scaling up likelihood-based models, potentially containing billions of parameters in autoregressive (AR) transformers~\cite{DBLP:journals/corr/abs-2102-12092, DBLP:conf/nips/RazaviOV19}.
In contrast, the promising results of GANs~\cite{goodfellow2014GAN, bigganbrock, karras2019stylebased} have been revealed to be mostly confined to data with comparably limited variability as their adversarial learning procedure does not easily scale to modeling complex, multi-modal distributions. %
Recently, diffusion models~\cite{DBLP:journals/corr/Sohl-DicksteinW15}, which are built from a hierarchy of denoising autoencoders, have shown to achieve impressive results in image synthesis~\cite{DBLP:conf/nips/HoJA20, DBLP:journals/corr/abs-2011-13456} and beyond~\cite{DBLP:journals/corr/abs-2107-00630, DBLP:conf/iclr/ChenZZWNC21, DBLP:conf/iclr/KongPHZC21, DBLP:journals/corr/abs-2103-16091}, and define the state-of-the-art in class-conditional image synthesis \cite{DBLP:journals/corr/abs-2105-05233, DBLP:journals/corr/abs-2106-15282} and super-resolution \cite{DBLP:journals/corr/abs-2104-07636}. Moreover, even unconditional DMs can readily be applied to tasks such as inpainting and colorization \cite{DBLP:journals/corr/abs-2011-13456} or stroke-based synthesis \cite{DBLP:journals/corr/abs-2108-01073}, in contrast to other types of generative models~\cite{VAE,VAE2,DBLP:conf/iclr/DinhSB17}.
Being likelihood-based models, they do not exhibit mode-collapse and training
instabilities as GANs and, by heavily exploiting parameter sharing,
they can model highly complex distributions of natural images without involving
billions of parameters as in AR
models~\cite{DBLP:conf/nips/RazaviOV19}. %
\vspace{-1.5em}
\paragraph{Democratizing High-Resolution Image Synthesis}
\enlargethispage{\baselineskip}
DMs belong to the class of likelihood-based models, whose mode-covering behavior makes them prone to spend excessive amounts of capacity (and thus compute resources) on modeling imperceptible details of the data \cite{dieleman2020typicality, DBLP:journals/corr/SalimansKCK17}. 
Although the reweighted variational objective \cite{DBLP:conf/nips/HoJA20} 
aims to address this
by undersampling the initial denoising steps, 
DMs are still computationally demanding, since training and evaluating such a model requires repeated function evaluations (and gradient computations) in the high-dimensional space of RGB images.
As an example, training the most powerful DMs often takes hundreds of GPU days (\eg 150 - 1000 V100 days in \cite{DBLP:journals/corr/abs-2105-05233})
and repeated evaluations on a noisy version of the input space render also inference expensive, so that producing 50k samples %
takes approximately 5 days \cite{DBLP:journals/corr/abs-2105-05233} %
on a single A100 GPU. 
This has two consequences for the research community and users in general: 
Firstly, training such a model 
requires massive computational resources only available to a small fraction of the field,
and leaves a huge carbon footprint \cite{DBLP:journals/corr/abs-2104-10350,DBLP:conf/aaai/StrubellGM20}. 
Secondly, evaluating an already trained model is also expensive in time and memory, since the same model architecture must
run sequentially for a large number of steps (\eg 25 - 1000 steps in \cite{DBLP:journals/corr/abs-2105-05233}).

To increase the accessibility of this powerful model class and at the same time
    reduce its significant resource consumption, a method is needed that
    reduces the computational complexity for both training and sampling.
Reducing the computational demands of DMs without impairing their performance is, therefore, key to enhance their accessibility.

\vspace{-1.5em}
\paragraph{Departure to Latent Space}
Our approach starts with the analysis of already trained diffusion models in pixel space: 
Fig.~\ref{fig:perceptualcompression} shows the rate-distortion trade-off of a trained model. %
As with any likelihood-based model, learning can be roughly divided into two stages: First is a \emph{perceptual compression} stage which removes high-frequency details 
but %
still learns little semantic variation. 
In the second stage, the actual generative model learns the semantic and conceptual composition of the data (\emph{semantic compression}).
We thus aim to first find a \textsl{perceptually equivalent, but computationally more suitable space}, in which we will train diffusion models for high-resolution image synthesis. 

Following common practice \cite{DBLP:conf/nips/OordVK17, DBLP:conf/nips/RazaviOV19, DBLP:journals/corr/abs-2012-09841, DBLP:conf/iclr/DaiW19, DBLP:journals/corr/abs-2102-12092}, 
we separate training into two distinct phases: First, we train an autoencoder %
which provides a lower-dimensional (and thereby efficient) representational space which is perceptually equivalent to the data space.
Importantly, and in contrast to previous work \cite{DBLP:journals/corr/abs-2012-09841, DBLP:journals/corr/abs-2102-12092}, 
we do not need to 
rely on excessive spatial compression, as we train DMs in the learned latent space, which
exhibits better scaling properties with respect to the spatial dimensionality. 
The reduced complexity also provides efficient image generation from the latent space with a single network pass. 
    We dub the resulting model class \emph{Latent Diffusion Models} (LDMs).

A notable
 advantage of this approach is that we need to train the universal
    autoencoding stage only once and can therefore reuse it for multiple DM
    trainings or to explore possibly completely different tasks
    \cite{clipguiding}.
This enables efficient exploration of a large number of diffusion models for various image-to-image and text-to-image tasks. 
For the latter, we design an architecture that connects transformers to the DM's UNet backbone \cite{DBLP:conf/miccai/RonnebergerFB15} 
and enables arbitrary types of token-based conditioning mechanisms, see Sec.~\ref{subsec:conditioning}.
\perceptualcompression
\enlargethispage{\baselineskip}
In sum, our work makes the following \textbf{contributions}:

(i) In contrast to purely transformer-based approaches \cite{DBLP:journals/corr/abs-2012-09841, DBLP:journals/corr/abs-2102-12092}, our 
method scales more graceful to higher dimensional data and can thus (a) work on a compression level which provides more faithful and detailed reconstructions than previous work (see Fig.~\ref{fig:firststagecomparison}) and (b) can be efficiently applied to high-resolution synthesis of megapixel images.

(ii) We achieve competitive performance on multiple tasks (unconditional image synthesis, inpainting, stochastic super-resolution)
and datasets while significantly lowering computational costs.
Compared to pixel-based diffusion approaches, we also significantly decrease inference costs.

(iii) We show that, in contrast to previous work \cite{DBLP:journals/corr/abs-2106-05931} which learns both an encoder/decoder architecture and a score-based prior simultaneously, our
approach does not require a delicate weighting of reconstruction and generative abilities. 
This ensures extremely faithful reconstructions and requires very little regularization of the latent space.

(iv) We find that for densely conditioned tasks such as super-resolution, inpainting and semantic synthesis, our model can be applied in a 
convolutional fashion and render large, consistent images of $\sim 1024^2$ px.

(v) Moreover, %
we design a general-purpose conditioning mechanism based on cross-attention, enabling multi-modal training. 
We use it to train class-conditional, text-to-image and layout-to-image models.

(vi) Finally, we release pretrained latent diffusion and autoencoding models at \url{\projectpath} 
which might be reusable for a various tasks besides 
training of DMs \cite{clipguiding}. 

\section{Related Work}
\enlargethispage{\baselineskip}
\vspace{-0.75em}
\textbf{Generative Models for Image Synthesis}
The high dimensional nature of images presents distinct challenges to generative modeling.
Generative Adversarial Networks (GAN) \cite{goodfellow2014GAN}
allow for efficient sampling of high resolution images with good perceptual quality \cite{bigganbrock, DBLP:journals/corr/abs-1912-04958}, but are difficult to optimize \cite{DBLP:journals/corr/abs-1801-04406,arjovsky2017wasserstein, gulrajani2017improved} and struggle to capture the full data distribution 
\cite{DBLP:conf/iclr/MetzPPS17}. 
In contrast, likelihood-based methods emphasize good density estimation which renders optimization more well-behaved.
Variational autoencoders (VAE) \cite{VAE}
and flow-based models \cite{dinh2015nice, DBLP:conf/iclr/DinhSB17} enable efficient synthesis of high resolution images \cite{DBLP:journals/corr/abs-2011-10650, DBLP:conf/nips/VahdatK20, glow}, but sample quality is not on par with GANs.
While autoregressive models (ARM) \cite{DBLP:journals/corr/OordKK16,
NIPS2016_b1301141, DBLP:conf/icml/ChenRC0JLS20,
DBLP:journals/corr/abs-1904-10509} achieve strong performance in density
estimation, computationally demanding architectures
\cite{DBLP:conf/nips/VaswaniSPUJGKP17} and a sequential sampling process limit them to low resolution images.
Because pixel based representations of images contain barely
perceptible, high-frequency details \cite{dieleman2020typicality,
DBLP:journals/corr/SalimansKCK17}, maximum-likelihood training spends a
disproportionate amount of capacity on modeling them, resulting in
long training times.
To scale to higher resolutions,
several two-stage approaches \cite{DBLP:journals/corr/abs-2104-10157, DBLP:conf/nips/RazaviOV19, DBLP:journals/corr/abs-2012-09841, yu2021vectorquantized}
use ARMs to model a compressed latent image space instead of raw pixels.

Recently, \textbf{Diffusion Probabilistic Models} (DM)~\cite{DBLP:journals/corr/Sohl-DicksteinW15}, have achieved state-of-the-art results in density estimation \cite{DBLP:journals/corr/abs-2107-00630} as well as in sample quality \cite{DBLP:journals/corr/abs-2105-05233}. The generative power of these models stems from a natural fit to the inductive biases of image-like data when their underlying neural backbone is implemented as a UNet~\cite{DBLP:conf/miccai/RonnebergerFB15, DBLP:conf/nips/HoJA20, DBLP:journals/corr/abs-2011-13456, DBLP:journals/corr/abs-2105-05233}.
The best synthesis quality is usually achieved when a reweighted objective \cite{DBLP:conf/nips/HoJA20}
is used for training. In this case, the DM corresponds to a lossy compressor and allow to trade image quality for compression capabilities.
Evaluating and optimizing these models in pixel space, however, has the downside of low inference speed and very high training costs.
While the former can be partially adressed by advanced sampling strategies~\cite{DBLP:conf/iclr/SongME21, DBLP:journals/corr/abs-2104-02600, DBLP:journals/corr/abs-2106-00132} and hierarchical approaches~\cite{DBLP:journals/corr/abs-2106-15282, DBLP:journals/corr/abs-2106-05931}, training on high-resolution image data always requires to calculate expensive gradients.
 We adress both drawbacks with our proposed \emph{LDMs}, which %
 work on a compressed latent space of lower dimensionality.
 This renders training computationally cheaper and speeds up inference with
 almost no reduction in synthesis quality (see
 Fig.~\ref{fig:firststagecomparison}).

\textbf{Two-Stage Image Synthesis}
\enlargethispage{\baselineskip}
To mitigate the shortcomings of individual generative approaches, a lot of research \cite{DBLP:conf/iclr/DaiW19,  DBLP:conf/nips/RombachEO20, DBLP:journals/corr/abs-2012-09841, yu2021vectorquantized, DBLP:journals/corr/abs-2104-10157, DBLP:conf/nips/RazaviOV19} has gone into combining the strengths of different methods into more efficient and performant models via a two stage approach. VQ-VAEs \cite{DBLP:journals/corr/abs-2104-10157, DBLP:conf/nips/RazaviOV19} use autoregressive models to learn an expressive prior over a discretized latent space.
\cite{DBLP:journals/corr/abs-2102-12092} extend this approach to text-to-image generation by learning a joint distributation over discretized image and text representations.
More generally, \cite{DBLP:conf/nips/RombachEO20} uses conditionally invertible networks to provide a generic transfer between latent spaces of diverse domains.
Different from VQ-VAEs, VQGANs \cite{DBLP:journals/corr/abs-2012-09841, yu2021vectorquantized} employ a first stage with an adversarial and perceptual objective to scale autoregressive transformers to larger images.
However, the high compression rates required for feasible ARM training, which introduces billions of trainable parameters~\cite{DBLP:journals/corr/abs-2102-12092,DBLP:journals/corr/abs-2012-09841}, limit the overall performance of such approaches and less compression comes at the price of high computational cost~\cite{DBLP:journals/corr/abs-2102-12092,DBLP:journals/corr/abs-2012-09841}.
Our work prevents such trade-offs, as our proposed \emph{LDMs} scale more gently to higher dimensional latent spaces due to their convolutional backbone. %
Thus, we are free to choose the level of compression which optimally mediates between learning a powerful first stage, without leaving too much perceptual compression up to the generative diffusion model while guaranteeing high-fidelity reconstructions (see Fig.~\ref{fig:firststagecomparison}).

While approaches to jointly \cite{DBLP:journals/corr/abs-2106-05931} or separately \cite{DBLP:journals/corr/abs-2106-06819} 
learn an encoding/decoding model together with a score-based prior exist,
the former still require a difficult weighting between reconstruction and generative capabilities \cite{DBLP:conf/iclr/DaiW19} and are outperformed by our approach (Sec.~\ref{sec:experiments}), and the latter focus on highly structured images such as human faces.

\section{Method}
\enlargethispage{\baselineskip}
\label{sec:method}
\vspace{-0.5em}
To lower the computational demands of training diffusion models towards high-resolution image synthesis, 
we observe
that although diffusion models 
allow
to ignore perceptually irrelevant details by undersampling the corresponding loss terms \cite{DBLP:conf/nips/HoJA20}, 
they still require costly function evaluations in pixel space,
which causes
huge demands in computation time and energy resources. 

We propose to circumvent this drawback by introducing an explicit separation
 of the compressive from the generative learning phase (see
 Fig.~\ref{fig:perceptualcompression}).
 To achieve this, we utilize an autoencoding model which learns a space that is perceptually equivalent to the image space, 
but offers significantly reduced computational complexity.%

Such an approach offers several advantages: (i)
By leaving the high-dimensional image space, we 
obtain DMs which are computationally much more efficient
because sampling is performed on a low-dimensional space.
(ii)
We exploit the inductive bias of DMs inherited from their UNet architecture
\cite{DBLP:conf/miccai/RonnebergerFB15}, which makes them particularly effective for data with spatial
structure and therefore
alleviates the need for aggressive, quality-reducing compression levels as required by previous
approaches \cite{DBLP:journals/corr/abs-2012-09841,
DBLP:journals/corr/abs-2102-12092}.
(iii)
Finally, we obtain general-purpose compression models whose latent space 
can be used to train multiple generative models
and which can also be utilized
for other downstream applications such as single-image CLIP-guided synthesis \cite{Frans2021CLIPDrawET}.
\subsection{Perceptual Image Compression}
\vspace{-0.5em}
\label{subsec:stageone}
Our perceptual compression model 
is based on previous work \cite{DBLP:journals/corr/abs-2012-09841} and
consists of an autoencoder trained by combination of a perceptual
loss \cite{lpips} and a patch-based \cite{DBLP:conf/cvpr/IsolaZZE17} adversarial objective \cite{dosovitskiy201perceptual, DBLP:journals/corr/abs-2012-09841, yu2021vectorquantized}. 
This ensures that the reconstructions are 
confined to the image manifold by enforcing local realism and avoids bluriness introduced by relying solely on pixel-space losses
such as $L_2$ or $L_1$ objectives.

More precisely, given an image $x \in \mathbb{R}^{\hpixel \times \wpixel \times \cpixel}$ in RGB space, 
the encoder $\encoder$ encodes $x$ into a latent representation $z=\encoder(x)$, and the decoder $\decoder$
reconstructs the image from the latent, giving $\xrec = \decoder(z) = \decoder(\encoder(x))$, where $z \in \mathbb{R}^{\hlatent \times \wlatent \times \clatent}$.
Importantly, the encoder \emph{downsamples} the image by a factor $f = H/h = W/w$, and we investigate different downsampling 
factors $f = 2^m$, with $m \in \mathbb{N}$. %

In order to avoid arbitrarily 
high-variance
latent spaces, we experiment with two different kinds of regularizations. The 
first variant, \emph{KL-reg.}, imposes a slight KL-penalty towards a standard
normal on the learned latent, similar to 
a VAE \cite{VAE,VAE2}, whereas \emph{VQ-reg.}
uses a vector quantization layer \cite{DBLP:conf/nips/OordVK17} within 
the decoder. This model can be interpreted as a VQGAN \cite{DBLP:journals/corr/abs-2012-09841} but with the quantization layer
absorbed by the decoder. %
Because our subsequent DM
is designed to work with the two-dimensional structure of
our learned latent space $z=\encoder(x)$, we can use relatively mild
compression rates and achieve very good reconstructions. This is in contrast to
previous works \cite{DBLP:journals/corr/abs-2012-09841,
DBLP:journals/corr/abs-2102-12092},
which relied on an arbitrary 1D ordering of the learned space $z$ to model its
distribution autoregressively and thereby ignored much of the
inherent structure of $z$.
Hence, our compression model preserves details of $x$ better (see Tab.~\ref{tab:firststagetablecomplete}).
The full objective and training details can be found in the supplement.

\subsection{Latent Diffusion Models}
\label{subsec:stagetwo}
\noindent \textbf{Diffusion Models} \cite{DBLP:journals/corr/Sohl-DicksteinW15} are probabilistic models designed to learn a data distribution $p(x)$ by 
gradually denoising a normally distributed variable, which corresponds to learning
the reverse process of a fixed Markov Chain of length $T$. 
For image synthesis, the most successful models \cite{DBLP:conf/nips/HoJA20, DBLP:journals/corr/abs-2105-05233, DBLP:journals/corr/abs-2104-07636}
rely on a reweighted variant of the variational lower bound on $p(x)$, which mirrors denoising score-matching \cite{DBLP:journals/corr/abs-2011-13456}.
These models can be interpreted as an equally weighted sequence of denoising autoencoders $\model(x_{t},t);\, t=1\dots T$, 
which are trained to predict a denoised variant of their input $x_t$, where $x_t$ is a noisy version of the input $x$.
The corresponding objective can be simplified to (Sec.~\ref{suppsec:dmdetails})
\begin{equation}
\lsimple = \expec_{x, \epsilon \sim \mathcal{N}(0, 1),  t }\Big[ \Vert \epsilon - \model(x_{t},t) \Vert_{2}^{2}\Big] \, ,
\label{eq:dmloss}
\end{equation}
with $t$ uniformly sampled from $\{1, \dots, T\}$.

\noindent \textbf{Generative Modeling of Latent Representations}
With our trained perceptual compression models consisting of $\encoder$ and $\decoder$, 
we now have access to an efficient, low-dimensional latent space in which high-frequency, imperceptible details are 
abstracted away.
Compared to the high-dimensional pixel space, this space is more suitable 
for likelihood-based generative models, as they can now (i) focus on the important, semantic bits of the data
and (ii) train in a lower dimensional, computationally much more efficient space.

Unlike previous work that relied on autoregressive, attention-based 
transformer models in a highly compressed, discrete latent space \cite{DBLP:journals/corr/abs-2102-12092, DBLP:journals/corr/abs-2012-09841, yu2021vectorquantized}, 
we can take advantage of image-specific inductive biases that our model offers. 
This includes the ability to build the underlying UNet primarily from 2D convolutional layers, 
and further focusing the objective on the perceptually most relevant bits using
the reweighted bound, which now reads
\crossattnfig
\begin{equation}
\lsimpleldm := \expec_{\encoder(x), \epsilon \sim \mathcal{N}(0, 1),  t}\Big[ \Vert \epsilon - \model(z_{t},t) \Vert_{2}^{2}\Big] \, .
\label{eq:ldmloss}
\end{equation}
The neural backbone $\model(\circ, t)$ of our model 
is realized as a time-conditional UNet~\cite{DBLP:conf/miccai/RonnebergerFB15}. 
Since the forward process is fixed, $\zt{t}$ can be efficiently obtained from
$\encoder$ during training, 
and samples from $p(z$) can be decoded to image space with a single pass through $\decoder$.
\smallsamples
\subsection{Conditioning Mechanisms}
\label{subsec:conditioning}
\vspace{-0.5em}
Similar to other types of generative
models~\cite{DBLP:journals/corr/MirzaO14,NIPS2015_8d55a249}, diffusion models
are in principle capable of modeling conditional distributions of the form
$p(z \vert \cond)$.
This can be implemented with
a conditional denoising autoencoder $\model(\zt{t},t,\cond)$ and
paves the way to controlling the synthesis process through
inputs $\cond$ such as text~\cite{Reed2016GenerativeAT}, semantic maps ~\cite{spade, DBLP:conf/cvpr/IsolaZZE17} or other image-to-image translation tasks \cite{Isola2017ImagetoImageTW}.
 
 In the context of image synthesis, however, combining the
  generative power of DMs with other types of conditionings beyond class-labels~\cite{DBLP:journals/corr/abs-2105-05233} 
or blurred variants
 of the input image~\cite{DBLP:journals/corr/abs-2104-07636}
  is so far an under-explored area of research.

We turn DMs into more flexible conditional image generators by
augmenting their underlying UNet backbone with the cross-attention mechanism~\cite{DBLP:conf/nips/VaswaniSPUJGKP17}, 
which 
is
effective for learning attention-based models of various input modalities~\cite{DBLP:conf/icml/JaegleGBVZC21,DBLP:journals/corr/abs-2107-14795}. 
To pre-process $y$ from various modalities (such as language prompts) we introduce a domain specific 
encoder $\conditioner$ that projects $y$ to an intermediate representation $\conditioner(y) \in \R^{M\times d_\tau}$, which 
is then mapped to the intermediate layers of the UNet via a cross-attention
layer implementing $\text{Attention}(Q, K, V) = \text{softmax}\left(\frac{QK^T}{\sqrt{d}}\right) \cdot V$, with
\begin{equation*}
Q = W^{(i)}_Q \cdot  \varphi_i(z_t), \; K = W^{(i)}_K \cdot \conditioner(y),
  \; V = W^{(i)}_V \cdot \conditioner(y) . \nonumber
\end{equation*}
Here, $\varphi_i(z_t) \in \R^{N \times d^i_\epsilon}$ denotes a (flattened) intermediate representation of the UNet implementing 
$\model$ and $W^{(i)}_V \in \R^{d \times d^i_\epsilon}$, $W^{(i)}_Q \in \R^{d \times d_\tau} $ \& $W^{(i)}_K \in \R^{d \times d_\tau}$ are learnable projection matrices \cite{DBLP:conf/nips/VaswaniSPUJGKP17, DBLP:conf/icml/JaegleGBVZC21}. See Fig.~\ref{fig:conditioning} for a visual depiction.

Based on image-conditioning pairs, we then learn the conditional LDM via
\begin{equation}
\lsimplelcm := \expec_{\encoder(x), y, \epsilon \sim \mathcal{N}(0, 1), t }\Big[ \Vert \epsilon - \model(z_{t},t, \conditioner(y)) \Vert_{2}^{2}\Big] \, ,
\label{eq:cond_loss}
\end{equation}
where 
both $\conditioner$ and $\model$ are jointly optimized via Eq.~\ref{eq:cond_loss}.
This conditioning mechanism is flexible as %
$\conditioner$ can be parameterized with domain-specific experts, 
\eg (unmasked) transformers~\cite{DBLP:conf/nips/VaswaniSPUJGKP17} when $y$ are
text prompts (see Sec.~\ref{subsubsec:crossattn2img})%

\section{Experiments}
\label{sec:experiments}
\newtexttoimagesamples
\cincompression
\speeeeed
\vspace{-0.5em}
\emph{LDMs} provide means to flexible and computationally tractable diffusion based image synthesis
 of various image modalities, which we empirically show in the following. 
Firstly, however, we analyze the gains of our models compared to pixel-based diffusion models in both training and inference. 
Interestingly, we find that \emph{LDMs} trained in \emph{VQ}-regularized latent spaces sometimes achieve better sample quality, even though the reconstruction capabilities of \emph{VQ}-regularized first stage models slightly fall behind those of their continuous counterparts, \cf Tab.~\ref{tab:firststagetablecomplete}. 
A visual comparison between the effects of first stage regularization schemes on \emph{LDM} training and their generalization abilities to resolutions $>256^2$ can be found in Appendix \ref{suppsec:rescale}. In \ref{suppsec:implementation_details} we list details on architecture, implementation, training and evaluation for all results presented in this section.
\subsection{On Perceptual Compression Tradeoffs}
\label{subsec:reduced_compute}

This section analyzes the behavior of our LDMs with different downsampling
factors $f\in\{1,2,4,8,16,32\}$ (abbreviated as \emph{LDM-}$f$, where
\emph{LDM-1} corresponds to pixel-based DMs).
To obtain a
comparable test-field, we fix the computational resources
to a
single NVIDIA A100 for all experiments in this section and train all models for
the same number of steps and with the same number of parameters.

Tab.~\ref{tab:firststagetablecomplete} shows hyperparameters and reconstruction performance of the first stage models used for the \emph{LDMs} compared in this section.
Fig.~\ref{fig:cin_traincourse} shows sample quality as a function of training
progress for 2M steps of class-conditional models on the
ImageNet~\cite{DBLP:conf/cvpr/DengDSLL009} dataset. We see that, i) small
downsampling factors for \emph{LDM-}$\{$\emph{1,2}$\}$ result in slow training
progress, whereas ii) overly large values of $f$ cause stagnating fidelity
after comparably few training steps. Revisiting the analysis above
(Fig.~\ref{fig:firststagecomparison} and \ref{fig:perceptualcompression}) we
attribute this to i) leaving most of perceptual compression to the diffusion
model and ii) too strong first stage compression resulting in information loss
and thus limiting the achievable quality. \emph{LDM-$\{$4-16$\}$}
strike a good balance between efficiency and perceptually faithful results,
which manifests in a significant FID~\cite{FID} gap of 38 between pixel-based diffusion (\emph{LDM-1}) and \emph{LDM-8} after 2M training steps.

In Fig.~\ref{fig:speedplot}, we compare models trained on
CelebA-HQ~\cite{DBLP:journals/corr/abs-1710-10196} and ImageNet in terms
sampling speed for different numbers of denoising steps with the DDIM
sampler~\cite{DBLP:conf/iclr/SongME21} and plot it against FID-scores~\cite{FID}.
\emph{LDM-$\{$4-8$\}$} outperform models 
with unsuitable ratios of perceptual and conceptual compression. 
Especially compared to
pixel-based \emph{LDM-1}, they achieve much lower FID scores while
simultaneously significantly increasing sample throughput.
Complex datasets such as ImageNet require reduced compression rates to avoid
reducing quality.
In summary, \emph{LDM-4} and \emph{-8} offer the best conditions for achieving high-quality synthesis results.
\fidsnew
\texttoimgquantcocotwo

\vspace{-1.0em}
\subsection{Image Generation with Latent Diffusion}
\label{subsec:uncond2img}
\vspace{-0.5em}
We train unconditional models of $256^2$ images on
CelebA-HQ~\cite{DBLP:journals/corr/abs-1710-10196}, FFHQ~\cite{stylegan},
LSUN-Churches and \mbox{-Bedrooms}~\cite{DBLP:journals/corr/YuZSSX15} 
and evaluate the i) sample quality and ii) their coverage of the data manifold using ii) FID~\cite{FID} and ii) Precision-and-Recall~\cite{DBLP:journals/corr/abs-1904-06991}. Tab.~\ref{tab:fids} summarizes our results. %
On CelebA-HQ, we report a new state-of-the-art FID of $5.11$, outperforming
previous likelihood-based models as well as GANs. We also outperform
LSGM~\cite{DBLP:journals/corr/abs-2106-05931} where a latent diffusion model is
trained jointly together with the first stage. 
In contrast, we train diffusion models in a fixed space and avoid the
difficulty of weighing reconstruction quality against learning the prior over
the latent space, see Fig.~\ref{fig:firststagecomparison}-\ref{fig:perceptualcompression}.

We outperform prior diffusion based approaches on all but the LSUN-Bedrooms
dataset, where our score is close to
ADM~\cite{DBLP:journals/corr/abs-2105-05233}, despite utilizing half its
parameters and requiring 4-times less train resources (see Appendix~\ref{suppsubsubsec:compute}). 
Moreover, \emph{LDMs} consistently improve upon GAN-based methods in Precision and Recall, thus confirming the advantages of their mode-covering likelihood-based training objective over adversarial approaches.
In Fig.~\ref{fig:samples_mix} we also show qualitative results on each dataset.

\subsection{Conditional Latent Diffusion}
\label{subsec:conditionallatentdiffusion}
\bboxandtexttoimgsamples
\subsubsection{Transformer Encoders for LDMs}
\label{subsubsec:crossattn2img}
\vspace{-0.5em}
By introducing cross-attention based conditioning into LDMs
we open them up for various
 conditioning modalities previously unexplored for diffusion models. 
For \textbf{text-to-image} image modeling, we train a 1.45B parameter \emph{KL}-regularized \emph{LDM} conditioned on language prompts on LAION-400M~\cite{schuhmann2021laion400m}. 
We employ the BERT-tokenizer~\cite{DBLP:journals/corr/abs-1810-04805} and implement $\tau_\theta$ as a transformer~\cite{DBLP:conf/nips/VaswaniSPUJGKP17} to infer a latent code which 
is mapped into the UNet via (multi-head) cross-attention (Sec.~\ref{subsec:conditioning}).
This combination of domain specific experts for learning a language representation and visual synthesis results in a powerful model, which generalizes well to complex, user-defined text prompts, \cf Fig.~\ref{fig:bboxandtxt2img} and \ref{fig:text2img_samples}. %
For quantitative analysis, we follow prior work and evaluate text-to-image generation on the MS-COCO~\cite{DBLP:journals/corr/LinMBHPRDZ14} validation set, where our model improves upon powerful AR~\cite{DBLP:journals/corr/abs-2102-12092,DBLP:journals/corr/abs-2105-13290} and GAN-based~\cite{DBLP:journals/corr/abs-2111-13792} methods, \cf Tab.~\ref{tab:txt2img}. We note that applying classifier-free diffusion guidance~\cite{ho2021classifier} greatly boosts sample quality, such that the guided \emph{LDM-KL-8-G} is on par with the recent state-of-the-art AR~\cite{DBLP:journals/corr/abs-2203-13131} and diffusion models~\cite{DBLP:journals/corr/abs-2112-10741} for text-to-image synthesis, while substantially reducing parameter count.
To further analyze the flexibility of the cross-attention based conditioning mechanism we also train models to synthesize images based on \textbf{semantic layouts} on OpenImages ~\cite{DBLP:journals/corr/abs-1811-00982},
and finetune on COCO~\cite{DBLP:conf/cvpr/CaesarUF18}, see Fig.~\ref{fig:bboxandtxt2img}.
See Sec.~\ref{suppsec:bboxtoimage} for the quantitative evaluation and implementation details.

Lastly, following prior work~\cite{DBLP:journals/corr/abs-2105-05233,bigganbrock,DBLP:journals/corr/abs-2012-09841,DBLP:journals/corr/abs-2108-08827}, we evaluate our best-performing \textbf{class-conditional} ImageNet models with $f\in\{4,8\}$ from Sec.~\ref{subsec:reduced_compute} %
in Tab.~\ref{tab:imagenet_main_numbers}, Fig.~\ref{fig:samples_mix} and 
Sec.~\ref{suppsec:cin}. Here we outperform the state of the art diffusion model ADM~\cite{DBLP:journals/corr/abs-2105-05233} while significantly reducing computational requirements and parameter count, \cf Tab~\ref{tab:compute_vs_fid}.
\cinmainmetrics
\vspace{-1.0em}
\subsubsection{Convolutional Sampling Beyond $256^2$}
\label{subsubsec:beyond}
\vspace{-0.75em}
By concatenating spatially aligned conditioning information to the input of
$\model$, \emph{LDMs} can serve
as efficient general-purpose image-to-image translation models.
We use this to train models for semantic synthesis, super-resolution (Sec.~\ref{subsec:superres}) and inpainting (Sec.~\ref{subsec:inpainting}).
For semantic synthesis, we use images of landscapes paired with semantic maps \cite{spade, DBLP:journals/corr/abs-2012-09841}
and concatenate downsampled versions of the semantic maps with the latent image
representation of a $f=4$ model (VQ-reg., see Tab.~\ref{tab:firststagetablecomplete}).
We train on an input resolution of $256^2$ (crops from $384^2$) but find that our model generalizes to larger resolutions
and can
generate images up to the megapixel regime when evaluated in a convolutional
manner (see Fig.~\ref{fig:thicksample}).
We exploit this behavior to also apply the super-resolution models in
Sec.~\ref{subsec:superres} and the inpainting models in
Sec.~\ref{subsec:inpainting} to generate large images between $512^2$ and $1024^2$.
For this application, the signal-to-noise ratio (induced by the scale of the latent space)
significantly affects the results.
In Sec.~\ref{suppsec:rescale} we illustrate this
when learning an LDM on (i) the latent space as provided by a $f=4$ model (KL-reg., see Tab.~\ref{tab:firststagetablecomplete}),
and (ii) a rescaled version, scaled by the component-wise standard deviation. 

The latter, in combination with classifier-free guidance~\cite{ho2021classifier}, also enables the direct synthesis of 
$>256^2$ images for the text-conditional \emph{LDM-KL-8-G} as in Fig.~\ref{fig:text2img_conv}.

\thicksample
\vspace{-0.5em}
\subsection{Super-Resolution with Latent Diffusion}
\label{subsec:superres}
\vspace{-0.5em}
LDMs can be efficiently trained for super-resolution by diretly conditioning on low-resolution images via concatenation (\cf Sec.~\ref{subsec:conditioning}).
In a first experiment, we follow SR3 \cite{DBLP:journals/corr/abs-2104-07636} and fix the image degradation to a bicubic interpolation with $4\times$-downsampling and train on ImageNet following SR3's data processing pipeline. We use the $f=4$ autoencoding model pretrained on OpenImages (VQ-reg., \cf Tab.~\ref{tab:firststagetablecomplete}) and concatenate the low-resolution conditioning $y$ and the inputs to the UNet, \ie $\conditioner$ is the identity.
Our qualitative and quantitative results (see Fig.~\ref{fig:srimagenet} and Tab.~\ref{tab:srtable}) show competitive performance and LDM-SR outperforms SR3 in FID while SR3 has a better IS. 
A simple image regression model achieves the highest PSNR and SSIM scores; 
however these metrics do not align well with human perception \cite{lpips} and favor blurriness over imperfectly aligned high frequency details \cite{DBLP:journals/corr/abs-2104-07636}. 
Further, we conduct a user study comparing the pixel-baseline with LDM-SR. We follow SR3~\cite{DBLP:journals/corr/abs-2104-07636} where human subjects were shown a low-res image in between two high-res images and asked for preference. The results in Tab.~\ref{tab:user_study} affirm the good performance of LDM-SR. 
PSNR and SSIM can be pushed by using a post-hoc guiding mechanism \cite{DBLP:journals/corr/abs-2105-05233} and we implement this \emph{image-based guider} via a perceptual loss, see Sec.~\ref{suppsec:superres}.
\srimagenet
\userstudy
Since the bicubic degradation process does not generalize well to images which do not follow this pre-processing, we also train a generic model, \emph{LDM-BSR}, by using more diverse degradation. The results are shown in Sec.~\ref{suppsubsubsec:bsr}.
\srtable

\subsection{Inpainting with Latent Diffusion}
\label{subsec:inpainting}
\vspace{-0.5em}
Inpainting is the task of filling masked regions of an image with new content either
because parts of the image are are corrupted or to
replace existing but undesired content within the image. We evaluate how our
general approach for conditional image generation compares to more specialized,
state-of-the-art approaches for this task. Our evaluation follows the protocol
of LaMa\cite{lama}, a recent inpainting model that introduces a specialized
architecture relying on Fast Fourier Convolutions\cite{Chi2020FastFC}. 
The exact training \& evaluation protocol on Places\cite{places} is described in Sec.~\ref{suppsec:inpainting}.

We first analyze the effect of different design choices for
the first stage. 
\inpaintingefficiency
\inpaintingremoval
In particular, we compare the inpainting efficiency of \emph{LDM-1} 
(\ie a pixel-based conditional DM) with
\emph{LDM-4}, for both \emph{KL} and \emph{VQ} regularizations,
as well as \emph{VQ-LDM-4} without any attention in the first stage
(see Tab.~\ref{tab:firststagetablecomplete}), where the latter reduces GPU memory for decoding at high resolutions. For comparability, we fix the number of parameters for all models.
Tab.~\ref{inpaintingefficiency} reports the training and sampling throughput 
at resolution $256^{2}$ and $512^{2}$, the total training time in hours
per epoch and the FID score on the validation split after six epochs. Overall,
we observe a speed-up of at least $2.7\times$ between pixel- and latent-based
diffusion models while improving FID scores by a factor of at least $1.6\times$. 

The comparison with other inpainting approaches in Tab.~\ref{inpaintingtable}
shows that our model with attention improves the overall image quality as
measured by FID over that of \cite{lama}. LPIPS between the unmasked
images and our samples is slightly higher than that of \cite{lama}.
We attribute this to \cite{lama} only producing a
single result which tends to recover more of an average image
compared to the diverse results produced by our LDM 
\cf Fig.~\ref{fig:suppinpaintingsamples}. 
Additionally in a user study (Tab.~\ref{tab:user_study}) human subjects favor our results over those of \cite{lama}.

Based on these initial results, we also trained a larger diffusion model
(\emph{big} in Tab.~\ref{inpaintingtable}) in the
latent space of the \emph{VQ}-regularized first stage without attention.
Following \cite{DBLP:journals/corr/abs-2105-05233}, the UNet of this diffusion model uses attention layers on
three levels of its feature hierarchy, the BigGAN \cite{bigganbrock} residual block
for up- and downsampling and has 387M parameters instead of 215M.
After training, we noticed a discrepancy in the quality of samples produced at
resolutions $256^2$ and $512^2$, which we hypothesize to be caused by the
additional attention modules. However, fine-tuning the model for half an epoch
at resolution $512^2$ allows the model to adjust to the new feature statistics
and sets a new state of the art FID on image inpainting (\emph{big, w/o attn, w/
ft} in Tab.~\ref{inpaintingtable}, Fig.~\ref{inpaintingremoval}.).

\inpaintingtable

\vspace{-0.5em}
\section{Limitations \& Societal Impact}
\label{sec:limitations}
\enlargethispage{\baselineskip}
\paragraph{Limitations}
While LDMs significantly reduce computational requirements compared to pixel-based approaches, 
their sequential sampling process is still slower than that of GANs.
Moreover, the use of LDMs can be questionable when high precision is required: 
although the loss of image quality is very small in our $f=4$ autoencoding models (see Fig.~\ref{fig:firststagecomparison}), 
their reconstruction capability can become a bottleneck for tasks that require fine-grained accuracy in pixel space. 
We assume that our superresolution models (Sec.~\ref{subsec:superres}) are already somewhat limited in this respect.

\vspace{-1.0em}
\paragraph{Societal Impact}
Generative models for media like imagery are a double-edged sword: On the one hand, they enable various creative applications, 
and in particular approaches like ours that reduce the cost of training and inference have the potential 
to facilitate access to this technology and democratize its exploration. 
On the other hand, it also means that it becomes easier to create and disseminate manipulated data or spread misinformation and spam. 
In particular, the deliberate manipulation of images (``deep fakes'') is a common problem in this context, 
and women in particular are disproportionately affected by it \cite{denton2021workshop, franks2018sex}.

Generative models can also reveal their training data \cite{carlini2021extracting, tinsley2021face}, 
which is of great concern when the data contain sensitive or personal information 
and were collected without explicit consent. 
However, the extent to which this also applies to DMs of images is not yet fully understood.

Finally, deep learning modules tend to reproduce or exacerbate biases that are already present in the data \cite{torralba2011unbiased, jain2020imperfect, esser2020note}. 
While diffusion models achieve better coverage of the data distribution than \eg GAN-based approaches, 
the extent to which our two-stage approach that combines adversarial training and a likelihood-based objective 
misrepresents the data remains an important research question.

For a more general, detailed discussion of the ethical considerations of deep generative models, see \eg \cite{denton2021workshop}.

\vspace{-0.5em}
\section{Conclusion}
\label{sec:conclusion}
\enlargethispage{\baselineskip}
\vspace{-0.5em}
We have presented latent diffusion models, a simple and efficient way to significantly improve both the
training and sampling efficiency of denoising diffusion models without
degrading their quality. Based on this and our cross-attention
conditioning mechanism, our experiments could demonstrate favorable results
compared to state-of-the-art methods across a wide range of conditional image
synthesis tasks without task-specific architectures. 
\blfootnote{This work has been supported by the German Federal Ministry for Economic Affairs and Energy within the project ’KI-Absicherung - Safe AI for automated driving’ and by the German Research Foundation (DFG) project 421703927.}

\newpage

\FloatBarrier

{\small
\bibliographystyle{ieee_fullname}
\bibliography{goodreferences}
}

\newpage

\FloatBarrier

\onecolumn

\appendix

\vspace{-2cm}
\begin{center}
\LARGE\textbf{Appendix}
\end{center}
\enlargethispage{\baselineskip}
\landscapestune
\texttoimgconvsamples
\newpage
\FloatBarrier

\section{Changelog}
Here we list changes between this version (\url{https://arxiv.org/abs/2112.10752v2}) of the paper and the previous version, \ie \url{https://arxiv.org/abs/2112.10752v1}. 

\begin{itemize}
\item We updated the results on text-to-image synthesis in Sec.~\ref{subsec:conditionallatentdiffusion} which were obtained by training a new, larger model (1.45B parameters). This also includes a new comparison to very recent competing methods on this task that were published on arXiv at the same time as (\cite{DBLP:journals/corr/abs-2112-10741, DBLP:journals/corr/abs-2111-13792}) or after (\cite{DBLP:journals/corr/abs-2203-13131}) the publication of our work.
\item We updated results on class-conditional synthesis on ImageNet in Sec.~\ref{subsec:reduced_compute}, Tab.~\ref{tab:imagenet_main_numbers} (see also Sec.~\ref{suppsec:cin}) obtained by retraining the model with a larger batch size.
The corresponding qualitative results in Fig.~\ref{fig:imagenet_samples_1} and Fig.~\ref{fig:imagenet_samples_2} were also updated. Both the updated text-to-image and the class-conditional model now use classifier-free guidance~\cite{ho2021classifier} as a measure to increase visual fidelity.
\item We conducted a user study (following the scheme suggested by Saharia et al~\cite{DBLP:journals/corr/abs-2104-07636}) which provides additional evaluation for our inpainting (Sec.~\ref{subsec:inpainting}) and superresolution models (Sec.~\ref{subsec:superres}).
\item Added Fig.~\ref{fig:text2img_samples} to the main paper, moved Fig.~\ref{fig:srgeneralization} to the appendix, added Fig.~\ref{fig:text2img_conv} to the appendix.
\end{itemize}

\newcommand{\normal}{\mathcal{N}}
\newcommand{\I}{\mathbb{I}}
\newcommand{\snr}{\text{SNR}}
\section{Detailed Information on Denoising Diffusion Models}
\label{suppsec:dmdetails}
Diffusion models can be specified in terms of a signal-to-noise ratio
$\snr(t)=\frac{\alpha_t^2}{\sigma_t^2}$ consisting
of sequences $(\alpha_t)_{t=1}^T$ and $(\sigma_t)_{t=1}^T$ which, starting from
a data sample $x_0$, define a forward diffusion process $q$ as
\begin{equation}
  q(x_t \vert x_0) = \normal(x_t \vert \alpha_t x_0, \sigma_t^2 \I)
\end{equation}
with the Markov structure for $s < t$:
\begin{align}
  q(x_t \vert x_s) &= \normal(x_t \vert \alpha_{t\vert s} x_s, \sigma_{t\vert s}^2 \I) \\
  \alpha_{t\vert s} &= \frac{\alpha_t}{\alpha_s} \\
  \sigma_{t\vert s}^2 &= \sigma_t^2 - \alpha_{t\vert s}^2 \sigma_s^2
\end{align}

Denoising diffusion models are generative models $p(x_0)$ which revert this
process with a similar Markov structure running backward in time, \ie they are
specified as
\begin{equation}
  p(x_0) = \int_{z} p(x_T) \prod_{t=1}^T p(x_{t-1} \vert x_t)
\end{equation}
The evidence lower bound (ELBO) associated with this model then decomposes over
the discrete time steps as
\begin{equation}
  -\log p(x_0) \le \KL(q(x_T \vert x_0) \vert p(x_T)) + \sum_{t=1}^T
  \expect_{q(x_t \vert x_0)} \KL(q(x_{t-1} \vert x_t, x_0) \vert p(x_{t-1}
  \vert x_t))
\end{equation}
The prior $p(x_T)$ is typically choosen as a standard normal distribution and
the first term of the ELBO then depends only on the final signal-to-noise ratio
$\snr(T)$. To minimize the remaining terms, a common choice to parameterize
$p(x_{t-1} \vert x_t)$ is to specify it in terms of the true posterior
$q(x_{t-1} \vert x_t, x_0)$ but with the unknown $x_0$ replaced by an estimate
$x_\theta(x_t, t)$ based on the current step $x_t$. This gives \cite{DBLP:journals/corr/abs-2107-00630}
\begin{align}
  p(x_{t-1} \vert x_t) &\coloneqq q(x_{t-1} \vert x_t, x_\theta(x_t, t)) \\
  &= \normal(x_{t-1} \vert \mu_\theta(x_t, t), \sigma_{t\vert t-1}^2
  \frac{\sigma_{t-1}^2}{\sigma_t^2}\I),
\end{align}
where the mean can be expressed as
\begin{equation}
  \mu_\theta(x_t, t) = \frac{\alpha_{t\vert t-1} \sigma_{t-1}^2}{\sigma_t^2}
  x_t + \frac{\alpha_{t-1} \sigma_{t\vert t-1}^2}{\sigma_t^2} x_\theta(x_t, t).
\end{equation}
In this case, the sum of the ELBO simplify to
\begin{equation}
  \sum_{t=1}^T \expect_{q(x_t \vert x_0)} \KL(q(x_{t-1} \vert x_t, x_0) \vert p(x_{t-1}) =
  \sum_{t=1}^T \expect_{\normal(\epsilon \vert 0, \I)} \frac{1}{2}(\snr(t-1) -
  \snr(t)) \Vert x_0 - x_\theta(\alpha_t x_0 + \sigma_t \epsilon, t) \Vert^2
\end{equation}
Following \cite{DBLP:conf/nips/HoJA20}, we use the reparameterization
\begin{equation}
  \epsilon_\theta(x_t, t) = (x_t - \alpha_t x_\theta(x_t, t))/\sigma_t
\end{equation}
to express the reconstruction term as a denoising objective,
\begin{equation}
  \Vert x_0 - x_\theta(\alpha_t x_0 + \sigma_t \epsilon, t) \Vert^2 =
  \frac{\sigma_t^2}{\alpha_t^2} \Vert \epsilon - \epsilon_\theta(\alpha_t x_0 +
  \sigma_t \epsilon, t) \Vert^2
\end{equation}
and the reweighting, which assigns each of the terms the same weight and
results in Eq.~\eqref{eq:dmloss}.

\newpage
\section{Image Guiding Mechanisms}
\label{subsec:imageguiding}
\convolutionalguiding
An intriguing feature of diffusion models is that unconditional models can be conditioned at test-time \cite{DBLP:journals/corr/abs-2011-13456, DBLP:journals/corr/Sohl-DicksteinW15, DBLP:journals/corr/abs-2105-05233}.
In particular, \cite{DBLP:journals/corr/abs-2105-05233} presented an algorithm to guide both unconditional and conditional models
trained on the ImageNet dataset with a classifier $\log p_{\Phi}(y\vert x_t)$, trained on each $x_t$ of the diffusion process.
We directly build on this formulation and introduce post-hoc \emph{image-guiding}:

For an epsilon-parameterized model with fixed variance, the guiding algorithm as introduced in \cite{DBLP:journals/corr/abs-2105-05233} reads: 
\begin{equation}
\hat{\epsilon} \leftarrow \model(z_t, t) + \sqrt{1-\alpha_t^2}\; \nabla_{z_t} \log p_{\Phi}(y\vert z_t) \; .
\label{eq:epsilonupdate}
\end{equation}

This can be interpreted as an update correcting the ``score'' $\epsilon_\theta$ with a conditional distribution
$\log p_{\Phi}(y\vert z_t)$. 

So far, this scenario has only been applied to single-class classification models. %
We re-interpret the guiding distribution $p_{\Phi}(y\vert T(\decoder(z_0(z_t))))$
as a general purpose image-to-image translation task given a target image $y$, 
where $T$ can be any differentiable transformation adopted to the image-to-image translation task at hand, 
such as the identity, a downsampling operation or similar.
\newline
As an example, we can assume a Gaussian guider with fixed variance $\sigma^2=1$, such that
\begin{equation}
\log p_{\Phi}(y\vert z_t) = -\frac{1}{2}\Vert y- T(\decoder(z_0(z_t))) \Vert^2_2
\end{equation}
becomes a $L_2$ regression objective. %

Fig.~\ref{fig:convolutionalguiding} demonstrates how this formulation can serve as an upsampling mechanism
of an unconditional model trained on $256^2$ images, where unconditional samples of size $256^2$ guide the 
convolutional synthesis of $512^2$ images
 and $T$ is a $2\times$ bicubic downsampling.
Following this motivation, 
we also experiment with a perceptual similarity guiding and replace the $L_2$ objective with
the LPIPS \cite{lpips} metric, see Sec.~\ref{subsec:superres}.

\newpage
\section{Additional Results}

\subsection{Choosing the Signal-to-Noise Ratio for High-Resolution Synthesis}
\label{suppsec:rescale}
\convolutionalrescaling
As discussed in Sec.~\ref{subsubsec:beyond}, the signal-to-noise ratio induced by the variance of the latent
space (\ie $\text{Var(z)}/\sigma^2_t$) significantly affects the results for convolutional sampling. 
For example, when training a LDM directly in the latent space of a KL-regularized model (see Tab.~\ref{tab:firststagetablecomplete}), 
this ratio is very high, such that the model allocates a lot of semantic detail early on in the reverse denoising process.
In contrast, when rescaling the latent space by the component-wise standard deviation of the latents as described in Sec.~\ref{suppsec:ae}, 
the SNR is descreased. We illustrate the effect on convolutional sampling for semantic image synthesis
in Fig.~\ref{fig:convolutionalrescaling}. Note that the VQ-regularized space has a variance close to $1$, 
such that it does not have to be rescaled.

\subsection{Full List of all First Stage Models}
We provide a complete list of various autoenconding  models trained on the OpenImages dataset in Tab.~\ref{tab:firststagetablecomplete}.

\firststagetablecomplete

\subsection{Layout-to-Image Synthesis}
\label{suppsec:bboxtoimage}
\layouttoimagesamples
\layouttoimagefids
Here we provide the quantitative evaluation and additional samples for our layout-to-image models from Sec.~\ref{subsubsec:crossattn2img}. We train a model on the COCO~\cite{DBLP:conf/cvpr/CaesarUF18} and one on the OpenImages~\cite{DBLP:journals/corr/abs-1811-00982} dataset, which we subsequently additionally finetune on COCO. Tab~\ref{tab:layout2img} shows the result. Our COCO model reaches the performance of recent state-of-the art models in layout-to-image synthesis, when following their training and evaluation protocol~\cite{DBLP:conf/aaai/SylvainZBH021}. When finetuning from the OpenImages model, we surpass these works. Our OpenImages model surpasses the results of Jahn et al~\cite{DBLP:journals/corr/abs-2105-06458} by a margin of nearly 11 in terms of FID. In Fig.~\ref{fig:lay2img_samples} we show additional samples of the model finetuned on COCO.

\subsection{Class-Conditional Image Synthesis on ImageNet}
\label{suppsec:cin}
Tab.~\ref{tab:imagenet_numbers} contains the results for our class-conditional LDM measured in FID and Inception score (IS). 
LDM-8 requires significantly fewer parameters and compute requirements (see Tab.~\ref{tab:compute_vs_fid}) to achieve very competitive performance. Similar to previous work, we can further boost the performance by training a classifier on each noise scale and guiding with it, see Sec.~\ref{subsec:imageguiding}.
Unlike the pixel-based methods, this classifier is trained very cheaply in latent space.
For additional qualitative results, see Fig.~\ref{fig:imagenet_samples_1} and Fig.~\ref{fig:imagenet_samples_2}.

\cinmetrics

\subsection{Sample Quality vs. V100 Days (Continued from Sec.~\ref{subsec:reduced_compute})}
\cinprogressvdays
For the assessment of sample quality over the training progress in Sec.~\ref{subsec:reduced_compute}, we reported FID and IS scores as a function of train steps. Another possibility is to report these metrics over the used resources in V100 days. Such an analysis is additionally provided in Fig.~\ref{fig:cin_traincourse_v100}, showing qualitatively similar results.

\subsection{Super-Resolution}
\label{suppsec:superres}
\srsuppptable
For better comparability between LDMs and diffusion models in pixel space, we extend our analysis from Tab.~\ref{tab:srtable} by comparing a diffusion model trained for the same number of steps and with a comparable number \footnote{It is not possible to exactly match both architectures since the diffusion model operates in the pixel space} of parameters to our LDM. The results of this comparison are shown in the last two rows of Tab.~\ref{tab:srsupptable} and demonstrate that LDM achieves better performance while allowing for significantly faster sampling. 
A qualitative comparison is given in Fig.~\ref{suppsrimagenet} which shows random samples from both LDM and the diffusion model in pixel space.

\subsubsection{LDM-BSR: General Purpose SR Model via Diverse Image Degradation}
\label{suppsubsubsec:bsr}
\srgeneralization
To evaluate generalization of our LDM-SR, we apply it both on synthetic LDM samples from a class-conditional ImageNet model (Sec.~\ref{subsec:reduced_compute}) and images crawled from the internet.
Interestingly, we observe that LDM-SR, trained only with a bicubicly
downsampled conditioning as in \cite{DBLP:journals/corr/abs-2104-07636}, does
not generalize well to images which do not follow this pre-processing.
Hence, to obtain a superresolution model for a wide range of real world images, which can contain complex superpositions of camera noise, compression artifacts, blurr and interpolations, we replace the bicubic downsampling operation in LDM-SR with the degration pipeline from \cite{bsrgan}. The BSR-degradation process is a degradation pipline which applies JPEG compressions noise, camera sensor noise, different image interpolations for downsampling, Gaussian blur kernels and Gaussian noise in a random order to an image. We found that using the bsr-degredation process with the original parameters as in \cite{bsrgan} leads to a very strong degradation process. Since a more moderate degradation process seemed apppropiate for our application, we adapted the parameters of the bsr-degradation (our adapted degradation process can be found in our code base at \url{\projectpath}).
Fig.~\ref{fig:srgeneralization} illustrates the effectiveness of this approach by directly comparing \emph{LDM-SR} with \emph{LDM-BSR}. The latter produces images much sharper than the models confined to a
fixed pre-processing, making it suitable for real-world applications.
Further results of \emph{LDM-BSR} are shown on LSUN-cows in Fig.~\ref{fig:supercows}.

\supercows
\suppsrimagenetbaseline

\newpage
\section{Implementation Details and Hyperparameters}
\subsection{Hyperparameters}
We provide an overview of the hyperparameters of all trained \emph{LDM} models in Tab.~\ref{tab:uncond_hyperparams}, Tab.~\ref{tab:cin_hyperparams}, Tab.~\ref{tab:celeba_hyperparams} and Tab.~\ref{tab:cond_hyperparams}.
\uncondhyperparams
\cinhyperparams
\celebahyperparams
\condhyperparams
\subsection{Implementation Details}
\label{suppsec:implementation_details}

\subsubsection{Implementations of $\tau_\theta$ for conditional \emph{LDMs}}
\label{suppsubsubsec:transformer}
For the experiments on text-to-image and layout-to-image (Sec.~\ref{subsubsec:crossattn2img}) synthesis, 
we implement the conditioner $\conditioner$ as an unmasked transformer which processes 
a tokenized version of the input $y$ and produces an output $\zeta := \conditioner(y)$, 
where $\zeta \in \R^{M \times d_\tau}$.
More specifically, the transformer is implemented from $N$ transformer blocks consisting of 
global self-attention layers, layer-normalization and 
position-wise MLPs as follows\footnote{adapted from \url{https://github.com/lucidrains/x-transformers}}: 

\begin{align}
&\zeta \leftarrow \text{TokEmb}(y) + \text{PosEmb(y)} \\
&\text{for } i=1,\dots,N: \nonumber \\
 &\quad \zeta_1 \leftarrow \text{LayerNorm}(\zeta) \\
 &\quad \zeta_2 \leftarrow \text{MultiHeadSelfAttention}(\zeta_1) + \zeta \\
 &\quad \zeta_3 \leftarrow \text{LayerNorm}(\zeta_2)  \\
 &\quad \zeta \leftarrow \text{MLP}(\zeta_3) + \zeta_2  \\ 
& \zeta \leftarrow \text{LayerNorm}(\zeta)  \\
\end{align}

With $\zeta$ available, the conditioning is mapped into the UNet via the cross-attention mechanism 
as depicted in Fig.~\ref{fig:conditioning}. We modify the ``ablated UNet''~\cite{DBLP:journals/corr/abs-2105-05233} architecture
and replace the self-attention layer with a shallow (unmasked) transformer consisting 
of $T$ blocks with alternating layers of (i) self-attention, (ii) a position-wise MLP and
(iii) a cross-attention layer; see Tab.~\ref{tab:transformertable}. Note that without (ii) and (iii), this
architecture is equivalent to the ``ablated UNet''.

While it would be possible to increase the representational power of $\conditioner$ 
by additionally conditioning on the time step $t$, we do not pursue this choice as it reduces the speed of inference. 
We leave a more detailed analysis of this modification to future work.

For the text-to-image model, we rely on a publicly available\footnote{\url{https://huggingface.co/transformers/model_doc/bert.html\#berttokenizerfast}} tokenizer 
\cite{DBLP:journals/corr/abs-1910-03771}. 
The layout-to-image model discretizes the spatial locations of the bounding boxes 
and encodes each box as a $(l, b, c)$-tuple, where $l$ denotes the (discrete) top-left and $b$ the 
bottom-right position. Class information is contained in $c$. \\
See Tab.~\ref{tab:transformerhyperparams} for the hyperparameters of $\conditioner$ and Tab.~\ref{tab:cin_hyperparams} for those of the UNet for both of the above tasks.\\

Note that the class-conditional model as described in Sec.~\ref{subsec:reduced_compute} %
is also implemented via cross-attention, where $\conditioner$ is a single learnable
embedding layer with a dimensionality of 512, mapping classes $y$ to 
$\zeta \in \R^{1\times 512}$.

\transformertable
\transformerhyperparams

\subsubsection{Inpainting}
\label{suppsec:inpainting}
\suppinpaintingsamples
\suppinpaintingremoval

For our experiments on image-inpainting in Sec.~\ref{subsec:inpainting}, we
used the code of \cite{lama} to generate synthetic masks. We use a fixed set of
2k validation and 30k testing samples from Places\cite{places}. During
training, we use random crops of size $256\times 256$ and evaluate on crops of
size $512\times 512$. This follows the training and testing protocol in
\cite{lama} and reproduces their reported metrics (see $^\dagger$ in
Tab.~\ref{inpaintingtable}). We include additional qualitative results of
\emph{LDM-4, w/ attn} in
Fig.~\ref{fig:suppinpaintingsamples} and of \emph{LDM-4, w/o attn, big, w/ ft} in
Fig.~\ref{suppinpaintingremoval}.

\subsection{Evaluation Details}
\label{suppsubsec:eval}
\noindent This section provides additional details on evaluation for the experiments shown in Sec.~\ref{sec:experiments}.

\subsubsection{Quantitative Results in Unconditional and Class-Conditional Image Synthesis}
\label{suppsubsubsec:fids}
We follow common practice and estimate the statistics for calculating the FID-, Precision- and Recall-scores~\cite{FID,DBLP:journals/corr/abs-1904-06991} shown in Tab.~\ref{tab:fids} and \ref{tab:imagenet_numbers} based on 50k samples from our models and the entire training set of each of the shown datasets. For calculating FID scores we use the \texttt{torch-fidelity} package~\cite{obukhov2020torchfidelity}. However, since different data processing pipelines might lead to different results~\cite{parmar2021cleanfid}, we also evaluate our models with the script provided by Dhariwal and Nichol~\cite{DBLP:journals/corr/abs-2105-05233}. We find that results mainly coincide, except for the ImageNet and LSUN-Bedrooms datasets, where we notice slightly varying scores of 7.76 (\texttt{torch-fidelity}) vs. 7.77 (Nichol and Dhariwal) and 2.95 vs 3.0. For the future we emphasize the importance of a unified procedure for sample quality assessment. Precision and Recall are also computed by using the script provided by Nichol and Dhariwal.

\subsubsection{Text-to-Image Synthesis}
\label{suppsubsubsec:text2img}
Following the evaluation protocol of~\cite{DBLP:journals/corr/abs-2102-12092} we compute FID and Inception Score for the Text-to-Image models from Tab.~\ref{tab:txt2img} by comparing generated samples with 30000 samples from the validation set of the MS-COCO dataset~\cite{DBLP:journals/corr/LinMBHPRDZ14}. FID and Inception Scores are computed with \texttt{torch-fidelity}.

\subsubsection{Layout-to-Image Synthesis}
\label{suppsubsubsec:layout2img}
For assessing the sample quality of our Layout-to-Image models from Tab.~\ref{tab:layout2img} on the COCO dataset, we follow common practice~\cite{DBLP:conf/aaai/SylvainZBH021,DBLP:journals/corr/abs-2105-06458,DBLP:journals/corr/abs-2003-11571} and compute FID scores the 2048 unaugmented examples of the COCO Segmentation Challenge split. To obtain better comparability, we use the exact same samples as in~\cite{DBLP:journals/corr/abs-2105-06458}. For the OpenImages dataset we similarly follow their protocol and use 2048 center-cropped test images from the validation set.

\subsubsection{Super Resolution}
\label{suppsubsubsec:sr}
We evaluate the super-resolution models on ImageNet following the pipeline suggested in \cite{DBLP:journals/corr/abs-2104-07636}, \ie images 
with a shorter size less than $256$ px are removed (both for training and evaluation). On ImageNet, the low-resolution images are 
produced using bicubic interpolation with anti-aliasing.
FIDs are evaluated using \texttt{torch-fidelity}~\cite{obukhov2020torchfidelity}, 
and we produce samples on the validation split.
For FID scores, we additionally compare to reference features computed on the train split, see Tab.~\ref{tab:srtable} and Tab.~\ref{tab:srsupptable}.

\subsubsection{Efficiency Analysis}
\label{suppsubsubsec:compute}
For efficiency reasons we compute the sample quality metrics plotted in Fig.~\ref{fig:cin_traincourse}, \ref{fig:cin_traincourse_v100} and \ref{fig:speedplot} based on 5k samples. Therefore, the results might vary from those shown in Tab.~\ref{tab:fids} and \ref{tab:imagenet_numbers}. All models have a comparable number of parameters as provided in Tab.~\ref{tab:cin_hyperparams} and \ref{tab:celeba_hyperparams}. We maximize the learning rates of the individual models such that they still train stably. Therefore, the learning rates slightly vary between different runs \cf Tab.~\ref{tab:cin_hyperparams} and \ref{tab:celeba_hyperparams}. 

\subsubsection{User Study}
\label{suppsubsubsec:user_study}
For the results of the user study presented in Tab.~\ref{tab:user_study} we followed the protocoll of \cite{DBLP:journals/corr/abs-2104-07636} and and use the 2-alternative force-choice paradigm to assess human preference scores for two distinct tasks. In Task-1 subjects were shown a low resolution/masked image between the corresponding ground truth high resolution/unmasked version and a synthesized image, which was generated by using the middle image as conditioning. For SuperResolution subjects were asked: \emph{'Which of the two images is a better high quality version of the low resolution image in the middle?'}. For Inpainting we asked \emph{'Which of the two images contains more realistic inpainted regions of the image in the middle?'}. In Task-2, humans were similarly shown the low-res/masked version and asked for preference between two corresponding images generated by the two competing methods. As in \cite{DBLP:journals/corr/abs-2104-07636} humans viewed the images for 3 seconds before responding.

\newpage
\section{Computational Requirements}
\label{suppsec:compute2}
\computevsfid
In Tab~\ref{tab:compute_vs_fid} we provide a more detailed analysis on our used compute ressources and compare our best performing models on the CelebA-HQ, FFHQ, LSUN and ImageNet datasets with the recent state of the art models by using their provided numbers, \cf \cite{DBLP:journals/corr/abs-2105-05233}.  As they report their used compute in V100 days and we train all our models on a single NVIDIA A100 GPU, we convert the A100 days to V100 days by assuming a $\times 2.2$ speedup of A100 vs V100~\cite{a100tov100}\footnote{This factor corresponds to the speedup of the A100 over the V100 for a U-Net, as defined in Fig. 1 in ~\cite{a100tov100}}. To assess sample quality, we additionally report FID scores on the reported datasets. We closely reach the performance of state of the art methods as StyleGAN2~\cite{DBLP:journals/corr/abs-1912-04958} and ADM~\cite{DBLP:journals/corr/abs-2105-05233} while significantly reducing the required compute resources.

\newpage
\section{Details on Autoencoder Models}
\label{suppsec:ae}
We train all our autoencoder models in an adversarial manner following \cite{DBLP:journals/corr/abs-2012-09841}, such that a patch-based discriminator $\disc$ is optimized to differentiate original images from reconstructions $\decoder(\encoder(x))$. 
To avoid arbitrarily scaled latent spaces, we regularize the latent $z$ to be zero centered and obtain small variance by introducing an regularizing loss term $\lreg$. \\
We investigate two different regularization methods: (i) a low-weighted Kullback-Leibler-term between $q_{\encoder}(z \vert x) = \mathcal{N}(z; \encoder_\mu, \encoder_{\sigma^2})$ and a standard normal distribution 
$\mathcal{N}(z; 0, 1)$ as in a standard variational autoencoder~\cite{VAE,VAE2}, and, (ii) regularizing the latent space with a vector quantization layer by learning a codebook of $\vert \mathcal{Z} \vert$ different exemplars~\cite{DBLP:conf/nips/OordVK17}. \\
To obtain high-fidelity reconstructions we only use a very small regularization for both scenarios, \ie we either weight the $\KL$ term
by a factor $\sim 10^{-6}$ or choose a high codebook dimensionality $\vert \mathcal{Z} \vert$.%

The full objective to train the autoencoding model $(\encoder, \decoder)$ reads: 
\begin{equation}
L_{\text{Autoencoder}} =  \min_{\encoder, \decoder} \max_\psi \Big( \lrec(x, \decoder(\encoder(x))) - \ladv(\decoder(\encoder(x))) + \log \disc(x) + \lreg(x; \encoder, \decoder) \Big)
\label{eq:firststageloss}
\end{equation}

\paragraph{DM Training in Latent Space}
Note that for training diffusion models on the learned latent space, we again distinguish two cases
when learning $p(z)$ or $p(z\vert y)$ (Sec.~\ref{subsec:conditionallatentdiffusion}):
(i) For a KL-regularized latent space, we sample $z = \encoder_\mu(x) + \encoder_\sigma(x) \cdot \varepsilon =: \encoder(x)$, where 
$\varepsilon \sim \mathcal{N}(0, 1)$.
When rescaling the latent, we estimate the component-wise variance $$\hat{\sigma}^2 = \frac{1}{b c h w}\sum_{b, c, h, w} (z^{b, c, h, w} - \hat{\mu})^2$$
from the first batch in the data, where $\hat{\mu} = \frac{1}{b c h w}\sum_{b, c, h, w} z^{b, c, h, w}$.
The output of $\encoder$ is scaled such that the rescaled latent has unit standard deviation, \ie
$z \leftarrow \frac{z}{\hat{\sigma}} = \frac{\encoder(x)}{\hat{\sigma}}$.
(ii) For a VQ-regularized latent space, we extract $z$ \emph{before} the quantization layer and absorb the quantization operation
into the decoder, \ie it can be interpreted as the first layer of $\decoder$.

\section{Additional Qualitative Results}
Finally, we provide additional qualitative results for our landscapes model (Fig.~\ref{fig:landscapestune},~\ref{fig:landscapestunewithmap},~\ref{fig:thickersample}~and~\ref{fig:landscapes1}), our class-conditional ImageNet model (Fig.~\ref{fig:imagenet_samples_1} - \ref{fig:imagenet_samples_2}) and our unconditional models for the CelebA-HQ, FFHQ and LSUN datasets (Fig.~\ref{fig:celeba_rsamples} - \ref{fig:beds_rsamples}). 
Similar as for the inpainting model in Sec.~\ref{subsec:inpainting} we also fine-tuned the semantic landscapes model from Sec.~\ref{subsubsec:beyond} directly on $512^2$ images and depict qualitative 
results in Fig.~\ref{fig:landscapestune} and Fig.~\ref{fig:landscapestunewithmap}.
For our those models trained on comparably small datasets, we additionally show nearest neighbors in VGG~\cite{simonyan2015VGG} feature space for samples from our models in Fig.~\ref{fig:celeba_nns} - \ref{fig:churches_nns}.

\landscapestunewithmap
\thickersample
\goodlandscapes

\imagenetsamplesone
\imagenetsamplestwo

\randomcelebasamples
\randomffhqsamples
\randomchurchsamples
\randombedssamples

\nnsceleba
\nnsffhq
\nnschurches

\end{document}